\documentclass[]{uiuc_template}

\usepackage{amsmath,amsfonts,bm}

\def\eqref#1{equation~\ref{#1}}

\def\1{\bm{1}}

\DeclareMathAlphabet{\mathsfit}{\encodingdefault}{\sfdefault}{m}{sl}
\SetMathAlphabet{\mathsfit}{bold}{\encodingdefault}{\sfdefault}{bx}{n}

\usepackage{url}

\usepackage{amsmath}
\usepackage{amssymb}
\usepackage{mathtools}
\usepackage{amsthm}
\usepackage{amsthm}
\usepackage{booktabs}
\usepackage{multirow}
\usepackage{multirow}
\usepackage{makecell}
\usepackage{graphicx} 
\usepackage[table]{xcolor}

\usepackage{xspace}
\usepackage{amssymb} 
\usepackage{multirow}
\usepackage{array}
\usepackage[most]{tcolorbox}
\usepackage{listings}
\usepackage{graphicx} 
\newcolumntype{P}[1]{>{\centering\arraybackslash}p{#1}}

\usepackage{enumitem}

\usepackage{tcolorbox}
\usepackage{algorithm}
\usepackage{algpseudocode}

\usepackage{enumitem}
\usepackage{natbib} 

\usepackage{tikz}
\usepackage{wrapfig}

\usepackage[toc,page,header]{appendix}
\usepackage{minitoc}

\newcommand{\cellml}{\cellcolor[HTML]{F5E9DD}}
\newcommand{\cellmr}{\cellcolor[HTML]{D8DDD6}}

\usepackage{makecell}
\usepackage{url}
\usepackage{breakurl}
\usepackage[breaklinks]{hyperref}

\definecolor{morandiGreen}{RGB}{80, 130, 100}
\definecolor{morandiYellow}{RGB}{200, 140, 80}
\definecolor{morandiBlue}{RGB}{70, 100, 140}
\definecolor{morandiPink}{RGB}{180, 80, 100}

\definecolor{airforceblue}{rgb}{0.36, 0.54, 0.66}
\definecolor{bluegray}{rgb}{0.4, 0.6, 0.8}
\definecolor{bleudefrance}{rgb}{0.19, 0.55, 0.91}
\hypersetup{colorlinks,linkcolor={airforceblue},citecolor={bleudefrance},urlcolor={bluegray}}

\NewDocumentCommand{\heng}
{ mO{} }{\textcolor{red}{\textsuperscript{\textit{Heng}}\textsf{\textbf{\small[#1]}}}}

\NewDocumentCommand{\mengchao}
{ mO{} }{\textcolor{red}{\textsuperscript{\textit{Mengchao}}\textsf{\small[#1]}}}

\NewDocumentCommand{\cheng}
{ mO{} }{\textcolor{purple}{\textsuperscript{\textit{Cheng}}\textsf{\textbf{\small[#1]}}}}

\NewDocumentCommand{\ke}
{ mO{} }{\textcolor{orange}{\textsuperscript{\textit{Ke}}\textsf{\textbf{\small[#1]}}}}

\NewDocumentCommand{\huan}
{ mO{} }{\textcolor{blue}{\textsuperscript{\textit{Huan}}\textsf{{\small[#1]}}}}

\title{ERA: Transforming VLMs into Embodied Agents via Embodied Prior Learning and Online Reinforcement Learning}

\author[]{%
  Hanyang Chen$^{1^{*}}$, Mark Zhao$^{1^{*} \dagger}$, Rui Yang$^{1^{*}}$, Qinwei Ma$^{1,\dagger,\ddagger}$,
  Ke Yang$^1$, Jiarui Yao$^1$, Kangrui Wang$^2$\\ Hao Bai$^1$, Zhenhailong Wang$^1$, Rui Pan$^1$, Mengchao Zhang$^3$, Jose Barreiros$^3$, Aykut Onol$^3$\\
  ChengXiang Zhai$^1$, Heng Ji$^1$, Manling Li$^2$,
  Huan Zhang$^{1}$, Tong Zhang$^{1}$
}

\affiliation{$^1$UIUC, $^2$Northwestern University, $^3$Toyota Research Institute \\
\vspace{3pt} 
\href{https://embodied-reasoning-agent.github.io/}{https://embodied-reasoning-agent.github.io/}
}

\abstract{
Recent advances in embodied AI highlight the potential of vision language models (VLMs) as agents capable of perception, reasoning, and interaction in complex environments. However, top-performing systems rely on large-scale models that are costly to deploy, while smaller VLMs lack the necessary knowledge and skills to succeed. To bridge this gap, we present \textit{Embodied Reasoning Agent (ERA)}, a two-stage framework that integrates prior knowledge learning and online reinforcement learning (RL).
The first stage, \textit{Embodied Prior Learning}, distills foundational knowledge from three types of data: (1) Trajectory-Augmented Priors, which enrich existing trajectory data with structured reasoning generated by stronger models; (2) Environment-Anchored Priors, which provide in-environment knowledge and grounding supervision; and (3) External Knowledge Priors, which transfer general knowledge from out-of-environment datasets. 
In the second stage, we develop an online RL pipeline that builds on these priors to further enhance agent performance. To overcome the inherent challenges in agent RL, including long horizons, sparse rewards, and training instability, we introduce three key designs: self-summarization for context management, dense reward shaping, and turn-level policy optimization. 
 Extensive experiments on both high-level planning (EB-ALFRED) and low-level control (EB-Manipulation) tasks demonstrate that ERA-3B surpasses both prompting-based large models and previous training-based baselines. Specifically, it achieves overall improvements of 8.4\% on EB-ALFRED and 19.4\% on EB-Manipulation over GPT-4o, and exhibits strong generalization to unseen tasks. Detailed Ablation studies further validate the effectiveness of different prior datasets and agent RL designs. Overall, ERA offers a practical path toward scalable embodied intelligence, providing methodological insights for future embodied AI systems.
}

\begin{document}

\maketitle


\begin{figure}[h!]
    \centering
    \includegraphics[width=1.0\linewidth]{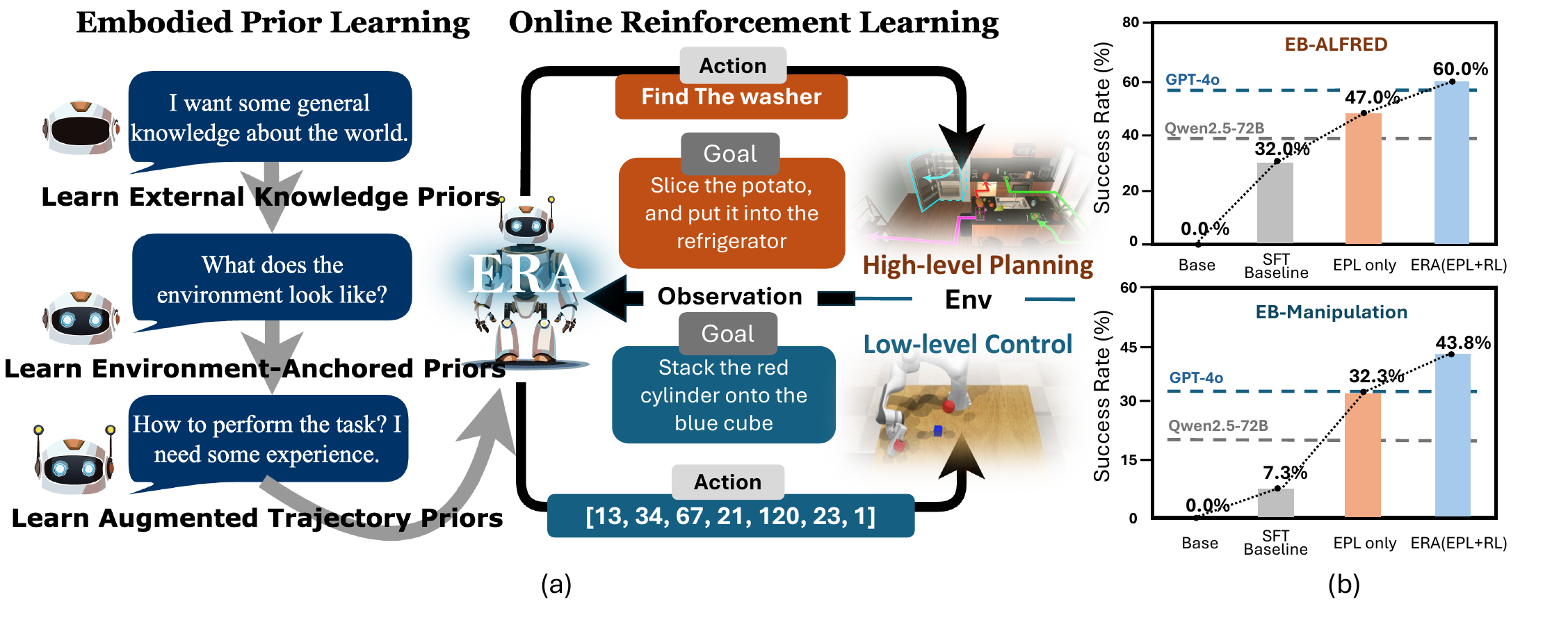}
    \vspace{-20pt}
    \caption{(a) Overview of the ERA framework: Embodied Prior Learning (EPL) finetunes on diverse data sources to provide foundational knowledge, and online RL further improves the agent. (b) ERA (i.e, EPL+RL) boosts a 3B base model to surpass GPT-4o on hold-out evaluation sets.}
    \label{fig:placeholder}
\end{figure}

\section{Introduction}
\label{sec:intro}

Vision language models (VLMs) have shown remarkable capabilities in instruction following, visual understanding, and commonsense as well as mathematical reasoning \citep{GPT-4o,liu2024llavanext,reid2024gemini,bai2025qwen25vl,zhu2025internvl3}. Building on these strengths, researchers are now exploring how to transform VLMs into embodied agents that can operate in interactive environments and tackle real-world tasks \citep{driess2023palm,huang2023voxposer,huang2024rekep,mu2024embodiedgpt,liu2024visualagentbench,kim2024openvla,szot2025multimodal}. Unlike single-turn question answering, embodied tasks require an agent to actively perceive, reason, and act within a dynamic environment to achieve its goals. This introduces new challenges for VLMs, including long-horizon planning, reliable visual grounding, and spatial awareness \citep{yang2025embodiedbench}.

Recent studies have systematically evaluated VLMs as embodied agents \citep{liu2024visualagentbench,yang2025embodiedbench,cheng2025embodiedeval,li2025sti}. With carefully designed prompting, large-scale VLMs can solve increasingly complex tasks, but their success comes at high cost: massive models demand expensive hardware, long training cycles, and costly inference, all of which hinder real-world deployment where efficiency is critical. Meanwhile, the performance gap between large and small models remains striking. For example, Claude-3.5-Sonnet achieves 64.0\% on EB-ALFRED, compared to only 4.7\% for Qwen2.5-VL-7B-Instruct \citep{yang2025embodiedbench}. This disparity highlights the limitations of smaller models, which often lack embodied knowledge, robust reasoning, and the synergy between high-level planning and low-level grounding. Thus, enabling compact models to master complex embodied tasks remains an open challenge. Recent efforts have explored reinforcement learning (RL) to enhance embodied agents’ reasoning capabilities \citep{zhai2024finetuning,kim2025robot,zhang2025embodied,wu2025reinforced,wang2025vagen}, but most approaches are restricted to static QA-style datasets or focus on either high-level or low-level reasoning in isolation. A unified framework that combines supervised fine-tuning (SFT) with RL, while systematically examining embodied data curation and the design of online agentic RL for VLM-based agents, has yet to be developed.

In this paper, we address the gap between large and small VLMs in embodied tasks with \textbf{Embodied Reasoning Agent (ERA)}, a two-stage training framework designed to unlock generalizable embodied skills in VLMs. In essence, ERA first introduces embodied priors into small VLMs and then refines them through online RL. Since general VLMs, especially small ones, lack domain-specific abilities in embodied environments, the first stage, \textit{Embodied Prior Learning}, injects tailored knowledge to strengthen reasoning, perception, and environmental understanding. We categorize three sources of prior knowledge: (\romannumeral 1) \textit{Trajectory-Augmented Priors}, which enrich existing trajectories with reasoning annotations from stronger VLMs and rule-based visual description augmentation;
(\romannumeral 2) \textit{Environment-Anchored Priors}, which provide in-environment knowledge and grounding in the form of QA pairs beyond agent-collected trajectories;
(\romannumeral 3) \textit{External Knowledge Priors}, which transfer general skills (e.g., mathematical reasoning, spatial reasoning) from large-scale out-of-environment data and can be curated at minimal cost.
The second stage applies online RL to further enhance agents' performance. Specifically, agents are trained with an improved PPO pipeline that incorporates three key designs: efficient context management via self-summarization, dense reward shaping through sub-goal and behavior-shaping rewards, and turn-level policy optimization. Together, these components enable stable and efficient policy learning in long-horizon, sparse-reward settings.

We evaluate ERA on EmbodiedBench \citep{yang2025embodiedbench}, focusing on both high-level planning (EB-ALFRED) and low-level control (EB-Manipulation), which together offer broad coverage of embodied reasoning tasks. ERA-3B not only surpasses prompting-based large models (e.g., GPT-4o) but also outperform 7B-scale training-based baselines, achieving an average score of 65.2\% on EB-ALFRED and 48.3\% on EB-Manipulation. Moreover, our ablation studies disentangle the contributions of different priors in the first stage, as well as context management, reward shaping, and turn-level optimization in the RL stage, providing practical insights for building effective training pipelines for embodied agents.

Our main contributions are threefold: (1) We present a comprehensive study on post-training compact VLMs for embodied agents, integrating prior knowledge curation for supervised fine-tuning with online agentic RL enhanced by key design choices. (2) We introduce a principled taxonomy of accessible prior knowledge for embodied agents, offering practical guidance for data curation across different task levels. (3) We demonstrate strong performance on both high- and low-level embodied tasks using a 3B model, and conduct detailed ablation studies to analyze the contribution of each data source and agentic RL design component.

\section{Related Work}\label{sec:relatedwork}

\newcommand{\greencheck}{\textcolor{green!60!black}{\checkmark}}
\newcommand{\redcross}{\textcolor{red!60!black}{\times}}

\begin{table}[t]
  \centering
  \captionsetup{position=top}
    \caption{Comparison of related work on embodied VLM agents. Most existing studies focus on either high-level or low-level tasks, and many incorporate chain-of-thought (CoT) reasoning. Columns under \textbf{SFT} represent three types of prior knowledge, while those under \textbf{RL} indicate whether the method includes interactive training, process-level rewards, or value learning. Our work presents the most comprehensive framework that unifies SFT and online agentic RL, addressing both high-level and low-level embodied tasks.}
  \setlength{\tabcolsep}{5pt}
  \renewcommand{\arraystretch}{1.15}
  \resizebox{\textwidth}{!}{%
  \begin{tabular}{l l l c c c c c c}
    \toprule
    \multicolumn{3}{c}{} &
    \multicolumn{3}{c}{\textbf{SFT}} &
    \multicolumn{3}{c}{\textbf{RL}} \\
    \cmidrule(lr){4-6} \cmidrule(lr){7-9}
    \textbf{Method} & \textbf{Task Level} & \textbf{Reasoning} & 
    \makecell{Traj.-aug.\\prior} &
    \makecell{Env.-anchored\\prior} &
    \makecell{Ext.-knowledge\\prior} &
    \makecell{In-env.\\interaction} &
    \makecell{Process-level\\reward} &
    \makecell{Value Learning} \\
    \midrule
    Reinforced Reasoning \citep{wu2025reinforced} & High & $\greencheck$ & $\greencheck$ & $\redcross$ & $\redcross$ & $\redcross$ & $\redcross$ & $\redcross$ \\
    CoSo \citep{feng2025towards} & High & $\greencheck$ & $\greencheck$ & $\redcross$ & $\redcross$ & $\greencheck$ & $\redcross$ & $\redcross$ \\
    Embodied-R1 \citep{yuan2025embodiedr1} & Low & $\greencheck$ & - & - & - & $\greencheck$ & $\redcross$ & $\redcross$ \\
    Robot-R1 \citep{kim2025robot} & Low & $\greencheck$ & $\greencheck$ & $\redcross$ & $\redcross$ & - & - & - \\
    GEA \citep{szot2025multimodal} & High \& Low & $\redcross$ & $\greencheck$ & $\redcross$ & $\redcross$ & $\greencheck$ & $\greencheck$ & $\greencheck$ \\
    MolmoAct \citep{lee2025molmoact} & Low & $\greencheck$ & $\greencheck$ & $\greencheck$ & $\greencheck$ & - & - & - \\
    RL4VLM \citep{zhai2024finetuning} & High & $\greencheck$ & $\greencheck$ & $\redcross$ & $\redcross$ & $\greencheck$ & $\redcross$ & $\greencheck$ \\
    RFTF \citep{shu2025rftf} & Low & $\redcross$ & - & - & - & $\greencheck$ & $\greencheck$ & $\greencheck$ \\
    VAGEN \citep{wang2025vagen} & High & $\greencheck$ & - & - & - & $\greencheck$ & $\redcross$ & $\greencheck$ \\
    VLA-RL \citep{lu2025vlarlmasterfulgeneralrobotic} & Low & $\redcross$ & - & - & - & $\greencheck$ & $\greencheck$ & $\greencheck$ \\
    Emma-X \citep{sun2024emmax} & Low & $\greencheck$ &  $\greencheck$ & $\greencheck$ & $\redcross$ & - & - & - \\
    \midrule
    ERA (Ours) & High \& Low & $\greencheck$ & $\greencheck$ & $\greencheck$ & $\greencheck$ & $\greencheck$ & $\greencheck$ & $\greencheck$ \\
    \bottomrule
  \end{tabular}}
\label{tab:related_work}
\end{table}

\textbf{Foundation Model–based Embodied Agents.}
LLMs and VLMs have been explored as embodied agents, enabling them to perceive complex environments and make sequential decisions. Early work relied on prompting strategies to harness the reasoning and planning capabilities of foundation models \citep{singh2022progpromptgeneratingsituatedrobot,song2023llmplanner, hu2023lookleapunveilingpower, kim2024contextawareplanningenvironmentawarememory, shin2025socraticplannerselfqabasedzeroshot}. Building on this foundation, subsequent research introduced mechanisms to improve decision-making, such as code-based tools \citep{liang2023codepolicieslanguagemodel, silver2024generalized}.
More recently, the availability of curated embodied datasets has facilitated supervised finetuning, which has proven effective across both low-level robotic control \citep{zawalski2024roboticcontrol, zhao2025cot, lee2025molmoact, liu2025towards,kim2024openvla, lu2025vla, huang2025tactilevlaunlockingvisionlanguageactionmodels, zhang2025up} and high-level embodied planning \citep{wu2023embodied, chen2024robogptintelligentagentmaking, ji2025robobrainunifiedbrainmodel}.

\textbf{RL for Embodied Agents.} Beyond supervised learning, RL has become a central approach for training embodied agents \citep{su2023subgoal, zhai2024finetuning, yang2024octopus, shu2025rftf, cao2025causal, liu2025whatcan, kim2025robot, szot2025multimodal, feng2025towards}. A key strength of RL lies in its ability to exploit suboptimal and even failed trajectories, thereby making efficient use of diverse data sources \citep{song2024trialanderror, wang2025worldmodelingmakesbetter}. Recent progress further shows that RL can foster reasoning abilities, enabling embodied agents to generalize more effectively to novel tasks \citep{wu2025reinforced, wei2025gtr, lin2025embrace}. Meanwhile, studies reveal that smaller LLMs and VLMs often lack basic embodied knowledge, such as spatial reasoning \citep{gao2024physicallygroundedvisionlanguagemodels, lee2025molmoact, sun2024emmax}. Therefore, grounding embodied knowledge into VLMs prior to RL training has emerged as a promising direction. A detailed comparison of related work is provided in Table \ref{tab:related_work}, with additional discussions deferred to Appendix \ref{app:related_work}.

Table~\ref{tab:related_work} provides a comprehensive comparison
with existing works, highlighting how ERA itself apart from prior works in several aspects.

\section{Problem Formulation}\label{sec:problem_formulation}
\textbf{VLM-based Agents.}
Formally, VLM-based agentic tasks can be modeled as a Partially Observable Markov Decision Process (POMDP) augmented with language, represented by the tuple $(\mathcal{S}, \mathcal{A}, \Omega, \mathcal{T}, \mathcal{O}, L, \mathcal{R})$. Here, $\mathcal{S}$ denotes the full environment state space; $\mathcal{A}$ is the action space; and $\Omega$ is the visual observation space, where each observation $I_t = \mathcal{O}(s_t)$ is generated from the underlying state. The agent also receives a language instruction $L$, which specifies the goal. The reward function $\mathcal{R}$ generally provides a binary signal: 1 if the current state satisfies the instruction, and 0 otherwise.
At timestep $t$, the agent maintains a history $h_t=\left(I_0,a_0,\ldots, I_{t-1}, a_{t-1},I_t\right)$ of past observations and actions, and acts according to a policy $\pi(a_t \mid L, h_t)$ parameterized by a VLM. The episode terminates either when the instruction is satisfied or when a maximum horizon is reached. The learning objective is to maximize the expected reward: $\max_{\pi} \, \mathbb{E}_{\pi}[\sum_{t=0}^\tau \gamma^t r_t]$, where $\tau$ denotes the terminal timestep and $\gamma$ is the discount factor.

\textbf{High-level and Low-level Embodied Tasks.} Following the embodied agent literature \citep{ma2024survey,yang2025embodiedbench}, we distinguish task levels based on the direct executability of actions in robotic systems. \textbf{Low-level tasks} operate on atomic action commands that can be executed directly by robots, typically specifying fine-grained translational or rotational displacements. For example, a robotic arm’s action can be represented as a 7-dimensional vector:
$\label{eq:low_level_action}
    a=[X, Y, Z, \rm{Roll}, \rm{Pitch}, \rm{Yaw}, \rm{Gripper}]$, 
where $(X, Y, Z)$ denotes incremental translations, $(\rm{Roll}, \rm{Pitch}, \rm{Yaw})$ represent rotational deltas in Euler angles, and \(\rm{Gripper}\) encodes the binary open/closed state of the end-effector. In contrast, \textbf{high-level tasks} consist of semantically meaningful actions that can be decomposed into sequences of low-level primitives. For instance, in EB-ALFRED \citep{yang2025embodiedbench}, a high-level action such as``find a HandTowel" may require iterating through multiple low-level behaviors, which are ultimately grounded into executable commands within the simulator. \emph{Prior work has revealed that high-level tasks primarily emphasize logical reasoning and long-term planning, whereas low-level tasks focus on precise perception and manipulation.}



\section{From Priors to Policies: Training VLMs as Embodied Agents}\label{sec:method}

In this section, we present the Embodied Reasoning Agent (ERA) framework, a two-stage training pipeline designed to equip compact VLMs with strong embodied intelligence. ERA consists of: (1) \textbf{Embodied Prior Learning}, which injects structured perception and reasoning capabilities into small VLMs through supervised-finetuning on curated prior data, and (2) \textbf{Online Reinforcement Learning}, which further enhances embodied performance via interactive online training with process-level rewards and turn-level advantage estimation.



\subsection{Embodied Prior Learning}
\label{sec:prior}

Because of the inherent gap between VLM pretraining and embodied agent tasks, small VLMs cannot be directly applied in this domain. Bridging this gap requires learning embodied priors before engaging in environment interactions.
A common approach to adapt VLMs for embodied tasks is to fine-tune them on task-specific trajectories \citep{liu2024visualagentbench, yuan2025embodiedr1, wu2025reinforced, feng2025towards}. However, this strategy faces two key limitations. First, \textit{data scarcity and cost}: collecting high-quality agent trajectories is computationally expensive and time-consuming. Second, \textit{limited reasoning supervision}: popular datasets such as ALFRED typically record only action sequences without detailed reasoning traces, since annotating such information is difficult and costly. We refer to trajectory data with little or no reasoning supervision as \emph{raw trajectory} data.

\begin{figure}[th!]
    \centering
    \includegraphics[width=1\linewidth]{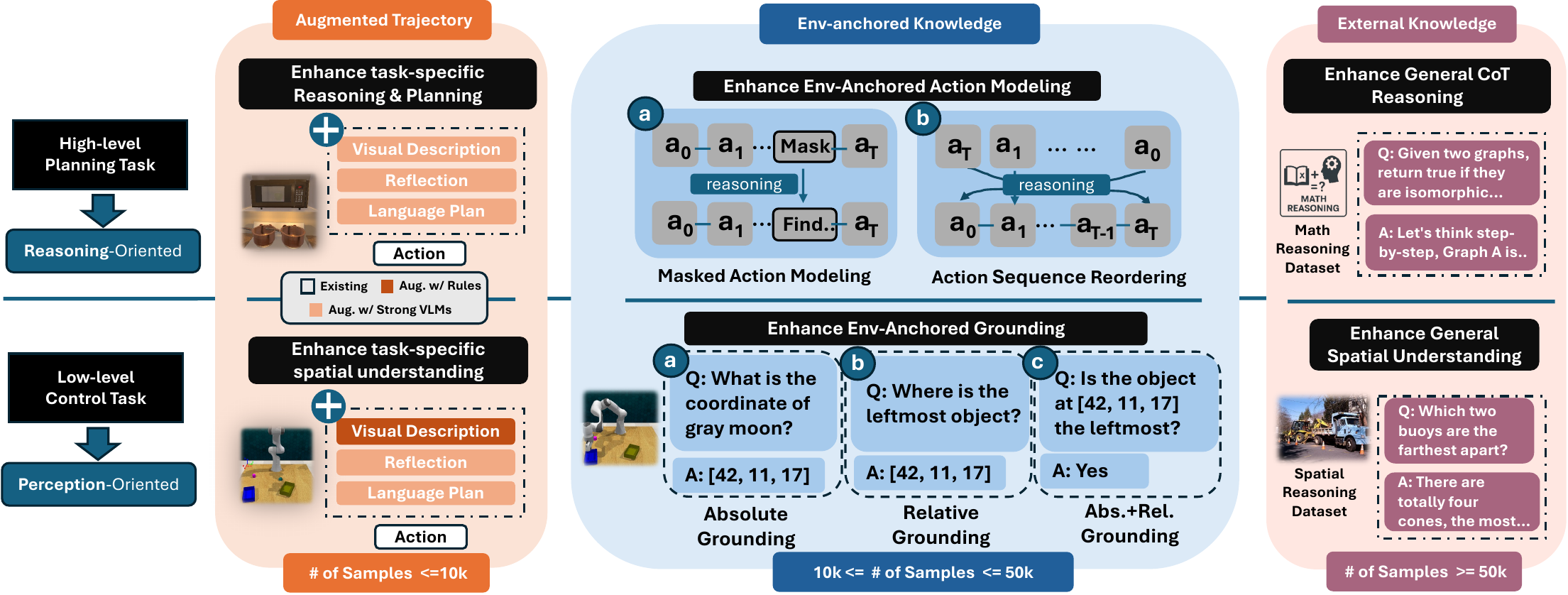}
    \vspace{-10pt}
    \caption{Illustration of Embodied Prior Learning (EPL). EPL leverages three data sources: Augmented trajectory priors, environment-anchored priors, and external knowledge priors.}
    \label{fig:embodied_prior_learning}
\end{figure}

To overcome these challenges, we propose to curate embodied prior data from diverse sources rather than relying solely on raw trajectories. We introduce three different prior data sources: \textbf{trajectory-augmented priors}, which enrich existing trajectories with reasoning annotations and detailed visual descriptions; \textbf{environment-anchored priors}, which provide contextual supervision and grounding signals from the same environment beyond the agent’s recorded trajectories; and \textbf{external knowledge priors}, which transfer general reasoning and perception skills from out-of-domain data. 
\textbf{An overview of these datasets is illustrated in Figure~\ref{fig:embodied_prior_learning}, and their statistics are summarized in Table~\ref{tab:dataset_stats}.}

Given a curated dataset or their combination \(\mathcal{D_{\text{EPL}}}=\{(\mathbf{x}_i,\mathbf{y}_i)\}_{i=1}^N\), where \(\mathbf{x}_i\) is a prompt and
\(\mathbf{y}_i=(\mathbf{y}_{i,1},\dots,\mathbf{y}_{i,|\mathbf{y}_i|})\) is the corresponding response, we perform supervised fine-tuning to inject embodied knowledge into the base model:
\[
\mathcal{L}_{\text{EPL}}(\theta)
= -\frac{1}{N}\sum_{i=1}^N \sum_{j=1}^{|\mathbf{y}_i|} 
\log \pi_\theta(\mathbf{y}_{i,j}\mid \mathbf{x}_i,\mathbf{y}_{i,<j}),
\]
where $\mathbf{y}_{i,<j}$ denotes the sequence of tokens preceding the $j-$th token in the response.
For downstream deployment, training on trajectory data, either raw or augmented, is indispensable. Thus, instead of mixing the dataset together, we first fine-tune VLMs on environment-anchored or external knowledge priors to acquire general embodied skills, and subsequently train them on raw or augmented trajectories to obtain the final agent model.

\begin{table*}[h]
\centering
\small
\caption{Statistics of the embodied prior learning datasets used for EB-ALFRED and EB-Manipulation. We report the number of trajectories (\#Traj), total samples (\#Samples), and average output token length (Avg. Token Len.) for each prior dataset type: Raw trajectory, Trajectory-Augmented Prior (Traj-Aug), Environment-Anchored Prior (Env-Anc), and External Knowledge Prior (Ext-Know).}
\begin{tabular}{lllccc}
\toprule
\textbf{Domain} & \textbf{Prior} & \textbf{Dataset} & \textbf{\#Traj} & \textbf{\#Samples} & \textbf{Avg. Token Len.} \\
\midrule

\multirow{4}{*}{EB-ALFRED} 
& Raw & -- & 913 & 8,834 & 74.0 \\
& Traj-Aug & -- & 913 & 8,834 & 159.8 \\
& Env-Anc & Masked Action Modeling & -- & 41,616 & 396.1 \\
& Env-Anc & Action Seq. Reordering & -- & 6,574 & 395.6 \\
& Ext-Know & OpenO1-SFT \citep{openo1_2024b} & -- & 10,000 & 1102.6 \\

\midrule

\multirow{5}{*}{EB-Manipulation} 
& Raw & -- & 622 & 4,249 & 104.4 \\
& Traj-Aug & -- & 622 & 4,023 & 284.2 \\
& Env-Anc & Absolute Grounding & -- & 4,444 & 7.6 \\
& Env-Anc & Relative Grounding & -- & 2,000 & 12.0 \\
& Env-Anc & Comb. Grounding & -- & 8,888 & 1.0 \\
& Ext-Know & SpaceThinker \citep{RemyxAI_2025_SpaceThinker_Dataset} & -- & 11,413 & 202.9 \\

\bottomrule
\end{tabular}

\label{tab:dataset_stats}
\end{table*}

\subsubsection{Trajectory-Augmented Priors}
\label{sec:traj}
Existing embodied trajectories typically consist only of observations and action sequences, without the intermediate reasoning steps that are crucial for solving complex agentic tasks. Some recent efforts \citep{yang2025embodiedbench} provide limited reasoning signals, such as a single high-level rationale corresponding to a multi-step plan. However, such coarse supervision does not offer structured guidance at intermediate steps, making it difficult for agents to adapt quickly when errors occur partway through execution.  


To address this limitation, we construct trajectory-augmented priors by enriching every step of the trajectory with explicit reasoning supervision from large VLMs such as GPT-4o. Specifically, for each timestep $t$, we define a structured reasoning trace $z_t = \{z^{vis}_t, z^{ref}_t, z^{plan}_t\}$, where $z^{vis}_t$ is a \textit{visual description} of the current state, $z^{ref}_t$ is a \textit{reflection} on the history to detect and correct potential errors, and $z^{plan}_t$ is a \textit{step-level plan} for achieving the task. By prompting GPT-4o with the language instruction $L$, current observation $I_t$, action history $\{a_0,\ldots,a_{t-1}\}$, and the current action $a_t$, we obtain $z_t$ that enriches the trajectory with structured ``inner monologue" \citep{huang2022inner,xu2024aguvis}. This step-level reasoning has been shown to significantly improve generalization in high-level embodied tasks \citep{feng2025towards, zhai2024finetuning}. Prompting used for generating this reasoning with GPT-4o is provided in Appendix~\ref{app:4o_prompt}.

For low-level manipulation tasks, however, GPT-4o often produces inaccurate visual descriptions, leading to misalignment between perception and action, and consequently, failure to solve low-level tasks. To overcome this issue, we employ a rule-based method that generates ground-truth visual descriptions using the backend information of the underlying simulator, ensuring consistency between perception and control. Further details and illustrative examples of this rule-based procedure are provided in Appendix~\ref{app:rule_based}. 

\subsubsection{Environment-Anchored Priors}
\label{sec:env}
While agent trajectories provide the most direct supervision for embodied tasks, such data is often limited in scale and expensive to collect.
This limitation raises a natural question: \textbf{can we leverage additional supervision that is not trajectory-based, yet still grounded in the same environment, to enhance the agent’s learning?} To address this, we introduce \textit{environment-anchored priors}: auxiliary data sources that encode environment-level knowledge beyond trajectories, such as semantic question answering, visual grounding, or spatial reasoning. These priors enrich agents with a deeper understanding of task semantics, temporal dependencies, visual grounding, and spatial relations. We illustrate these datasets in the middle of Figure \ref{fig:embodied_prior_learning}.

\textbf{For EB-ALFRED, we curate two semantic supervision datasets designed to enhance task comprehension, action feasibility, and embodied planning}: \textit{masked action modeling} and \textit{action sequence reordering}. Both datasets are derived from the original ALFRED training set, which contains large-scale raw action sequences. Because the original actions do not perfectly align with the EB-ALFRED action space, we apply rule-based matching and replacement to adapt them for effective use. Detailed processing steps are provided in Appendix~\ref{app:rule_based_alfred}.
\begin{itemize}[leftmargin=1em, itemsep=0pt, topsep=0pt]
    \item \textbf{Masked Action Modeling.}  
    Given an instruction $L$ and an action sequence $\{a_0, a_1, \ldots, a_T\}$, we mask a randomly selected timestep $t \in \{0, \dots, T\}$, replacing $a_t$ with $[\text{MASK}]$. This produces a query–output pair:  
    \[
    q = (L, \{a_0, \ldots, a_{t-1}, [\text{MASK}], a_{t+1}, \ldots, a_T\}), 
    \quad y = (z, a_t),
    \]
    where $q$ is the masked trajectory with its instruction, and $y$ contains the missing action $a_t$ along with a reasoning trace $z$. The reasoning traces, generated by GPT-4o, explain why $a_t$ is correct, providing explicit supervision that strengthens both prediction accuracy and interpretability.

    \item \textbf{Action Sequence Reordering.}  
    Here, an action sequence $\{a_0, a_1, \ldots, a_T\}$ is randomly shuffled into a permuted sequence $\{a_{m_0}, a_{m_1}, \ldots, a_{m_T}\}$. The query–output pairs are organized as:  
    \[
    q = (L, \{a_{m_0}, a_{m_1}, \ldots, a_{m_T}\}), 
    \quad y = (z, \{a_0, a_1, \ldots, a_T\}),
    \]  
    where $q$ contains the permuted sequence and instruction, and $y$ provides the correctly ordered sequence with a reasoning trace $z$ from GPT-4o. These traces explain the correct temporal order, helping the model capture action dependencies and temporal coherence.
\end{itemize}

\textbf{For EB-Manipulation, we curate environment-anchored priors that emphasize spatial understanding}, a critical component towards fine-grained control. Using simulated episodes from the VLMbench training set \citep{zheng2022vlmbench}, we combine image observations with ground-truth 3D coordinates to construct three complementary subsets based on the question type: \textit{absolute coordinate grounding}, \textit{relative coordinate grounding}, and \textit{combined grounding}. Detailed dataset descriptions are provided in Appendix~\ref{app:app_env_anchored}.
\begin{itemize}[leftmargin=1em, itemsep=0pt, topsep=0pt]
    \item \textbf{Absolute Coordinate Grounding.}  
    This dataset builds direct mappings between objects and their 3D coordinates. Two query formats are included: (1) given a scene image (with a robot arm and various objects on a desk) and an object description, predict its coordinates; or (2) given a scene and coordinate, describe the object at that location.
    \item \textbf{Relative Coordinate Grounding.}  
This subset captures spatial relations such as ``the leftmost" and ``the second leftmost". Each query pairs an image with a relational description, and the output is the coordinate of the corresponding object, allowing the agent to learn spatial reasoning and relational grounding.
    \item \textbf{Combined Grounding.}  
This subset combines both absolute and relative grounding information within each query to enable joint reasoning. Each data sample pairs an observation with a coordinate and a spatial relation (e.g., ``Is the object at [42, 11, 17] the leftmost?"). The response is binary (``Yes" or ``No"), allowing models to learn both coordinate grounding and relational verification.  
\end{itemize}

\subsubsection{External Knowledge Priors}
\label{sec:ext}

While environment-anchored priors enrich embodied agents with knowledge tied to their task environments, they remain limited in scale compared to the massive datasets commonly used in general LLM/VLM training. This raises a natural question: \textbf{can broader datasets, curated outside embodied environments, provide complementary supervision for embodied agent learning?} To explore this, we define \textit{external knowledge priors}: large-scale datasets that potentially transfer general reasoning and cross-domain grounding capabilities from out-of-environment data, thereby enabling agents to generalize beyond what environment-specific data alone can provide.

For \textbf{high-level planning tasks} such as EB-ALFRED, we investigate whether external reasoning datasets can strengthen agents' planning and logical reasoning abilities. Specifically, we adopt the \textit{OpenO1-SFT dataset} \citep{openo1_2024b}, a text-based corpus designed to activate chain-of-thought reasoning. OpenO1-SFT is multilingual and contains 77,685 examples. From its English subset, we sample 10,000 QA pairs to construct our external knowledge prior dataset.

For \textbf{low-level control tasks} such as EB-Manipulation, we examine whether external spatial reasoning datasets can improve agents’ visual perception and spatial understanding. To this end, we utilize the \textit{SpaceThinker dataset} \citep{RemyxAI_2025_SpaceThinker_Dataset}, a synthesized multimodal spatial reasoning corpus. We use all 11,413 QA pairs from this dataset to curate the corresponding external knowledge prior.

By incorporating these external knowledge priors, we complement environment-anchored supervision with large-scale reasoning and grounding signals. This combination equips VLMs with richer generalization capabilities that extend beyond the limitations of purely embodied data.


\subsection{Online Reinforcement Learning}
\label{sec:agent_rl}


While Embodied Prior Learning (\S\ref{sec:prior}) equips agents with essential foundational skills, the dynamic nature of embodied tasks requires going beyond static priors to achieve adaptive planning, reasoning, and reflection. The natural next stage is online reinforcement learning (RL), where agents interact with the environment through trial and error, gradually discovering how to best leverage and refine their priors for autonomous task completion. However, applying RL in embodied agents introduces three major challenges:  
(1) \textbf{Efficient VLM-based agent design}: due to the high computational cost of RL training, it is crucial to build an efficient agent framework that manages training overhead through careful design choices, such as context management.  
(2) \textbf{Long-horizon tasks with sparse rewards}: tasks often require many steps (e.g., 20+) to complete, making it difficult for RL to propagate meaningful supervision signals, even when prior knowledge is available.  
(3) \textbf{Stable RL optimization}: conventional RL methods for LLMs/VLMs tend to be unstable in embodied agent settings, since token-level value functions capture limited semantic meaning and are not suitable for effective agent policy optimization.  

To address these challenges, we introduce our online RL training pipeline with three key components:  
(1) an efficient agent framework that incorporates structured observation description, reflection, planning, and self-summarization context management; (2) a process-level reward function that provides fine-grained supervision at each step; and (3) a turn-level advantage value estimation for PPO, which aligns better with agent-level decision-making and delivers more stable training compared to existing token-level or bi-level methods.

\subsubsection{Agent Framework}
\label{sec:framework}

\begin{figure}[t!]
     \centering
     \includegraphics[width=1\linewidth]{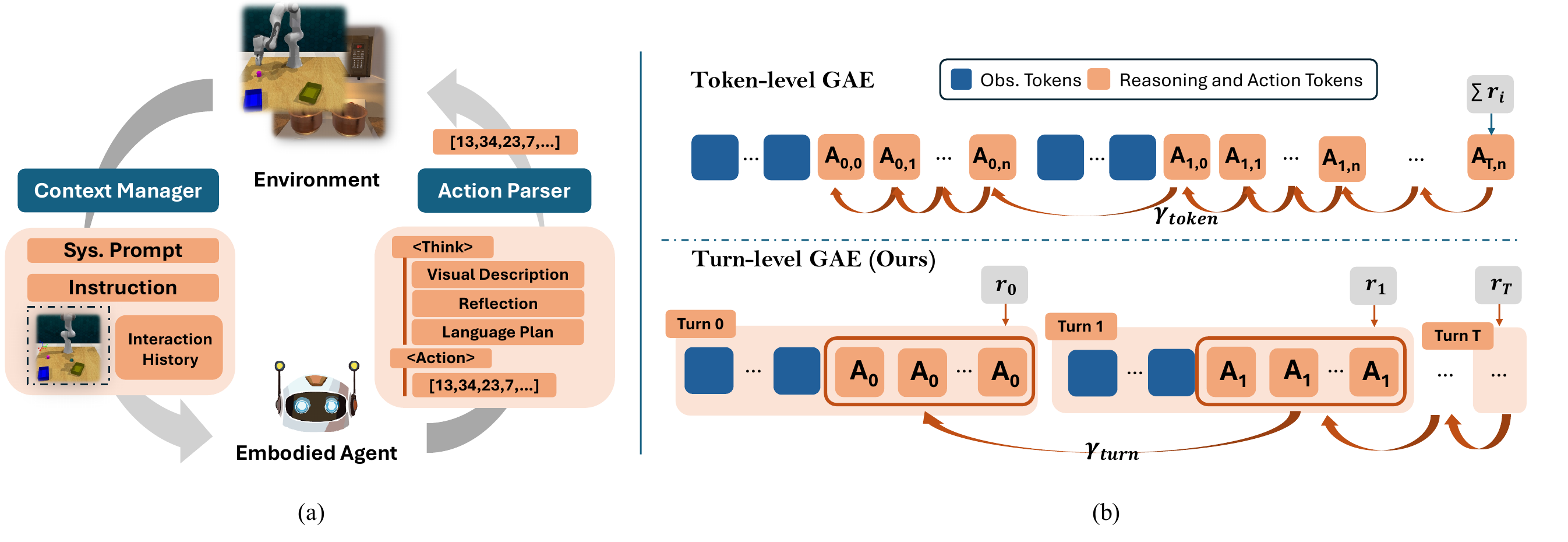}
     \vspace{-20pt}
     \caption{(a) Our agent framework, and (b) a comparison of advantage estimation in turn-level GAE and token-level GAE. $A_{0,0}, \ldots, A_{T,n}$, and $A_0, \ldots,A_T$ refer to Advantage values for each response token.}
     \label{fig:agent_framework_and_gae}
 \end{figure}

We design a unified VLM-based embodied agent pipeline, illustrated in Figure~\ref{fig:agent_framework_and_gae}(a). 
This pipeline provides a powerful framework for processing multimodal inputs, generating structured reasoning and reflection, and producing executable actions. 
The agent processes three types of inputs: language instructions, visual observations, and interaction history. For visual perception, the agent takes the current step image as input, following prior work's agent design \citep{yang2025embodiedbench,xu2024aguvis}. 
Interaction history is maintained by a context manager that integrates environment feedback with the agent’s previous reasoning traces and actions.
Below, we describe the key components of our agent framework.

\textbf{Structured Reasoning and Action.}
At each turn, the agent generates a response that combines a reasoning trace with an executable action, separated by special tokens:  
\texttt{<|think\_start|>}~…~\texttt{<|think\_end|>} for reasoning, and \texttt{<|action\_start|>}~…~\texttt{<|action\_end|>} for actions. This ReAct-style \citep{yao2023react} format leverages the reasoning ability of VLMs to improve decision-making, while also ensuring that actions can be directly parsed by downstream modules.  
The reasoning trace is further structured into three components, $
z_{t} = \{z_t^\text{vis}, z_t^\text{ref}, z_t^\text{plan}\}$, 
where $z_t^\text{vis}$ is a natural language description of the current visual input, $z_t^\text{ref}$ is a reflection on past actions, environment feedback, and progress toward the goal, and $z_t^\text{plan}$ is a multi-step plan that guides subsequent actions.  
The action enclosed between action tokens is formatted according to environment requirements: either as an action string for high-level planning tasks (EB-ALFRED) or as a 7-dimensional vector for low-level control (EB-Manipulation).  
This structured design elicits the reasoning capability of foundation models while enabling fine-grained optimization during agent training.  

\textbf{Self-Summarization Context Management.}
The context manager organizes historical information and feeds it into the model to ensure continuity of reasoning. A key challenge in long-horizon tasks is the \textit{context explosion problem}: naively retaining the full history of agent outputs (reasoning and action) and environment feedback, $h_t = (z_{1}, a_1, e_1, \dots, z_{t-1}, a_{t-1},e_{t-1})$, causes input length to grow linearly with turn number $t$, i.e., $\mathcal{O}(t)$.  
This is computationally inefficient and may harm performance by diverting attention to irrelevant history.
Sliding-window approaches are a common workaround, but their window sizes are often chosen heuristically rather than principled. In our framework, trajectory-augmented priors train the model to explicitly summarize the interaction history into a reflection at each step. 
This design allows the agent to compress the entire past into its most recent thought $z_{t-1}$ and we only need a one-step context $h_t = (z_{t-1}, a_{t-1}, e_{t-1})$, effectively reducing context size to $\mathcal{O}(1)$ while retaining essential information.  
We refer to this lightweight mechanism as \textit{Self-Summarization}. It enables efficient long-horizon reasoning by maintaining crucial context without being hindered by non-critical historical details.

\subsubsection{Reward Design}
\label{sec:reward_design}

Embodied tasks are typically long-horizon (e.g., 20 steps) and often suffer from sparse supervision, where rewards are only given upon task success. To provide richer learning signals, we design a process-level reward function $r_t$ that integrates task completion, intermediate progress, and behavior shaping. At each turn $t$, the reward is defined as
\[
r_t = r_t^{\text{success}} + r_t^{\text{subgoal}} + r_t^{\text{behavior}},
\]
with the three components described below and further detailed in Appendix~\ref{app:reward_design}.

\textbf{Success-based Reward ($r_t^{\text{success}}$).}
This sparse reward is assigned a positive value when the task is successfully completed before reaching the maximum step limit, and zero otherwise.

\textbf{Subgoal-based Reward ($r_t^{\text{subgoal}}$).}  
To provide denser feedback for reinforcement learning, subgoal rewards are granted the first time the agent achieves rule-defined subgoals. For \emph{high-level planning} tasks, subgoals correspond to satisfying one of the conditions specified in the Planning Domain Definition Language (PDDL) configuration of the simulator.
For \emph{low-level manipulation} tasks, subgoals are defined as the first successful approach of the end-effector to the instruction-referenced object within a predefined distance threshold.

\textbf{Behavior-Shaping Reward ($r_t^{\text{behavior}}$).}
This component shapes task-specific behaviors by rewarding desirable actions and penalizing undesirable ones. For \emph{high-level planning}, penalties are applied to invalid actions that the environment cannot execute (e.g., attempting to pick up an object while already holding another, or issuing an erroneous action string). For \emph{low-level manipulation}, rewards are based on the accuracy of the agent’s visual grounding, quantified by the ratio of correctly matched attributes against the ground truth. Thresholds on this ratio determine positive or negative rewards. Full implementation details are provided in Appendix~\ref{app:reward_design}.

\subsubsection{Turn-level Policy Optimization}
\label{sec:turnwise_rl_main}

Conventional token-level optimization, widely adopted in RLHF and recent agent RL works \citep{wu2025reinforced,kim2025robot}, is not well-suited for multi-turn embodied agents. In embodied tasks, interactions and rewards are inherently defined at the \textit{turn level}. Learning a value function for individual reasoning or action tokens is therefore less meaningful and often leads to unstable policy optimization.

To address this challenge, we propose a turn-level policy optimization scheme, where the agent’s entire response in a turn is treated as a single ``action." At each turn $t$, rather than estimating values for every token, we pass only the state input $\mathbf{x}_t$ (including current observation, task instruction, and interaction history) to the value function $V_\phi$ to obtain a single estimate $V_\phi(\mathbf{x}_t)$. Given turn-level rewards $\{r_t\}_{t=0}^{T}$ for a trajectory of $T+1$ turns, we compute the temporal-difference (TD) residual for each turn: $
\delta_t = r_t + \gamma V_\phi(\mathbf{x}_{t+1}) - V_\phi(\mathbf{x}_t) $,
with terminal bootstrap $V_\phi(\mathbf{x}_{T+1})=0$. The turn-level generalized advantage estimate (GAE) is then calculated as:
\begin{equation}
A_t = \sum_{l=0}^{T-t} (\gamma\lambda)^l \, \delta_{t+l},
\end{equation}
where $\lambda$ controls the bias–variance trade-off. This advantage estimate $A_t$ is \textbf{shared across all tokens within the response} $\mathbf{y}_t$, ensuring that credit assignment aligns with the natural unit of environment interaction. Figure~\ref{fig:agent_framework_and_gae}(b) illustrates the difference between turn-level and token-level value estimation.

We perform parallel rollouts with multiple environments and collect an online buffer $\mathcal{D}$ of turn-level state–response pairs for PPO updates. The policy objective is defined as
\begin{equation}
\label{eq:turnwise_policy_loss_main}
\mathcal{L}_{\text{policy}}(\theta) 
= \mathbb{E}_{(\mathbf{x}_t, \mathbf{y}_t) \sim \mathcal{D}} \left[ 
\frac{1}{|\mathbf{y}_t|} \sum_{i=1}^{|\mathbf{y}_t|}
\min\!\left( 
\frac{\pi_\theta(\mathbf{y}_{t,i}\,|\,\mathbf{x}_t)}{\pi_{\theta_{\text{old}}}(\mathbf{y}_{t,i}\,|\,\mathbf{x}_t)} A_t,\;
\text{clip}\!\left(\frac{\pi_\theta(\mathbf{y}_{t,i}\,|\,\mathbf{x}_t)}{\pi_{\theta_{\text{old}}}(\mathbf{y}_{t,i}\,|\,\mathbf{x}_t)},\,1-\epsilon,\,1+\epsilon\right) A_t
\right) \right],
\end{equation}
where $\pi_{\theta}$ and $\pi_{\theta_{\text{old}}}$ are the current and old policies, $\mathbf{y}_{t,i}$ is the $i$-th token in response $\mathbf{y}_t$, and $\epsilon$ is the clipping threshold. Note that $A_t$ is shared by all tokens of $\mathbf{y}_t$. The value function is trained concurrently by regressing toward a detached target:
\begin{equation}
\label{eq:value_loss_main}
\mathcal{L}_{\text{value}}(\phi) 
= \mathbb{E}_{\mathbf{x}_t \sim \mathcal{D}} \left[ 
\frac{1}{2}\,\big( V_\phi(\mathbf{x}_t) - \text{no\_grad} \left( A_t + V_\phi(\mathbf{x}_t) \right) \big)^2 \right].
\end{equation}
Overall, this turn-wise formulation reduces variance in advantage estimation and leads to more stable policy learning for embodied agents. An empirical comparison between token-level and turn-level GAE is presented in Section~\ref{exp:rl_reward_gae_ablations}.

\section{Experiments}
\label{sec:exp}

\begin{figure}[h!]
    \centering
    \includegraphics[width=1\textwidth,height=0.92\textheight]{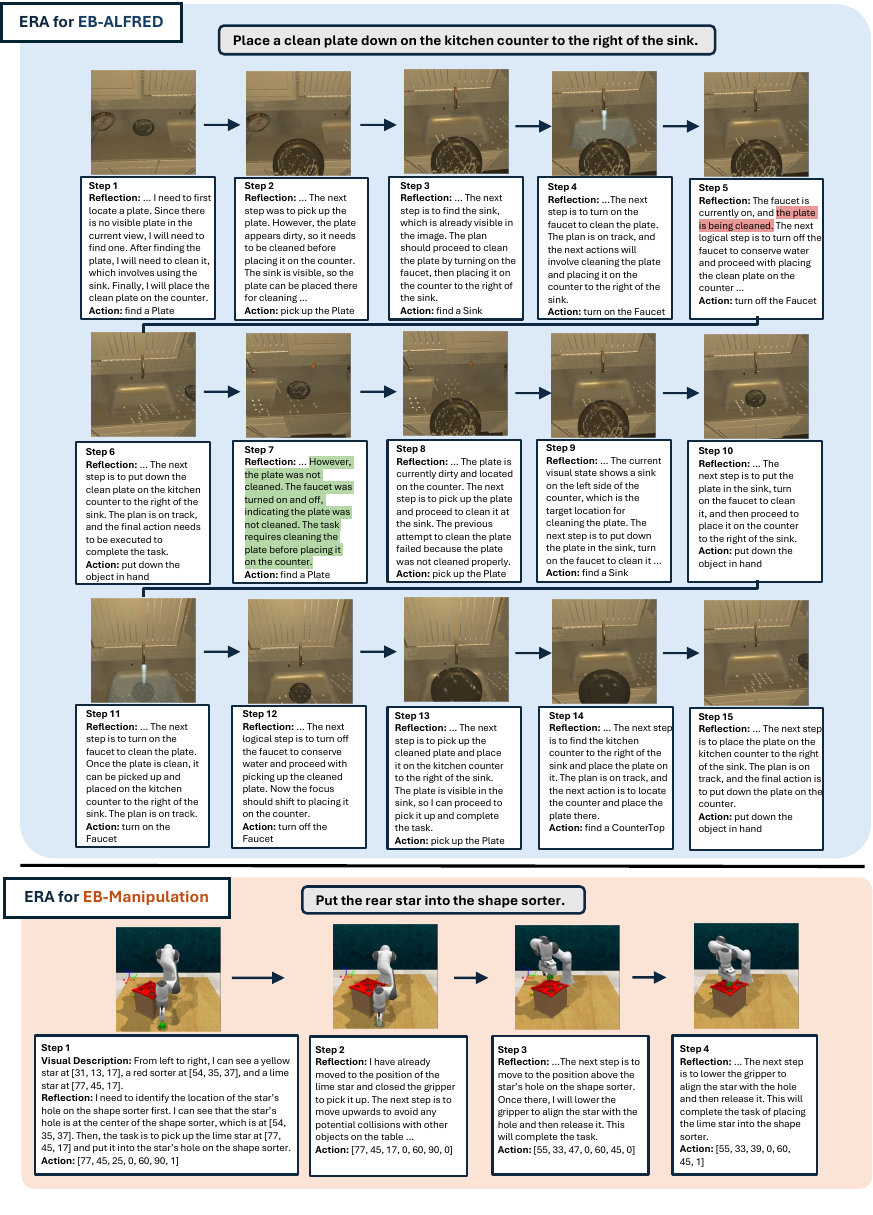}
    \caption{Examples of ERA performing step-by-step reasoning and actions: (a) on EB-ALFRED, it identifies and reflects on earlier errors; (b) on EB-Manipulation, it accurately places the star into the correct slot of the shape sorter.}
    \label{fig:successful_examples}
\end{figure}

We conduct comprehensive experiments on both high-level planning and low-level manipulation tasks, aiming to gain deeper insights into how different design choices contribute to embodied agent learning. Specifically, we address the following research questions:

\noindent
{\textcircled{\scriptsize 1}}~\textbf{Q1}: What performance does ERA achieve compared to strong baselines? \\
{\textcircled{\scriptsize 2}}~\textbf{Q2}: What role do different prior datasets play in agent performance? \\
{\textcircled{\scriptsize 3}}~\textbf{Q3}: Is Self-Summarization effective for context management? \quad \\
{\textcircled{\scriptsize 2}}~\textbf{Q4}: How do reward design and turn-level value impact RL performance? \quad

\textbf{Experiment Setup.}
We build ERA on top of Qwen2.5-VL-3B-Instruct and evaluate models on \emph{EmbodiedBench}, a comprehensive benchmark covering both high-level planning (EB-ALFRED) and low-level control (EB-Manipulation). The benchmark includes diverse subsets designed to assess distinct embodied capabilities. To evaluate both in-distribution learning and out-of-distribution generalization, we use three subsets, i.e., base skills (``Base"), complex instruction following (``Complex"), and visual perception (``Visual"), for \textbf{\emph{training (seen)}}, and hold out commonsense reasoning (``Common") and spatial awareness (``Spatial") subsets for \textbf{\emph{testing (unseen)}} ERA and other training-based baselines. We report task success rate as the primary evaluation metric. Additional details on dataset curation, training hyperparameters, and evaluation procedures are provided in Appendix~\ref{app:additional_experiments}.


\begin{table*}[t!]
    \centering\small
    \caption{
    Task success rates on the five subsets of EB-ALFRED and EB-Manipulation. The best result in each column is highlighted in \textbf{bold}. ``Base,’’ ``Complex,’’ and ``Visual’’ are seen subsets, while ``Common’’ and ``Spatial’’ are unseen subsets.
    }
    \renewcommand{\arraystretch}{1.1}
    \setlength\tabcolsep{2pt}
    \setlength\extrarowheight{2pt}
    \resizebox{1\linewidth}{!}{

    \begin{tabular}{
        >{\centering\arraybackslash}p{3.3cm} 
        >{\centering\arraybackslash}p{1.17cm} 
        >{\centering\arraybackslash}p{1.17cm} 
        >{\centering\arraybackslash}p{1.17cm} 
        >{\centering\arraybackslash}p{1.17cm} 
        >{\centering\arraybackslash}p{1.17cm} 
        >{\centering\arraybackslash}p{1.17cm} 
        @{\hskip 10pt} 
        >{\centering\arraybackslash}p{1.17cm} 
        >{\centering\arraybackslash}p{1.17cm} 
        >{\centering\arraybackslash}p{1.17cm} 
        >{\centering\arraybackslash}p{1.17cm} 
        >{\centering\arraybackslash}p{1.17cm} 
        >{\centering\arraybackslash}p{1.17cm} 
    }
    
        \toprule
        
        \multirow{2}{*}{\textbf{ Model}} & \multicolumn{6}{c}{\bf EB-Alfred} & \multicolumn{6}{c}{ \bf EB-Manipulation} \\

        \cmidrule(lr){2-7} \cmidrule(lr){8-13}
        
        ~ & \textbf{Avg} & \cellml \textbf{Base} & \cellml \textbf{Complex} & \cellml \textbf{Visual} & \cellmr \textbf{Common}  & \cellmr \textbf{Spatial} 
        & \textbf{Avg} & \cellml \textbf{Base} & \cellml \textbf{Complex} & \cellml \textbf{Visual} & \cellmr \textbf{Common}  & \cellmr \textbf{Spatial}   \\

        \addlinespace[2pt]
        \midrule
        \addlinespace[2pt]

        
        \multicolumn{13}{c}{ \textit{Prompting-based MLLMs} }  \\ 
        \midrule
        
        {\fontsize{8}{10}\selectfont GPT-4o} & 56.8 & 64 & 68 & 46 & 54 & 52
        &  28.9 & 39.6 & 29.2 & 19.4 & 29.2 & 25.0 \\ 

        {\fontsize{8}{10}\selectfont Claude-3.5-Sonnet}&  \textbf{66.4} & \textbf{72} & \textbf{76} & 60 & \textbf{66} & 58 
        &  25.4 & 37.5 & 29.2 & 19.4 & 16.7 & 22.9 \\

        {\fontsize{8}{10}\selectfont Gemini-1.5-Pro} &  63.2 & 70 & 72 & 58 & 64 & 52
        &  21.1 & 14.6 & 22.9 & 16.7 & 14.6 & 35.4 \\
        
        {\fontsize{8}{10}\selectfont Gemini-2.0-flash} & 51.2 & 62 & 54 & 46 & 48 & 46
        &  16.7 & 14.6 & 14.6 & 13.9 & 8.3 & 31.3\\

         {\fontsize{8}{10}\selectfont Llama-3.2-90B-Vision} &  35.2 & 38 & 44 & 28 & 34 & 32 
        & 14.9 & 10.4 & 16.7 & 10.4 & 12.5 & 20.8 \\

        {\fontsize{8}{10}\selectfont InternVL3-78B} & 39.6 & 38 & 46 & 42 & 34 & 38 
        & 26.3 & 29.2 & 22.9 & 25.0 & 22.9 & 31.3 \\

        {\fontsize{8}{10}\selectfont Qwen2.5-VL-72B-Ins} & 40.8 & 50 & 42 & 36 & 42 & 34
        & 16.2 & 12.5 & 16.7 & 22.2 & 12.5 & 18.8 \\

        {\fontsize{8}{10}\selectfont Qwen2.5-VL-7B-Ins} & 5.2 & 10 & 6 & 2 & 8 & 0 
        & 9.6 & 8.3 & 8.3 & 5.6 & 8.3 & 16.7 \\

        {\fontsize{8}{10}\selectfont Qwen2.5-VL-3B-Ins} & 0 & 0 & 0 & 0 & 0 & 0 
        & 0 & 0 & 0 & 0 & 0 & 0 \\

        \addlinespace[2pt]
        \midrule
        \addlinespace[2pt]

        \multicolumn{13}{c}{ \textit{Training-based MLLMs} }   \\
        \midrule

        {\fontsize{8}{10}\selectfont RL4VLM (3B)} & 51.2  & 70 & 70 & 56 & 32 & 28 
        & 21.9 & 33.3 & 29.2 & 30.6 & 8.3 & 8.3 \\

        {\fontsize{8}{10}\selectfont VAGEN (3B)} &  52.8 & 70 & 70 & 58 & 38 & 28 
        & 22.9 & 35.4 & 31.3 & 29.2 & 8.3 & 10.4 \\

        {\fontsize{8}{10}\selectfont Reinforced Reasoner (7B)} & 41.6   &   54 & 46 & 28 & 42 & 38 
        &  - & - & - & - & - & - \\

        {\fontsize{8}{10}\selectfont Robot-R1 (7B)} &  - & - & - & - & - & -
        & 11.7  &   12.5    &   6.3    &   16.7   &   8.3   &  14.6   \\

        \addlinespace[2pt] 
        \midrule
        \addlinespace[2pt]


        \rowcolor{gray!20} {\fontsize{8}{10}\selectfont \textbf{ERA-3B (EPL-only)} } & 56.0 & 68  & 66 & 52 & 44 & 50
        & 40.0 & 45.8 & 41.7 & 47.9 & 37.5 & 27.1 \\

        \rowcolor{gray!20} {\fontsize{8}{10}\selectfont \textbf{ERA-3B (EPL+RL)} } & 65.2 & \textbf{72}  & 72 & \textbf{62} & 54 & \textbf{66}
        & \textbf{48.3} & \textbf{56.3} & \textbf{47.9} & \textbf{50.0} & \textbf{47.9} & \textbf{39.6} \\

        \bottomrule
    \end{tabular}\label{tb:overall_table}
    }
\end{table*}

\subsection{\textbf{Q1:} What performance does ERA achieve compared to strong baselines?}

In Table~\ref{tb:overall_table}, we compare ERA-3B against a diverse set of baselines. These include \emph{prompting-based models}, such as GPT-4o, Claude-3.5-Sonnet, Gemini-1.5-Pro, Gemini-2.0-Flash, Llama-3.2-90B-Vision-Ins \citep{patterson2022llama90v}, InternVL3-78B \citep{zhu2025internvl3}, and Qwen2.5-VL (3B, 7B, and 72B) \citep{bai2025qwen25vl}. We also evaluate \emph{RL-based models}, including reproduced RL4VLM \citep{zhai2024finetuning} and VAGEN \citep{wang2025vagen}, which are trained from our best EPL checkpoint and fine-tuned with identical setting as ERA, ensuring a fair comparison of the RL stage. In addition, we report results for Reinforced Reasoner \citep{wu2025reinforced} on EB-ALFRED and Robot-R1 \citep{kim2025robot} on EB-Manipulation, where scores are taken from their original papers. 

We also present two trajectory examples in Figure~\ref{fig:successful_examples}, which demonstrate that ERA exhibits sophisticated long-term reasoning, planning, and action execution across both tasks.

\textbf{Overall Results.} ERA establishes a new state-of-the-art among training-based agents, substantially outperforming previous RL baselines on both high-level planning (+12.4\% over VAGEN) and low-level manipulation (+25.4\% over VAGEN). ERA achieves average success rates of 65.2\% on EB-ALFRED and 48.3\% on EB-Manipulation, exceeding proprietary models such as GPT-4o by 8.4 and 19.4 points, respectively. Remarkably, these results are obtained with a compact 3B model, underscoring the parameter efficiency of ERA for embodied agents. On EB-ALFRED, ERA-3B is also competitive with the top-performing large proprietary model Claude-3.5-Sonnet (66.4\%).

\textbf{Unseen Task Generalization.} On EB-ALFRED, RL4VLM and VAGEN perform comparably to ERA on the three seen subsets but fall far behind on the two unseen subsets. For example, ERA achieves 66\% on the \emph{Spatial} subset, outperforming VAGEN (28\%) by 38 points. Similar patterns are observed on EB-Manipulation. These results show that ERA learns robust and transferable skills rather than overfitting to the training tasks.

\textbf{Benefits of EPL and RL in ERA.} The results further highlight the complementary roles of EPL and RL in ERA. EPL alone provides a strong foundation, reaching success rates of 56.0\% on EB-ALFRED and 40.0\% on EB-Manipulation. Adding the RL stage yields substantial gains, improving average success rates by 9.2 and 8.3 points, respectively. These improvements are especially pronounced on unseen subsets, with average gains of 13.0 points on EB-ALFRED and 11.5 points on EB-Manipulation. Together, these findings confirm that EPL imparts essential foundational knowledge, while online RL effectively refines the policy to boost generalization.

\begin{table*}[t]
\centering
\caption{Ablation results on different prior datasets. We report average success rates on both seen and unseen splits. ``Traj-Aug", ``Env-Anc", and ``Ext-Know" denotes Trajectory-Augmented, Environment-Anchored, and External Knowledge Priors. Stage 1 and Stage 2 correspond to EPL and RL, respectively. Numbers in parentheses indicate gains over the raw trajectory baseline. }
\renewcommand{\arraystretch}{1.4}
\setlength{\tabcolsep}{1pt} 
\resizebox{1\linewidth}{!}{
\begin{tabular}{
    >{\raggedright\arraybackslash}p{5cm}@{}
    *{4}{>{\raggedright\arraybackslash}p{2.1cm}@{}}
    *{4}{>{\raggedright\arraybackslash}p{2.1cm}@{}}
}
\toprule
\multirow{3}{*}{\textbf{Methods}} 
& \multicolumn{4}{c}{\textbf{EB-ALFRED}} 
& \multicolumn{4}{c}{\textbf{EB-Manipulation}} \\
\cmidrule(lr){2-5} \cmidrule(lr){6-9}
& \multicolumn{2}{c}{\textbf{Stage 1}} & \multicolumn{2}{c}{\textbf{Stage 1 \& 2}} 
& \multicolumn{2}{c}{\textbf{Stage 1}} & \multicolumn{2}{c}{\textbf{Stage 1 \& 2}} \\
\cmidrule(lr){2-3} \cmidrule(lr){4-5} \cmidrule(lr){6-7} \cmidrule(lr){8-9}
& \textbf{Seen} & \textbf{Unseen} 
& \textbf{Seen} & \textbf{Unseen} 
& \textbf{Seen} & \textbf{Unseen} 
& \textbf{Seen} & \textbf{Unseen} \\
\midrule
Base Model (No prior injected)            & -- & -- & 0 & 0 & -- & -- & 0 & 0 \\
\midrule
Raw Trajectory  (baseline)             & 59.3 & 32.0 & 64.0 & 36.0 & 25.0 & 7.3 & 44.0 & 21.9 \\
+ Traj-Aug  
& 62.0 (+2.7) & 37.0 (+5.0) & 66.7 (+2.7) & 49.0 (+13.0) 
& 41.4 (+16.4) & 26.1 (+18.8) & 50.3 (+6.3) & 35.5 (+13.6) \\
+ Env-Anc  
& \textbf{63.3 (+4.0)} & 39.0 (+7.0) & \textbf{70.0 (+6.0)} & 42.0 (+6.0) 
& 25.0 (+0.0) & 9.4 (+2.1) & 47.2 (+3.2) & 22.9 (+1.0) \\
+ Ext-Know  
& \textbf{63.3 (+4.0)} & 35.0 (+3.0) & 68.7 (+4.7) & 46.0 (+10.0) 
& 30.3 (+5.3) & 18.8 (+11.5) & 48.5 (+4.5) & 27.1 (+5.2)\\
+ Traj-Aug   + Env-Anc 
& 62.0 (+2.7) & \textbf{47.0 (+15.0)} & 68.7 (+4.7) & \textbf{60.0 (+24.0)} 
& \textbf{45.1 (+20.1)} & \textbf{32.3 (+25.0)} & \textbf{51.4 (+7.4)} & \textbf{43.8 (+21.9)} \\
+ Traj-Aug   + Ext-Know  
& \textbf{63.3 (+4.0)} & 44.0 (+12.0) & 68.7 (+4.7) & 55.0 (+19.0) 
& 37.9 (+12.9) & 31.3 (+24.0) & \textbf{51.4 (+7.4)} & 37.5 (+15.6)\\

\bottomrule
\end{tabular}}
\label{tab:EPL_ablations}
\end{table*}


\subsection{\textbf{Q2:} What role do different prior datasets play in agent performance?}

To evaluate the impact of different prior datasets, we conducted an ablation study on performance using various prior datasets both after stage 1 (i.e., EPL) and after stages 1 \& 2 (i.e., EPL+RL), as reported in Table~\ref{tab:EPL_ablations}. 
Without embodied prior learning, direct RL training from the 3B base model completely fails, yielding scores of 0 on all tasks. Incorporating raw trajectory data substantially improves performance, but generalization to unseen tasks remains limited. In contrast, \textbf{EPL with each prior dataset consistently improves performance over the raw-trajectory baseline in both stages and significantly reduces the seen–unseen gap.} Below, we summarize the key findings.

\begin{figure*}[t]
    \centering
    \includegraphics[width=1.0\textwidth]{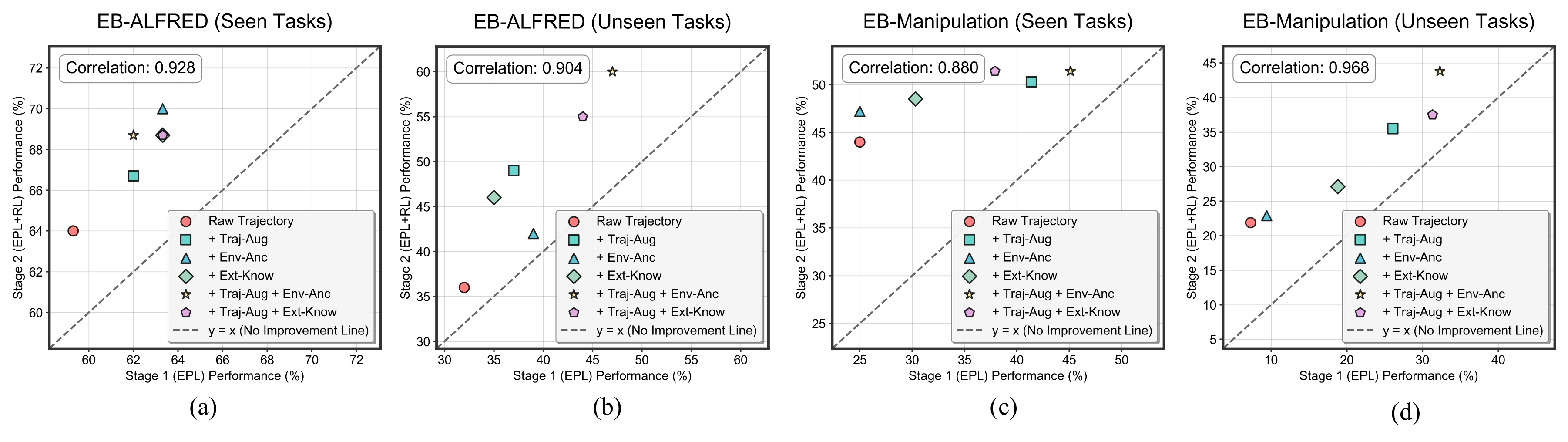}
    \vspace{-20pt}
    \caption{Performance correlation between stage 1 (EPL) and stage 2 (EPL+RL) on both seen and unseen sets. The Pearson correlation coefficient is reported in the top-left corner.}
    \label{fig:correlation_before_after_RL}
\end{figure*}

\textbf{Trajectory-Augmented Priors Achieve the Largest Individual Gains in Generalization.}
Among the three individual datasets, trajectory augmentation yields the strongest improvements on unseen tasks. On EB-ALFRED, augmenting raw trajectories with structured reasoning improves unseen performance by +13.0\% after Stage~2, relative to the raw-trajectory baseline. The effect is even stronger on EB-Manipulation, with a +13.6\% gain on unseen tasks. In comparison, Environment-Anchored and External Knowledge Priors achieve relatively modest improvements of +6.0\%/+10.0\% on EB-ALFRED (unseen) and +1.0\%/+5.2\% on EB-Manipulation (unseen). These results highlight the importance of structured reasoning in enhancing transfer to novel tasks.

\textbf{Environment-Anchored Priors Improve Seen and Unseen Tasks Equally, While External Knowledge Priors Favor Unseen Tasks.}  Environment-Anchored Priors produce balanced improvements across seen and unseen subsets. For instance, on EB-ALFRED they deliver a +6\% gain over the raw-trajectory baseline for both seen and unseen tasks after stage~2. This consistency suggests that environment-anchored data encode environment-level knowledge that is useful across tasks in similar environments. In contrast, External Knowledge Priors lead to larger improvements on unseen tasks than on seen ones. This indicates that while external data are less environment-specific, they capture general reasoning and grounding skills that support generalization to novel tasks. However, their overall gains remain smaller than trajectory augmentation.

\textbf{Combining Trajectory-Augmented and Environment-Anchored Priors Elicit the Best Performance.} We next examine whether combining different priors leads to further gains. Notably, the combination of Trajectory-Augmented and Environment-Anchored Priors achieves the strongest overall results: after stages~1 and~2, performance reaches 60\% on the unseen subsets of EB-ALFRED (+24\% over baseline) and 43.8\% on the unseen subsets of EB-Manipulation (+21.9\% over baseline). While combining Trajectory-Augmented and External Knowledge Priors also produces substantial improvements, the gains are relatively smaller. These findings suggest that environment-anchored data provide explicit, task-relevant supervision that complements trajectory-based priors, whereas external knowledge offers a weaker but more easily obtainable alternative when environment-specific data are costly to curate.

\textbf{Performance Correlation between EPL and RL.} Figure~\ref{fig:correlation_before_after_RL} illustrates the relationship between model performance after stage~1 (EPL) and stage~2 (EPL+RL) across different prior datasets. Using the Pearson correlation coefficient, we observe an often near-linear correlation between the two stages, with correlation coefficients ranging from 0.88 to 0.97. This finding indicates that models achieving higher performance during the EPL stage tend to maintain or even amplify their advantage after RL fine-tuning.
Such a trend indicates that EPL performance is a reliable predictor of final RL outcomes, emphasizing the crucial role of embodied prior learning. In other words, \textbf{investing in stronger prior learning substantially improves the downstream effectiveness of reinforcement learning, leading to more capable and generalizable embodied agents.}

\begin{figure*}[t]
    \centering
    \includegraphics[width=1.0\textwidth]{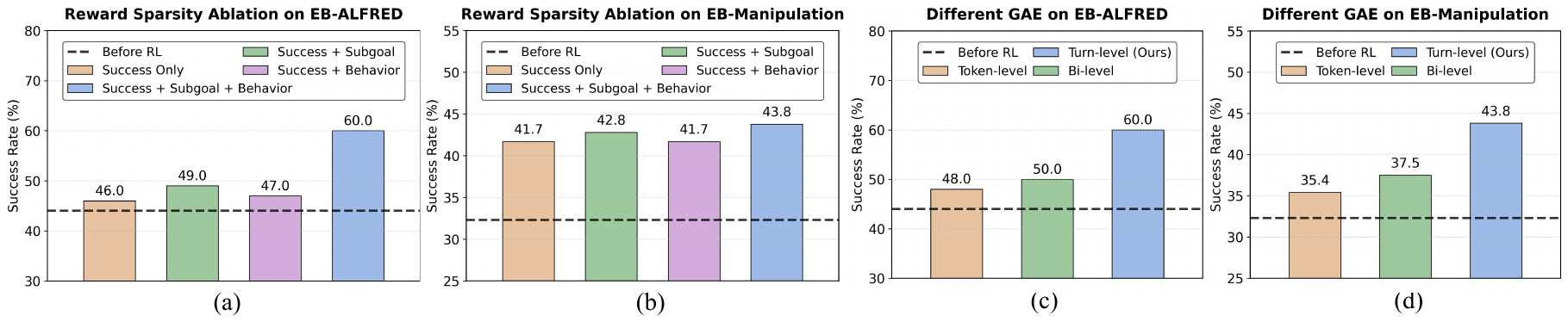}
    \vspace{-15pt}
    \caption{
    (a)(b) Reward design ablations and (c)(d) Value estimation method comparisons in ERA. We report average success rates on the unseen subsets of EB-ALFRED and EB-Manipulation.}
    \label{fig:combined_ablation}
\end{figure*}

\vspace{15pt}

\subsection{\textbf{Q3:} Is Self-Summarization Effective for Context Management?}
\label{exp:context}


\begin{wraptable}{r}{0.55\linewidth}
\centering\small
\caption{Comparison of average success rate (SR) and average input tokens with varying number of history steps included in the context. SR is averaged over unseen subsets.}
\label{tab:ablation_interaction_history_compact}
\renewcommand{\arraystretch}{1.1}
\setlength\tabcolsep{6pt}
\resizebox{1\linewidth}{!}{%
\begin{tabular}{
>{\centering\arraybackslash}p{1.9cm}
>{\centering\arraybackslash}p{1.5cm}
>{\centering\arraybackslash}p{2.4cm}
>{\centering\arraybackslash}p{1.4cm}
>{\centering\arraybackslash}p{2.4cm}
}
\toprule
\multirow{2}{*}{\textbf{History Length}}
& \multicolumn{2}{c}{\bf EB-ALFRED}
& \multicolumn{2}{c}{\bf EB-Manipulation} \\
\cmidrule(lr){2-3}\cmidrule(lr){4-5}
~ & \textbf{SR (\%) $\uparrow$} & \textbf{\#Input Tokens $\downarrow$} & \textbf{SR (\%) $\uparrow$} & \textbf{\#Input Tokens $\downarrow$} \\
\midrule
\multicolumn{5}{c}{\textit{No History}} \\
\midrule
0 step & 2 & \textbf{43} & 3.1 & \textbf{179} \\
\midrule
\multicolumn{5}{c}{\textit{\textbf{w/} Self-Summarization}} \\
\midrule
\rowcolor{gray!20} 1 step (Ours) & \textbf{47} & 217.4 & \textbf{32.3} & 399.5 \\
3 steps       & 45 & 490.5 & 30.3 & 798.3 \\
5 steps       & 46 & 680.4 & 29.1 & 998.3 \\
\midrule
\multicolumn{5}{c}{\textit{\textbf{w/o} Self-Summarization}} \\
\midrule
1 step        & 40 & 157.0 & 29.2 & 305.3 \\
3 steps       & 38 & 332.7 & 24.4 & 564.4 \\
5 steps       & 37 & 455.8 & 25.4 & 694.5 \\
\bottomrule
\end{tabular}}
\end{wraptable}

To evaluate the effectiveness of the proposed self-summarization mechanism, we conduct an ablation study in Table~\ref{tab:ablation_interaction_history_compact} comparing unseen task performance with and without self-summarization after Stage~1. The key difference is that the \textit{w/o Self-Summarization} setting excludes the model’s generated reflection from the context, while retaining other history information identical to the self-summarization setting. Overall, interaction history plays a crucial role in embodied agents. Removing it causes success rates to drop drastically to 2\% and 3.1\% on EB-ALFRED and EB-Manipulation, respectively. Incorporating self-summarization substantially improves performance, yielding an 8–10\% increase across different context lengths. Notably, with self-summarization, using only a one-step history already outperforms the 3–5 step history variants, while requiring fewer tokens, likely due to the distraction introduced by redundant information. \textbf{These findings demonstrate that a concise summary generated by the model can provide a more efficient history representation, enabling the agent to proactively focus on relevant context without being hindered by lengthy histories.}

\subsection{\textbf{Q4:} How do reward design and turn-level value impact RL performance?}\label{exp:rl_reward_gae_ablations}
To assess the effect of two key RL design choices, we conduct ablations in Figure~\ref{fig:combined_ablation}, reporting average success rates on unseen subsets of EB-ALFRED and EB-Manipulation. All comparisons are trained under identical settings as ERA, ensuring a fair comparison.

\textbf{Synergistic Dense Reward Improves Long-Horizon RL.}
Reward sparsity poses a major challenge for credit assignment, particularly in long-horizon tasks. As shown in Figure~\ref{fig:combined_ablation}, supplementing sparse success-based rewards with two process-level signals (subgoal-based and behavior-shaping rewards) substantially improves performance, particularly for high-level planning tasks. On EB-ALFRED, the average success rate rises by 14\% (46\% $\to$ 60\%) compared to training with only success-based rewards. In contrast, the gain on the shorter-horizon EB-Manipulation benchmark is modest (+2.1\%). This disparity shows that \textbf{dense rewards are especially critical for long-horizon tasks}, where they can guide exploration and stabilize credit assignment. Moreover, using only subgoal-based or only behavior-shaping rewards produces limited gains, highlighting the synergistic effect of combining multiple dense reward signals.

\textbf{Turn-Level Value Estimation Enhances Policy Learning.}
We also compare three value learning schemes: token-level, bi-level \citep{wang2025vagen}, and our turn-level GAE. Token-level and bi-level approaches require learning a value function over individual tokens, distributing fine-grained credit across reasoning and action tokens. In contrast, turn-level GAE treats the entire response as a single action and learns a turn-level value function. Results show that turn-level GAE achieves the strongest performance, improving average success rates by 12 points on EB-ALFRED (48\% $\to$ 60\%) and by 8.4 points on EB-Manipulation (35.4\% $\to$ 43.8\%) compared to token-level GAE. Bi-level GAE offers modest gains over token-level GAE, likely because it incorporates partial turn-level credit assignment, but its reliance on unstable token-level value estimation still limits effectiveness. These results validate that turn-level credit assignment yields more stable policy learning.

\section{Conclusion}
\label{sec:conclusion}
In this work, we propose a two-stage framework for transforming VLMs into capable embodied reasoning agents. In the first stage, we show that enriching existing trajectory data with structured reasoning, incorporating environment-anchored supervision, and leveraging external knowledge each improve agent performance, with their combination yielding even greater gains. In the second stage, we highlight the importance of careful design choices in context management, dense reward shaping, and turn-level value learning, providing insights into the factors that drive effective RL for embodied agents. These contributions establish a practical and scalable recipe for developing more powerful and efficient VLM-based agents in more realistic settings.


\section*{Acknowledgments}

This material is based upon work supported partially by NSF under Grant No. 2416897, Grant No. 2505932, and by ORN under Grant No. N000142512318.
Any opinions, findings, and conclusions or recommendations expressed in this material are those of the author(s) and do not necessarily reflect the views of the funding agencies. This research used both Delta (NSF award OAC 2005572) and DeltaAI (NSF  award OAC 2320345) advanced computing systems, as well as TAAC computing resources via NAIRR Pilot NAIRR250157.

Huan Zhang acknowledges the support of the University Research Program (URP) from Toyota Research Institute (TRI). This research reflects solely the opinions and conclusions of its authors, not those of TRI or any other Toyota entity.

\bibliography{arxiv}

\begin{thebibliography}{113}
\providecommand{\natexlab}[1]{#1}
\providecommand{\url}[1]{\texttt{#1}}
\expandafter\ifx\csname urlstyle\endcsname\relax
  \providecommand{\doi}[1]{doi: #1}\else
  \providecommand{\doi}{doi: \begingroup \urlstyle{rm}\Url}\fi

\bibitem[Awais et~al.(2025)Awais, Naseer, Khan, Anwer, Cholakkal, Shah, Yang, and Khan]{awais2025foundation}
Muhammad Awais, Muzammal Naseer, Salman Khan, Rao~Muhammad Anwer, Hisham Cholakkal, Mubarak Shah, Ming-Hsuan Yang, and Fahad~Shahbaz Khan.
\newblock Foundation models defining a new era in vision: a survey and outlook.
\newblock \emph{IEEE Transactions on Pattern Analysis and Machine Intelligence}, 2025.

\bibitem[Bai et~al.(2025)Bai, Chen, Liu, Wang, Ge, Song, Dang, Wang, Wang, Tang, Zhong, Zhu, Yang, Li, Wan, Wang, Ding, Fu, Xu, Ye, Zhang, Xie, Cheng, Zhang, Yang, Xu, and Lin]{bai2025qwen25vl}
Shuai Bai, Keqin Chen, Xuejing Liu, Jialin Wang, Wenbin Ge, Sibo Song, Kai Dang, Peng Wang, Shijie Wang, Jun Tang, Humen Zhong, Yuanzhi Zhu, Mingkun Yang, Zhaohai Li, Jianqiang Wan, Pengfei Wang, Wei Ding, Zheren Fu, Yiheng Xu, Jiabo Ye, Xi~Zhang, Tianbao Xie, Zesen Cheng, Hang Zhang, Zhibo Yang, Haiyang Xu, and Junyang Lin.
\newblock Qwen2.5-vl technical report, 2025.

\bibitem[Cao et~al.(2025)Cao, Feng, Huo, and Gao]{cao2025causal}
Hongye Cao, Fan Feng, Jing Huo, and Yang Gao.
\newblock Causal action empowerment for efficient reinforcement learning in embodied agents.
\newblock \emph{Science China Information Sciences}, 68\penalty0 (5):\penalty0 150201, 2025.

\bibitem[Chen et~al.(2025{\natexlab{a}})Chen, Li, Wang, Jiang, Liu, Lu, Huang, Byeon, Le, Rintamaki, et~al.]{chen2025eagle}
Guo Chen, Zhiqi Li, Shihao Wang, Jindong Jiang, Yicheng Liu, Lidong Lu, De-An Huang, Wonmin Byeon, Matthieu Le, Tuomas Rintamaki, et~al.
\newblock Eagle 2.5: Boosting long-context post-training for frontier vision-language models.
\newblock \emph{arXiv preprint arXiv:2504.15271}, 2025{\natexlab{a}}.

\bibitem[Chen et~al.(2025{\natexlab{b}})Chen, Wu, Liu, Pan, Liu, Xie, Yu, and Ruan]{chen2025janus}
Xiaokang Chen, Zhiyu Wu, Xingchao Liu, Zizheng Pan, Wen Liu, Zhenda Xie, Xingkai Yu, and Chong Ruan.
\newblock Janus-pro: Unified multimodal understanding and generation with data and model scaling.
\newblock \emph{arXiv preprint arXiv:2501.17811}, 2025{\natexlab{b}}.

\bibitem[Chen et~al.(2024{\natexlab{a}})Chen, Cui, Chen, Tan, Zhang, Zhao, and Wang]{chen2024robogptintelligentagentmaking}
Yaran Chen, Wenbo Cui, Yuanwen Chen, Mining Tan, Xinyao Zhang, Dongbin Zhao, and He~Wang.
\newblock Robogpt: an intelligent agent of making embodied long-term decisions for daily instruction tasks, 2024{\natexlab{a}}.
\newblock URL \url{https://arxiv.org/abs/2311.15649}.

\bibitem[Chen et~al.(2024{\natexlab{b}})Chen, Wang, Cao, Liu, Gao, Cui, Zhu, Ye, Tian, Liu, et~al.]{chen2024expanding}
Zhe Chen, Weiyun Wang, Yue Cao, Yangzhou Liu, Zhangwei Gao, Erfei Cui, Jinguo Zhu, Shenglong Ye, Hao Tian, Zhaoyang Liu, et~al.
\newblock Expanding performance boundaries of open-source multimodal models with model, data, and test-time scaling.
\newblock \emph{arXiv preprint arXiv:2412.05271}, 2024{\natexlab{b}}.

\bibitem[Cheng et~al.(2025)Cheng, Tu, Li, Dai, Hu, Hu, Li, Shi, Yu, Chen, et~al.]{cheng2025embodiedeval}
Zhili Cheng, Yuge Tu, Ran Li, Shiqi Dai, Jinyi Hu, Shengding Hu, Jiahao Li, Yang Shi, Tianyu Yu, Weize Chen, et~al.
\newblock Embodiedeval: Evaluate multimodal llms as embodied agents.
\newblock \emph{arXiv preprint arXiv:2501.11858}, 2025.

\bibitem[Choi et~al.(2024)Choi, Yoon, Ong, Kim, and Jang]{choi2024lotabench}
Jae-Woo Choi, Youngwoo Yoon, Hyobin Ong, Jaehong Kim, and Minsu Jang.
\newblock Lota-bench: Benchmarking language-oriented task planners for embodied agents.
\newblock \emph{arXiv preprint arXiv:2402.08178}, 2024.

\bibitem[Deitke et~al.(2020)Deitke, Kembhavi, Rastegari, Farhadi, and Mottaghi]{deitke2020robothor}
Matt Deitke, Aniruddha Kembhavi, Mohammad Rastegari, Ali Farhadi, and Roozbeh Mottaghi.
\newblock {RoboTHOR}: An open simulation-to-real embodied ai platform.
\newblock In \emph{Proceedings of the IEEE/CVF Conference on Computer Vision and Pattern Recognition}, 2020.

\bibitem[Deitke et~al.(2024)Deitke, Clark, Lee, Tripathi, Yang, Park, Salehi, Muennighoff, Lo, Soldaini, et~al.]{deitke2024molmo}
Matt Deitke, Christopher Clark, Sangho Lee, Rohun Tripathi, Yue Yang, Jae~Sung Park, Mohammadreza Salehi, Niklas Muennighoff, Kyle Lo, Luca Soldaini, et~al.
\newblock Molmo and pixmo: Open weights and open data for state-of-the-art multimodal models.
\newblock \emph{arXiv e-prints}, pp.\  arXiv--2409, 2024.

\bibitem[Deng et~al.(2025)Deng, Zhu, Li, Gou, Li, Wang, Zhong, Yu, Nie, Song, et~al.]{deng2025emerging}
Chaorui Deng, Deyao Zhu, Kunchang Li, Chenhui Gou, Feng Li, Zeyu Wang, Shu Zhong, Weihao Yu, Xiaonan Nie, Ziang Song, et~al.
\newblock Emerging properties in unified multimodal pretraining.
\newblock \emph{arXiv preprint arXiv:2505.14683}, 2025.

\bibitem[Driess et~al.(2023)Driess, Xia, Sajjadi, Lynch, Chowdhery, Ichter, Wahid, Tompson, Vuong, Yu, et~al.]{driess2023palm}
Danny Driess, Fei Xia, Mehdi~SM Sajjadi, Corey Lynch, Aakanksha Chowdhery, Brian Ichter, Ayzaan Wahid, Jonathan Tompson, Quan Vuong, Tianhe Yu, et~al.
\newblock Palm-e: an embodied multimodal language model.
\newblock In \emph{Proceedings of the 40th International Conference on Machine Learning}, pp.\  8469--8488, 2023.

\bibitem[Feng et~al.(2025{\natexlab{a}})Feng, Tan, Lyu, Zheng, Xu, Yan, Huang, and An]{feng2025towards}
Lang Feng, Weihao Tan, Zhiyi Lyu, Longtao Zheng, Haiyang Xu, Ming Yan, Fei Huang, and Bo~An.
\newblock Towards efficient online tuning of vlm agents via counterfactual soft reinforcement learning.
\newblock \emph{arXiv preprint arXiv:2505.03792}, 2025{\natexlab{a}}.

\bibitem[Feng et~al.(2025{\natexlab{b}})Feng, Xue, Liu, and An]{feng2025gigpo}
Lang Feng, Zhenghai Xue, Tingcong Liu, and Bo~An.
\newblock Group-in-group policy optimization for llm agent training, 2025{\natexlab{b}}.
\newblock URL \url{https://arxiv.org/abs/2505.10978}.

\bibitem[Gao et~al.(2024)Gao, Sarkar, Xia, Xiao, Wu, Ichter, Majumdar, and Sadigh]{gao2024physicallygroundedvisionlanguagemodels}
Jensen Gao, Bidipta Sarkar, Fei Xia, Ted Xiao, Jiajun Wu, Brian Ichter, Anirudha Majumdar, and Dorsa Sadigh.
\newblock Physically grounded vision-language models for robotic manipulation, 2024.
\newblock URL \url{https://arxiv.org/abs/2309.02561}.

\bibitem[Gulcehre et~al.(2023)Gulcehre, Paine, Srinivasan, Konyushkova, Weerts, Sharma, Siddhant, Ahern, Wang, Gu, et~al.]{gulcehre2023reinforced}
Caglar Gulcehre, Tom~Le Paine, Srivatsan Srinivasan, Ksenia Konyushkova, Lotte Weerts, Abhishek Sharma, Aditya Siddhant, Alex Ahern, Miaosen Wang, Chenjie Gu, et~al.
\newblock Reinforced self-training (rest) for language modeling.
\newblock \emph{arXiv preprint arXiv:2308.08998}, 2023.

\bibitem[Guo et~al.(2025)Guo, Yang, Zhang, Song, Zhang, Xu, Zhu, Ma, Wang, Bi, et~al.]{guo2025deepseek}
Daya Guo, Dejian Yang, Haowei Zhang, Junxiao Song, Ruoyu Zhang, Runxin Xu, Qihao Zhu, Shirong Ma, Peiyi Wang, Xiao Bi, et~al.
\newblock Deepseek-r1: Incentivizing reasoning capability in llms via reinforcement learning.
\newblock \emph{arXiv preprint arXiv:2501.12948}, 2025.

\bibitem[Hu et~al.(2023)Hu, Lin, Zhang, Yi, and Gao]{hu2023lookleapunveilingpower}
Yingdong Hu, Fanqi Lin, Tong Zhang, Li~Yi, and Yang Gao.
\newblock Look before you leap: Unveiling the power of gpt-4v in robotic vision-language planning, 2023.
\newblock URL \url{https://arxiv.org/abs/2311.17842}.

\bibitem[Huang et~al.(2025)Huang, Wang, Lin, Hu, Wen, and Gao]{huang2025tactilevlaunlockingvisionlanguageactionmodels}
Jialei Huang, Shuo Wang, Fanqi Lin, Yihang Hu, Chuan Wen, and Yang Gao.
\newblock Tactile-vla: Unlocking vision-language-action model's physical knowledge for tactile generalization, 2025.
\newblock URL \url{https://arxiv.org/abs/2507.09160}.

\bibitem[Huang et~al.(2022)Huang, Xia, Xiao, Chan, Liang, Florence, Zeng, Tompson, Mordatch, Chebotar, et~al.]{huang2022inner}
Wenlong Huang, Fei Xia, Ted Xiao, Harris Chan, Jacky Liang, Pete Florence, Andy Zeng, Jonathan Tompson, Igor Mordatch, Yevgen Chebotar, et~al.
\newblock Inner monologue: Embodied reasoning through planning with language models.
\newblock \emph{arXiv preprint arXiv:2207.05608}, 2022.

\bibitem[Huang et~al.(2023)Huang, Wang, Zhang, Li, Wu, and Fei-Fei]{huang2023voxposer}
Wenlong Huang, Chen Wang, Ruohan Zhang, Yunzhu Li, Jiajun Wu, and Li~Fei-Fei.
\newblock Voxposer: Composable 3d value maps for robotic manipulation with language models.
\newblock \emph{arXiv preprint arXiv:2307.05973}, 2023.

\bibitem[Huang et~al.(2024)Huang, Wang, Li, Zhang, and Fei-Fei]{huang2024rekep}
Wenlong Huang, Chen Wang, Yunzhu Li, Ruohan Zhang, and Li~Fei-Fei.
\newblock Rekep: Spatio-temporal reasoning of relational keypoint constraints for robotic manipulation.
\newblock \emph{arXiv preprint arXiv:2409.01652}, 2024.

\bibitem[Jaech et~al.(2024)Jaech, Kalai, Lerer, Richardson, El-Kishky, Low, Helyar, Madry, Beutel, Carney, et~al.]{jaech2024openai}
Aaron Jaech, Adam Kalai, Adam Lerer, Adam Richardson, Ahmed El-Kishky, Aiden Low, Alec Helyar, Aleksander Madry, Alex Beutel, Alex Carney, et~al.
\newblock Openai o1 system card.
\newblock \emph{arXiv preprint arXiv:2412.16720}, 2024.

\bibitem[Ji et~al.(2025)Ji, Tan, Shi, Hao, Zhang, Zhang, Wang, Zhao, Mu, An, Xue, Su, Lyu, Zheng, Liu, Wang, and Zhang]{ji2025robobrainunifiedbrainmodel}
Yuheng Ji, Huajie Tan, Jiayu Shi, Xiaoshuai Hao, Yuan Zhang, Hengyuan Zhang, Pengwei Wang, Mengdi Zhao, Yao Mu, Pengju An, Xinda Xue, Qinghang Su, Huaihai Lyu, Xiaolong Zheng, Jiaming Liu, Zhongyuan Wang, and Shanghang Zhang.
\newblock Robobrain: A unified brain model for robotic manipulation from abstract to concrete, 2025.
\newblock URL \url{https://arxiv.org/abs/2502.21257}.

\bibitem[Kim et~al.(2024{\natexlab{a}})Kim, Kim, Kim, Min, and Choi]{kim2024contextawareplanningenvironmentawarememory}
Byeonghwi Kim, Jinyeon Kim, Yuyeong Kim, Cheolhong Min, and Jonghyun Choi.
\newblock Context-aware planning and environment-aware memory for instruction following embodied agents, 2024{\natexlab{a}}.
\newblock URL \url{https://arxiv.org/abs/2308.07241}.

\bibitem[Kim et~al.(2025)Kim, Park, Jang, Shin, Kim, and Seo]{kim2025robot}
Dongyoung Kim, Sumin Park, Huiwon Jang, Jinwoo Shin, Jaehyung Kim, and Younggyo Seo.
\newblock Robot-r1: Reinforcement learning for enhanced embodied reasoning in robotics.
\newblock \emph{arXiv preprint arXiv:2506.00070}, 2025.

\bibitem[Kim et~al.(2024{\natexlab{b}})Kim, Pertsch, Karamcheti, Xiao, Balakrishna, Nair, Rafailov, Foster, Lam, Sanketi, et~al.]{kim2024openvla}
Moo~Jin Kim, Karl Pertsch, Siddharth Karamcheti, Ted Xiao, Ashwin Balakrishna, Suraj Nair, Rafael Rafailov, Ethan Foster, Grace Lam, Pannag Sanketi, et~al.
\newblock Openvla: An open-source vision-language-action model.
\newblock \emph{arXiv preprint arXiv:2406.09246}, 2024{\natexlab{b}}.

\bibitem[Kingma \& Ba(2014)Kingma and Ba]{kingma2014adam}
Diederik~P Kingma and Jimmy Ba.
\newblock Adam: A method for stochastic optimization.
\newblock \emph{arXiv preprint arXiv:1412.6980}, 2014.

\bibitem[Kolve et~al.(2017)Kolve, Mottaghi, Gordon, Zhu, Gupta, and Farhadi]{kolve2017ai2thor}
Eric Kolve, Roozbeh Mottaghi, Daniel Gordon, Yuke Zhu, Abhinav Gupta, and Ali Farhadi.
\newblock {AI2-THOR}: An interactive 3d environment for visual {AI}.
\newblock In \emph{arXiv preprint arXiv:1712.05474}, 2017.

\bibitem[Lee et~al.(2025)Lee, Duan, Fang, Deng, Liu, Li, Fang, Zhang, Wang, Lee, et~al.]{lee2025molmoact}
Jason Lee, Jiafei Duan, Haoquan Fang, Yuquan Deng, Shuo Liu, Boyang Li, Bohan Fang, Jieyu Zhang, Yi~Ru Wang, Sangho Lee, et~al.
\newblock Molmoact: Action reasoning models that can reason in space.
\newblock \emph{arXiv preprint arXiv:2508.07917}, 2025.

\bibitem[Li et~al.(2024{\natexlab{a}})Li, Zhang, Guo, Zhang, Li, Zhang, Zhang, Zhang, Li, Liu, et~al.]{li2024llava}
Bo~Li, Yuanhan Zhang, Dong Guo, Renrui Zhang, Feng Li, Hao Zhang, Kaichen Zhang, Peiyuan Zhang, Yanwei Li, Ziwei Liu, et~al.
\newblock Llava-onevision: Easy visual task transfer.
\newblock \emph{arXiv preprint arXiv:2408.03326}, 2024{\natexlab{a}}.

\bibitem[Li et~al.(2024{\natexlab{b}})Li, Zhang, Wang, Zhong, Chen, Chu, Liu, and Jia]{li2024mini}
Yanwei Li, Yuechen Zhang, Chengyao Wang, Zhisheng Zhong, Yixin Chen, Ruihang Chu, Shaoteng Liu, and Jiaya Jia.
\newblock Mini-gemini: Mining the potential of multi-modality vision language models.
\newblock \emph{arXiv preprint arXiv:2403.18814}, 2024{\natexlab{b}}.

\bibitem[Li et~al.(2025{\natexlab{a}})Li, Lai, Bao, Tan, Dao, Sui, Shen, Liu, Liu, and Kong]{li2025visual}
Yifan Li, Zhixin Lai, Wentao Bao, Zhen Tan, Anh Dao, Kewei Sui, Jiayi Shen, Dong Liu, Huan Liu, and Yu~Kong.
\newblock Visual large language models for generalized and specialized applications.
\newblock \emph{arXiv preprint arXiv:2501.02765}, 2025{\natexlab{a}}.

\bibitem[Li et~al.(2025{\natexlab{b}})Li, Zhang, Lin, Liu, Cai, Liu, and Zhao]{li2025sti}
Yun Li, Yiming Zhang, Tao Lin, XiangRui Liu, Wenxiao Cai, Zheng Liu, and Bo~Zhao.
\newblock Sti-bench: Are mllms ready for precise spatial-temporal world understanding?
\newblock \emph{arXiv preprint arXiv:2503.23765}, 2025{\natexlab{b}}.

\bibitem[Liang et~al.(2023)Liang, Huang, Xia, Xu, Hausman, Ichter, Florence, and Zeng]{liang2023codepolicieslanguagemodel}
Jacky Liang, Wenlong Huang, Fei Xia, Peng Xu, Karol Hausman, Brian Ichter, Pete Florence, and Andy Zeng.
\newblock Code as policies: Language model programs for embodied control, 2023.
\newblock URL \url{https://arxiv.org/abs/2209.07753}.

\bibitem[Lightman et~al.(2023)Lightman, Kosaraju, Burda, Edwards, Baker, Lee, Leike, Schulman, Sutskever, and Cobbe]{lightman2023let}
Hunter Lightman, Vineet Kosaraju, Yuri Burda, Harrison Edwards, Bowen Baker, Teddy Lee, Jan Leike, John Schulman, Ilya Sutskever, and Karl Cobbe.
\newblock Let's verify step by step.
\newblock In \emph{The Twelfth International Conference on Learning Representations}, 2023.

\bibitem[Lin et~al.(2025)Lin, Huang, Li, Jiang, Wu, Zhong, Qian, Wang, and Qi]{lin2025embrace}
Mingxian Lin, Wei Huang, Yitang Li, Chengjie Jiang, Kui Wu, Fangwei Zhong, Shengju Qian, Xin Wang, and Xiaojuan Qi.
\newblock Embrace-3k: Embodied reasoning and action in complex environments.
\newblock \emph{arXiv preprint arXiv:2507.10548}, 2025.

\bibitem[Liu et~al.(2024{\natexlab{a}})Liu, Li, Li, Li, Zhang, Shen, and Lee]{liu2024llavanext}
Haotian Liu, Chunyuan Li, Yuheng Li, Bo~Li, Yuanhan Zhang, Sheng Shen, and Yong~Jae Lee.
\newblock Llava-next: Improved reasoning, ocr, and world knowledge, January 2024{\natexlab{a}}.
\newblock URL \url{https://llava-vl.github.io/blog/2024-01-30-llava-next/}.

\bibitem[Liu et~al.(2025{\natexlab{a}})Liu, Li, Li, Liu, Wang, Liu, Kang, Ma, Kong, and Zhang]{liu2025towards}
Huaping Liu, Xinghang Li, Peiyan Li, Minghuan Liu, Dong Wang, Jirong Liu, Bingyi Kang, Xiao Ma, Tao Kong, and Hanbo Zhang.
\newblock Towards generalist robot policies: What matters in building vision-language-action models.
\newblock 2025{\natexlab{a}}.

\bibitem[Liu et~al.(2025{\natexlab{b}})Liu, Gao, Wei, Chen, Liao, Wu, Yu, and Wang]{liu2025whatcan}
Jijia Liu, Feng Gao, Bingwen Wei, Xinlei Chen, Qingmin Liao, Yi~Wu, Chao Yu, and Yu~Wang.
\newblock What can rl bring to vla generalization? an empirical study.
\newblock \emph{arXiv preprint arXiv:2505.19789}, 2025{\natexlab{b}}.

\bibitem[Liu et~al.(2024{\natexlab{b}})Liu, Zhang, Gu, Iong, Xu, Song, Zhang, Lai, Liu, Zhao, et~al.]{liu2024visualagentbench}
Xiao Liu, Tianjie Zhang, Yu~Gu, Iat~Long Iong, Yifan Xu, Xixuan Song, Shudan Zhang, Hanyu Lai, Xinyi Liu, Hanlin Zhao, et~al.
\newblock Visualagentbench: Towards large multimodal models as visual foundation agents.
\newblock \emph{arXiv preprint arXiv:2408.06327}, 2024{\natexlab{b}}.

\bibitem[Lu et~al.(2025{\natexlab{a}})Lu, Guo, Zhang, Zhou, Jiang, Gao, Tang, and Wang]{lu2025vla}
Guanxing Lu, Wenkai Guo, Chubin Zhang, Yuheng Zhou, Haonan Jiang, Zifeng Gao, Yansong Tang, and Ziwei Wang.
\newblock Vla-rl: Towards masterful and general robotic manipulation with scalable reinforcement learning.
\newblock \emph{arXiv preprint arXiv:2505.18719}, 2025{\natexlab{a}}.

\bibitem[Lu et~al.(2025{\natexlab{b}})Lu, Guo, Zhang, Zhou, Jiang, Gao, Tang, and Wang]{lu2025vlarlmasterfulgeneralrobotic}
Guanxing Lu, Wenkai Guo, Chubin Zhang, Yuheng Zhou, Haonan Jiang, Zifeng Gao, Yansong Tang, and Ziwei Wang.
\newblock Vla-rl: Towards masterful and general robotic manipulation with scalable reinforcement learning, 2025{\natexlab{b}}.
\newblock URL \url{https://arxiv.org/abs/2505.18719}.

\bibitem[Lu et~al.(2023)Lu, Bansal, Xia, Liu, Li, Hajishirzi, Cheng, Chang, Galley, and Gao]{lu2023mathvista}
Pan Lu, Hritik Bansal, Tony Xia, Jiacheng Liu, Chunyuan Li, Hannaneh Hajishirzi, Hao Cheng, Kai-Wei Chang, Michel Galley, and Jianfeng Gao.
\newblock Mathvista: Evaluating mathematical reasoning of foundation models in visual contexts.
\newblock \emph{arXiv preprint arXiv:2310.02255}, 2023.

\bibitem[Luo et~al.(2025)Luo, Yang, Sun, Deng, Yao, Shen, Zhang, and Chen]{luo2025rethinking}
Feng Luo, Rui Yang, Hao Sun, Chunyuan Deng, Jiarui Yao, Jingyan Shen, Huan Zhang, and Hanjie Chen.
\newblock Rethinking diverse human preference learning through principal component analysis.
\newblock \emph{arXiv preprint arXiv:2502.13131}, 2025.

\bibitem[Ma et~al.(2024)Ma, Song, Zhuang, Hao, and King]{ma2024survey}
Yueen Ma, Zixing Song, Yuzheng Zhuang, Jianye Hao, and Irwin King.
\newblock A survey on vision-language-action models for embodied ai.
\newblock \emph{arXiv preprint arXiv:2405.14093}, 2024.

\bibitem[McDermott et~al.(1998)McDermott, Ghallab, Howe, Knoblock, Ram, Veloso, Weld, and Wilkins]{mcdermott1998pddl}
Drew McDermott, Malik Ghallab, Adele Howe, Craig Knoblock, Ashwin Ram, Manuela Veloso, Daniel Weld, and David Wilkins.
\newblock Pddl---the planning domain definition language.
\newblock Technical report, Yale Center for Computational Vision and Control / Yale University / DCS, 1998.

\bibitem[Mu et~al.(2024)Mu, Zhang, Hu, Wang, Ding, Jin, Wang, Dai, Qiao, and Luo]{mu2024embodiedgpt}
Yao Mu, Qinglong Zhang, Mengkang Hu, Wenhai Wang, Mingyu Ding, Jun Jin, Bin Wang, Jifeng Dai, Yu~Qiao, and Ping Luo.
\newblock Embodiedgpt: Vision-language pre-training via embodied chain of thought.
\newblock \emph{Advances in Neural Information Processing Systems}, 36, 2024.

\bibitem[{Open O1 Team}(2024)]{openo1_2024b}
{Open O1 Team}.
\newblock Open o1, 2024.
\newblock Dataset.

\bibitem[OpenAI(2024)]{GPT-4o}
OpenAI.
\newblock Hello gpt-4o, 2024.
\newblock URL \url{https://openai.com/index/hello-gpt-4o/}.

\bibitem[Ouyang et~al.(2022)Ouyang, Wu, Jiang, Almeida, Wainwright, Mishkin, Zhang, Agarwal, Slama, Ray, et~al.]{ouyang2022training}
Long Ouyang, Jeffrey Wu, Xu~Jiang, Diogo Almeida, Carroll Wainwright, Pamela Mishkin, Chong Zhang, Sandhini Agarwal, Katarina Slama, Alex Ray, et~al.
\newblock Training language models to follow instructions with human feedback.
\newblock \emph{Advances in neural information processing systems}, 35:\penalty0 27730--27744, 2022.

\bibitem[Patterson et~al.(2022)Patterson, Gonzalez, H{\"o}lzle, Le, Liang, Munguia, Rothchild, So, Texier, and Dean]{patterson2022llama90v}
David Patterson, Joseph Gonzalez, Urs H{\"o}lzle, Quoc Le, Chen Liang, Lluis-Miquel Munguia, Daniel Rothchild, David~R So, Maud Texier, and Jeff Dean.
\newblock The carbon footprint of machine learning training will plateau, then shrink.
\newblock \emph{Computer}, 55\penalty0 (7):\penalty0 18--28, 2022.

\bibitem[Rafailov et~al.(2023)Rafailov, Sharma, Mitchell, Manning, Ermon, and Finn]{rafailov2023direct}
Rafael Rafailov, Archit Sharma, Eric Mitchell, Christopher~D Manning, Stefano Ermon, and Chelsea Finn.
\newblock Direct preference optimization: Your language model is secretly a reward model.
\newblock \emph{Advances in neural information processing systems}, 36:\penalty0 53728--53741, 2023.

\bibitem[Rajbhandari et~al.(2020)Rajbhandari, Ruwase, Rasley, and He]{rajbhandari2020}
Samyam Rajbhandari, Olatunji Ruwase, Jeff Rasley, and Yuxiong He.
\newblock Zero: Memory optimizations toward training trillion parameter models.
\newblock In \emph{Proceedings of the International Conference for High Performance Computing, Networking, Storage and Analysis (SC)}, pp.\  1--16. IEEE, 2020.

\bibitem[Reid et~al.(2024)Reid, Savinov, Teplyashin, Lepikhin, Lillicrap, Alayrac, Soricut, Lazaridou, Firat, Schrittwieser, et~al.]{reid2024gemini}
Machel Reid, Nikolay Savinov, Denis Teplyashin, Dmitry Lepikhin, Timothy Lillicrap, Jean-baptiste Alayrac, Radu Soricut, Angeliki Lazaridou, Orhan Firat, Julian Schrittwieser, et~al.
\newblock Gemini 1.5: Unlocking multimodal understanding across millions of tokens of context.
\newblock \emph{arXiv preprint arXiv:2403.05530}, 2024.

\bibitem[{Remyx AI}(2025)]{RemyxAI_2025_SpaceThinker_Dataset}
{Remyx AI}.
\newblock Spacethinker: Synthetic spatial reasoning traces (vqasynth).
\newblock Hugging Face dataset, 2025.
\newblock URL \url{https://huggingface.co/datasets/remyxai/SpaceThinker}.

\bibitem[Rohmer et~al.(2013)Rohmer, Singh, and Freese]{rohmer2013v}
Elodie Rohmer, Surya Singh, and Daniel Freese.
\newblock Coppeliasim (formerly v-rep): a versatile and scalable robot simulation framework.
\newblock In \emph{Proceedings of the IEEE/RSJ International Conference on Intelligent Robots and Systems (IROS)}, 2013.

\bibitem[Shao et~al.(2024)Shao, Wang, Zhu, Xu, Song, Bi, Zhang, Zhang, Li, Wu, et~al.]{shao2024deepseekmath}
Zhihong Shao, Peiyi Wang, Qihao Zhu, Runxin Xu, Junxiao Song, Xiao Bi, Haowei Zhang, Mingchuan Zhang, YK~Li, Yang Wu, et~al.
\newblock Deepseekmath: Pushing the limits of mathematical reasoning in open language models.
\newblock \emph{arXiv preprint arXiv:2402.03300}, 2024.

\bibitem[Shen et~al.(2025)Shen, Yao, Yang, Sun, Luo, Pan, Zhang, and Zhao]{shen2025micro}
Jingyan Shen, Jiarui Yao, Rui Yang, Yifan Sun, Feng Luo, Rui Pan, Tong Zhang, and Han Zhao.
\newblock Micro: Mixture modeling and context-aware routing for personalized preference learning.
\newblock \emph{arXiv preprint arXiv:2505.24846}, 2025.

\bibitem[Sheng et~al.(2025)Sheng, Zhang, Ye, Wu, Zhang, Zhang, Peng, Lin, and Wu]{sheng2025hybridflow}
Guangming Sheng, Chi Zhang, Zilingfeng Ye, Xibin Wu, Wang Zhang, Ru~Zhang, Yanghua Peng, Haibin Lin, and Chuan Wu.
\newblock Hybridflow: A flexible and efficient rlhf framework.
\newblock In \emph{Proceedings of the Twentieth European Conference on Computer Systems}, pp.\  1279--1297, 2025.

\bibitem[Shin et~al.(2025)Shin, jeon, Kim, Kang, and Zhang]{shin2025socraticplannerselfqabasedzeroshot}
Suyeon Shin, Sujin jeon, Junghyun Kim, Gi-Cheon Kang, and Byoung-Tak Zhang.
\newblock Socratic planner: Self-qa-based zero-shot planning for embodied instruction following, 2025.
\newblock URL \url{https://arxiv.org/abs/2404.15190}.

\bibitem[Shridhar et~al.(2020)Shridhar, Thomason, Gordon, Bisk, Han, Mottaghi, Zettlemoyer, and Fox]{Shridhar_2020_CVPR}
Mohit Shridhar, Jesse Thomason, Daniel Gordon, Yonatan Bisk, Winson Han, Roozbeh Mottaghi, Luke Zettlemoyer, and Dieter Fox.
\newblock Alfred: A benchmark for interpreting grounded instructions for everyday tasks.
\newblock In \emph{Proceedings of the IEEE/CVF Conference on Computer Vision and Pattern Recognition (CVPR)}, June 2020.

\bibitem[Shu et~al.(2025)Shu, Lin, and Wang]{shu2025rftf}
Junyang Shu, Zhiwei Lin, and Yongtao Wang.
\newblock Rftf: Reinforcement fine-tuning for embodied agents with temporal feedback.
\newblock \emph{arXiv preprint arXiv:2505.19767}, 2025.

\bibitem[Silver et~al.(2024)Silver, Dan, Srinivas, Tenenbaum, Kaelbling, and Katz]{silver2024generalized}
Tom Silver, Soham Dan, Kavitha Srinivas, Joshua~B Tenenbaum, Leslie Kaelbling, and Michael Katz.
\newblock Generalized planning in pddl domains with pretrained large language models.
\newblock In \emph{Proceedings of the AAAI conference on artificial intelligence}, volume~38, pp.\  20256--20264, 2024.

\bibitem[Sima et~al.(2024)Sima, Renz, Chitta, Chen, Zhang, Xie, Bei{\ss}wenger, Luo, Geiger, and Li]{sima2024drivelm}
Chonghao Sima, Katrin Renz, Kashyap Chitta, Li~Chen, Hanxue Zhang, Chengen Xie, Jens Bei{\ss}wenger, Ping Luo, Andreas Geiger, and Hongyang Li.
\newblock Drivelm: Driving with graph visual question answering.
\newblock In \emph{European conference on computer vision}, pp.\  256--274. Springer, 2024.

\bibitem[Singh et~al.(2022)Singh, Blukis, Mousavian, Goyal, Xu, Tremblay, Fox, Thomason, and Garg]{singh2022progpromptgeneratingsituatedrobot}
Ishika Singh, Valts Blukis, Arsalan Mousavian, Ankit Goyal, Danfei Xu, Jonathan Tremblay, Dieter Fox, Jesse Thomason, and Animesh Garg.
\newblock Progprompt: Generating situated robot task plans using large language models, 2022.
\newblock URL \url{https://arxiv.org/abs/2209.11302}.

\bibitem[Song et~al.(2023)Song, Wu, Washington, Sadler, Chao, and Su]{song2023llmplanner}
Chan~Hee Song, Jiaman Wu, Clayton Washington, Brian~M Sadler, Wei-Lun Chao, and Yu~Su.
\newblock Llm-planner: Few-shot grounded planning for embodied agents with large language models.
\newblock In \emph{Proceedings of the IEEE/CVF international conference on computer vision}, pp.\  2998--3009, 2023.

\bibitem[Song et~al.(2024)Song, Yin, Yue, Huang, Li, and Lin]{song2024trialanderror}
Yifan Song, Da~Yin, Xiang Yue, Jie Huang, Sujian Li, and Bill~Yuchen Lin.
\newblock Trial and error: Exploration-based trajectory optimization for llm agents.
\newblock \emph{arXiv preprint arXiv:2403.02502}, 2024.

\bibitem[Stiennon et~al.(2020)Stiennon, Ouyang, Wu, Ziegler, Lowe, Voss, Radford, Amodei, and Christiano]{stiennon2020learning}
Nisan Stiennon, Long Ouyang, Jeffrey Wu, Daniel Ziegler, Ryan Lowe, Chelsea Voss, Alec Radford, Dario Amodei, and Paul~F Christiano.
\newblock Learning to summarize with human feedback.
\newblock \emph{Advances in neural information processing systems}, 33:\penalty0 3008--3021, 2020.

\bibitem[Su \& Zhang(2023)Su and Zhang]{su2023subgoal}
Jianhai Su and Qi~Zhang.
\newblock Subgoal proposition using a vision-language model.
\newblock In \emph{CoRL 2023 Workshop on Learning Effective Abstractions for Planning (LEAP)}, 2023.

\bibitem[Sun(2023)]{sun2023reinforcement}
Hao Sun.
\newblock Reinforcement learning in the era of llms: What is essential? what is needed? an rl perspective on rlhf, prompting, and beyond.
\newblock \emph{arXiv preprint arXiv:2310.06147}, 2023.

\bibitem[Sun et~al.(2025)Sun, Shen, and Ton]{sun2025rethinking}
Hao Sun, Yunyi Shen, and Jean-Francois Ton.
\newblock Rethinking reward modeling in preference-based large language model alignment.
\newblock In \emph{The Thirteenth International Conference on Learning Representations}, 2025.

\bibitem[Sun et~al.(2024)Sun, Hong, Pala, Toh, Tan, Ghosal, Poria, et~al.]{sun2024emmax}
Qi~Sun, Pengfei Hong, Tej~Deep Pala, Vernon Toh, U~Tan, Deepanway Ghosal, Soujanya Poria, et~al.
\newblock Emma-x: An embodied multimodal action model with grounded chain of thought and look-ahead spatial reasoning.
\newblock \emph{arXiv preprint arXiv:2412.11974}, 2024.

\bibitem[Sun et~al.(2023{\natexlab{a}})Sun, Yu, Cui, Zhang, Zhang, Wang, Gao, Liu, Huang, and Wang]{sun2023emu}
Quan Sun, Qiying Yu, Yufeng Cui, Fan Zhang, Xiaosong Zhang, Yueze Wang, Hongcheng Gao, Jingjing Liu, Tiejun Huang, and Xinlong Wang.
\newblock Emu: Generative pretraining in multimodality.
\newblock \emph{arXiv preprint arXiv:2307.05222}, 2023{\natexlab{a}}.

\bibitem[Sun et~al.(2023{\natexlab{b}})Sun, Shen, Cao, Liu, Li, Shen, Gan, Gui, Wang, Yang, et~al.]{sun2023aligning}
Zhiqing Sun, Sheng Shen, Shengcao Cao, Haotian Liu, Chunyuan Li, Yikang Shen, Chuang Gan, Liang-Yan Gui, Yu-Xiong Wang, Yiming Yang, et~al.
\newblock Aligning large multimodal models with factually augmented rlhf.
\newblock \emph{arXiv preprint arXiv:2309.14525}, 2023{\natexlab{b}}.

\bibitem[Szot et~al.(2025)Szot, Mazoure, Attia, Timofeev, Agrawal, Hjelm, Gan, Kira, and Toshev]{szot2025multimodal}
Andrew Szot, Bogdan Mazoure, Omar Attia, Aleksei Timofeev, Harsh Agrawal, Devon Hjelm, Zhe Gan, Zsolt Kira, and Alexander Toshev.
\newblock From multimodal llms to generalist embodied agents: Methods and lessons.
\newblock In \emph{Proceedings of the Computer Vision and Pattern Recognition Conference}, pp.\  10644--10655, 2025.

\bibitem[Tang et~al.(2024)Tang, Lu, Wu, Xu, Ma, Fang, Guo, Lu, Chen, and Chen]{tang2024hawk}
Jiaqi Tang, Hao Lu, Ruizheng Wu, Xiaogang Xu, Ke~Ma, Cheng Fang, Bin Guo, Jiangbo Lu, Qifeng Chen, and Yingcong Chen.
\newblock Hawk: Learning to understand open-world video anomalies.
\newblock \emph{Advances in Neural Information Processing Systems}, 37:\penalty0 139751--139785, 2024.

\bibitem[Tian et~al.(2025)Tian, Wang, Guo, Wu, Dong, Wang, Yang, Zhang, Zhu, and Liu]{tian2025ego}
Shulin Tian, Ruiqi Wang, Hongming Guo, Penghao Wu, Yuhao Dong, Xiuying Wang, Jingkang Yang, Hao Zhang, Hongyuan Zhu, and Ziwei Liu.
\newblock Ego-r1: Chain-of-tool-thought for ultra-long egocentric video reasoning.
\newblock \emph{arXiv preprint arXiv:2506.13654}, 2025.

\bibitem[Wang* et~al.(2025)Wang*, Zhang*, Wang*, Gao*, Li*, Wang, Chen, Wan, Lu, Yang, Wang, Krishna, Wu, Fei-Fei, Choi, and Li]{wang2025vagen}
Kangrui Wang*, Pingyue Zhang*, Zihan Wang*, Yaning Gao*, Linjie Li*, Qineng Wang, Hanyang Chen, Chi Wan, Yiping Lu, Zhengyuan Yang, Lijuan Wang, Ranjay Krishna, Jiajun Wu, Li~Fei-Fei, Yejin Choi, and Manling Li.
\newblock Reinforcing visual state reasoning for multi-turn vlm agents, 2025.
\newblock URL \url{https://github.com/RAGEN-AI/VAGEN}.

\bibitem[Wang et~al.(2024)Wang, Zhang, Zhang, Hu, Li, Zhang, Li, Wu, Wang, and Hovy]{wang2024reinforcement}
Shuhe Wang, Shengyu Zhang, Jie Zhang, Runyi Hu, Xiaoya Li, Tianwei Zhang, Jiwei Li, Fei Wu, Guoyin Wang, and Eduard Hovy.
\newblock Reinforcement learning enhanced llms: A survey.
\newblock \emph{arXiv preprint arXiv:2412.10400}, 2024.

\bibitem[Wang et~al.(2025)Wang, Fei, Cheng, Zhang, Cai, Fu, and Qiu]{wang2025worldmodelingmakesbetter}
Siyin Wang, Zhaoye Fei, Qinyuan Cheng, Shiduo Zhang, Panpan Cai, Jinlan Fu, and Xipeng Qiu.
\newblock World modeling makes a better planner: Dual preference optimization for embodied task planning, 2025.
\newblock URL \url{https://arxiv.org/abs/2503.10480}.

\bibitem[Wei et~al.(2025)Wei, Yang, Xing, Shi, Lu, and Ye]{wei2025gtr}
Tong Wei, Yijun Yang, Junliang Xing, Yuanchun Shi, Zongqing Lu, and Deheng Ye.
\newblock Gtr: Guided thought reinforcement prevents thought collapse in rl-based vlm agent training.
\newblock \emph{arXiv preprint arXiv:2503.08525}, 2025.

\bibitem[Wu et~al.(2025{\natexlab{a}})Wu, Fan, Zang, Wang, Yin, Li, and Jin]{wu2025reinforced}
Di~Wu, Jiaxin Fan, Junzhe Zang, Guanbo Wang, Wei Yin, Wenhao Li, and Bo~Jin.
\newblock Reinforced reasoning for embodied planning.
\newblock \emph{arXiv preprint arXiv:2505.22050}, 2025{\natexlab{a}}.

\bibitem[Wu et~al.(2025{\natexlab{b}})Wu, Cheng, Yang, Zhang, Yang, Jiang, Mu, Peng, Qiao, Tan, et~al.]{wu2025gui}
Qianhui Wu, Kanzhi Cheng, Rui Yang, Chaoyun Zhang, Jianwei Yang, Huiqiang Jiang, Jian Mu, Baolin Peng, Bo~Qiao, Reuben Tan, et~al.
\newblock Gui-actor: Coordinate-free visual grounding for gui agents.
\newblock \emph{arXiv preprint arXiv:2506.03143}, 2025{\natexlab{b}}.

\bibitem[Wu et~al.(2023)Wu, Wang, Xu, Lu, and Yan]{wu2023embodied}
Zhenyu Wu, Ziwei Wang, Xiuwei Xu, Jiwen Lu, and Haibin Yan.
\newblock Embodied task planning with large language models.
\newblock \emph{arXiv preprint arXiv:2307.01848}, 2023.

\bibitem[Wu et~al.(2024)Wu, Wu, Xu, Wang, Sun, Jia, Cheng, Ding, Chen, Liang, et~al.]{wu2024atlas}
Zhiyong Wu, Zhenyu Wu, Fangzhi Xu, Yian Wang, Qiushi Sun, Chengyou Jia, Kanzhi Cheng, Zichen Ding, Liheng Chen, Paul~Pu Liang, et~al.
\newblock Os-atlas: A foundation action model for generalist gui agents.
\newblock \emph{arXiv preprint arXiv:2410.23218}, 2024.

\bibitem[Xu et~al.(2024)Xu, Wang, Wang, Lu, Xie, Saha, Sahoo, Yu, and Xiong]{xu2024aguvis}
Yiheng Xu, Zekun Wang, Junli Wang, Dunjie Lu, Tianbao Xie, Amrita Saha, Doyen Sahoo, Tao Yu, and Caiming Xiong.
\newblock Aguvis: Unified pure vision agents for autonomous gui interaction.
\newblock \emph{arXiv preprint arXiv:2412.04454}, 2024.

\bibitem[Yang et~al.(2024{\natexlab{a}})Yang, Dong, Liu, Li, Wang, Tan, Jiang, Kang, Zhang, Zhou, et~al.]{yang2024octopus}
Jingkang Yang, Yuhao Dong, Shuai Liu, Bo~Li, Ziyue Wang, Haoran Tan, Chencheng Jiang, Jiamu Kang, Yuanhan Zhang, Kaiyang Zhou, et~al.
\newblock Octopus: Embodied vision-language programmer from environmental feedback.
\newblock In \emph{European conference on computer vision}, pp.\  20--38. Springer, 2024{\natexlab{a}}.

\bibitem[Yang et~al.(2024{\natexlab{b}})Yang, Ding, Lin, Zhang, and Zhang]{yang2024regularizing}
Rui Yang, Ruomeng Ding, Yong Lin, Huan Zhang, and Tong Zhang.
\newblock Regularizing hidden states enables learning generalizable reward model for llms.
\newblock \emph{Advances in Neural Information Processing Systems}, 37:\penalty0 62279--62309, 2024{\natexlab{b}}.

\bibitem[Yang et~al.(2024{\natexlab{c}})Yang, Pan, Luo, Qiu, Zhong, Yu, and Chen]{yang2024rewards}
Rui Yang, Xiaoman Pan, Feng Luo, Shuang Qiu, Han Zhong, Dong Yu, and Jianshu Chen.
\newblock Rewards-in-context: Multi-objective alignment of foundation models with dynamic preference adjustment.
\newblock \emph{arXiv preprint arXiv:2402.10207}, 2024{\natexlab{c}}.

\bibitem[Yang et~al.(2025)Yang, Chen, Zhang, Zhao, Qian, Wang, Wang, Koripella, Movahedi, Li, et~al.]{yang2025embodiedbench}
Rui Yang, Hanyang Chen, Junyu Zhang, Mark Zhao, Cheng Qian, Kangrui Wang, Qineng Wang, Teja~Venkat Koripella, Marziyeh Movahedi, Manling Li, et~al.
\newblock Embodiedbench: Comprehensive benchmarking multi-modal large language models for vision-driven embodied agents.
\newblock \emph{arXiv preprint arXiv:2502.09560}, 2025.

\bibitem[Yao et~al.(2023)Yao, Zhao, Yu, Du, Shafran, Narasimhan, and Cao]{yao2023react}
Shunyu Yao, Jeffrey Zhao, Dian Yu, Nan Du, Izhak Shafran, Karthik Narasimhan, and Yuan Cao.
\newblock React: Synergizing reasoning and acting in language models.
\newblock In \emph{International Conference on Learning Representations (ICLR)}, 2023.

\bibitem[Yin et~al.(2024)Yin, Fu, Zhao, Li, Sun, Xu, and Chen]{yin2024survey}
Shukang Yin, Chaoyou Fu, Sirui Zhao, Ke~Li, Xing Sun, Tong Xu, and Enhong Chen.
\newblock A survey on multimodal large language models.
\newblock \emph{National Science Review}, 11\penalty0 (12), 2024.

\bibitem[Yu et~al.(2024)Yu, Yao, Zhang, He, Han, Cui, Hu, Liu, Zheng, Sun, et~al.]{yu2024rlhf}
Tianyu Yu, Yuan Yao, Haoye Zhang, Taiwen He, Yifeng Han, Ganqu Cui, Jinyi Hu, Zhiyuan Liu, Hai-Tao Zheng, Maosong Sun, et~al.
\newblock Rlhf-v: Towards trustworthy mllms via behavior alignment from fine-grained correctional human feedback.
\newblock In \emph{Proceedings of the IEEE/CVF Conference on Computer Vision and Pattern Recognition}, pp.\  13807--13816, 2024.

\bibitem[Yuan et~al.(2025)Yuan, Cui, Huang, Chen, Ni, Dong, Li, Zheng, and Hao]{yuan2025embodiedr1}
Yifu Yuan, Haiqin Cui, Yaoting Huang, Yibin Chen, Fei Ni, Zibin Dong, Pengyi Li, Yan Zheng, and Jianye Hao.
\newblock Embodied-r1: Reinforced embodied reasoning for general robotic manipulation.
\newblock \emph{arXiv preprint arXiv:2508.13998}, 2025.

\bibitem[Yuan et~al.(2024)Yuan, Liu, Cao, Sun, Jia, Chen, Li, Lin, Yuan, He, et~al.]{yuan2024mora}
Zhengqing Yuan, Yixin Liu, Yihan Cao, Weixiang Sun, Haolong Jia, Ruoxi Chen, Zhaoxu Li, Bin Lin, Li~Yuan, Lifang He, et~al.
\newblock Mora: Enabling generalist video generation via a multi-agent framework.
\newblock \emph{arXiv preprint arXiv:2403.13248}, 2024.

\bibitem[Zawalski et~al.(2024)Zawalski, Chen, Pertsch, Mees, Finn, and Levine]{zawalski2024roboticcontrol}
Micha{\l} Zawalski, William Chen, Karl Pertsch, Oier Mees, Chelsea Finn, and Sergey Levine.
\newblock Robotic control via embodied chain-of-thought reasoning.
\newblock \emph{arXiv preprint arXiv:2407.08693}, 2024.

\bibitem[Zhai et~al.(2024)Zhai, Bai, Lin, Pan, Tong, Zhou, Suhr, Xie, LeCun, Ma, et~al.]{zhai2024finetuning}
Simon Zhai, Hao Bai, Zipeng Lin, Jiayi Pan, Peter Tong, Yifei Zhou, Alane Suhr, Saining Xie, Yann LeCun, Yi~Ma, et~al.
\newblock Fine-tuning large vision-language models as decision-making agents via reinforcement learning.
\newblock \emph{Advances in neural information processing systems}, 37:\penalty0 110935--110971, 2024.

\bibitem[Zhang et~al.(2025{\natexlab{a}})Zhang, Li, Cheng, Hu, Yuan, Chen, Leng, Jiang, Zhang, Li, et~al.]{zhang2025videollama}
Boqiang Zhang, Kehan Li, Zesen Cheng, Zhiqiang Hu, Yuqian Yuan, Guanzheng Chen, Sicong Leng, Yuming Jiang, Hang Zhang, Xin Li, et~al.
\newblock Videollama 3: Frontier multimodal foundation models for image and video understanding.
\newblock \emph{arXiv preprint arXiv:2501.13106}, 2025{\natexlab{a}}.

\bibitem[Zhang et~al.(2024{\natexlab{a}})Zhang, Zhoubian, Hu, Yue, Dong, and Tang]{zhang2024rest}
Dan Zhang, Sining Zhoubian, Ziniu Hu, Yisong Yue, Yuxiao Dong, and Jie Tang.
\newblock Rest-mcts*: Llm self-training via process reward guided tree search.
\newblock \emph{Advances in Neural Information Processing Systems}, 37:\penalty0 64735--64772, 2024{\natexlab{a}}.

\bibitem[Zhang et~al.(2025{\natexlab{b}})Zhang, Guo, Hu, Chen, Zhu, and Chen]{zhang2025up}
Jianke Zhang, Yanjiang Guo, Yucheng Hu, Xiaoyu Chen, Xiang Zhu, and Jianyu Chen.
\newblock Up-vla: A unified understanding and prediction model for embodied agent.
\newblock \emph{arXiv preprint arXiv:2501.18867}, 2025{\natexlab{b}}.

\bibitem[Zhang et~al.(2024{\natexlab{b}})Zhang, Huang, Jin, and Lu]{zhang2024vision}
Jingyi Zhang, Jiaxing Huang, Sheng Jin, and Shijian Lu.
\newblock Vision-language models for vision tasks: A survey.
\newblock \emph{IEEE transactions on pattern analysis and machine intelligence}, 46\penalty0 (8):\penalty0 5625--5644, 2024{\natexlab{b}}.

\bibitem[Zhang et~al.(2024{\natexlab{c}})Zhang, Hu, Xu, Yan, Xu, Jin, Zhang, and Huang]{zhang2024tinychart}
Liang Zhang, Anwen Hu, Haiyang Xu, Ming Yan, Yichen Xu, Qin Jin, Ji~Zhang, and Fei Huang.
\newblock Tinychart: Efficient chart understanding with visual token merging and program-of-thoughts learning.
\newblock \emph{arXiv preprint arXiv:2404.16635}, 2024{\natexlab{c}}.

\bibitem[Zhang et~al.(2024{\natexlab{d}})Zhang, Jiang, Zhang, Lin, Guo, Qiu, Zhou, Lu, Chang, Qiao, et~al.]{zhang2024mathverse}
Renrui Zhang, Dongzhi Jiang, Yichi Zhang, Haokun Lin, Ziyu Guo, Pengshuo Qiu, Aojun Zhou, Pan Lu, Kai-Wei Chang, Yu~Qiao, et~al.
\newblock Mathverse: Does your multi-modal llm truly see the diagrams in visual math problems?
\newblock In \emph{European Conference on Computer Vision}, pp.\  169--186. Springer, 2024{\natexlab{d}}.

\bibitem[Zhang et~al.(2024{\natexlab{e}})Zhang, Wei, Jiang, Guo, Li, Zhang, Tong, Liu, Zhou, Wei, et~al.]{zhang2024mavis}
Renrui Zhang, Xinyu Wei, Dongzhi Jiang, Ziyu Guo, Shicheng Li, Yichi Zhang, Chengzhuo Tong, Jiaming Liu, Aojun Zhou, Bin Wei, et~al.
\newblock Mavis: Mathematical visual instruction tuning with an automatic data engine.
\newblock \emph{arXiv preprint arXiv:2407.08739}, 2024{\natexlab{e}}.

\bibitem[Zhang et~al.(2025{\natexlab{c}})Zhang, Wang, Liu, Huixin, Jiang, Shen, Hou, Zheng, Zhang, Li, et~al.]{zhang2025embodied}
Wenqi Zhang, Mengna Wang, Gangao Liu, Xu~Huixin, Yiwei Jiang, Yongliang Shen, Guiyang Hou, Zhe Zheng, Hang Zhang, Xin Li, et~al.
\newblock Embodied-reasoner: Synergizing visual search, reasoning, and action for embodied interactive tasks.
\newblock \emph{arXiv preprint arXiv:2503.21696}, 2025{\natexlab{c}}.

\bibitem[Zhao et~al.(2025)Zhao, Lu, Kim, Fu, Zhang, Wu, Li, Ma, Han, Finn, et~al.]{zhao2025cot}
Qingqing Zhao, Yao Lu, Moo~Jin Kim, Zipeng Fu, Zhuoyang Zhang, Yecheng Wu, Zhaoshuo Li, Qianli Ma, Song Han, Chelsea Finn, et~al.
\newblock Cot-vla: Visual chain-of-thought reasoning for vision-language-action models.
\newblock In \emph{Proceedings of the Computer Vision and Pattern Recognition Conference}, pp.\  1702--1713, 2025.

\bibitem[Zheng et~al.(2025)Zheng, Liu, Li, Chen, Yu, Gao, Dang, Liu, Men, Yang, et~al.]{zheng2025group}
Chujie Zheng, Shixuan Liu, Mingze Li, Xiong-Hui Chen, Bowen Yu, Chang Gao, Kai Dang, Yuqiong Liu, Rui Men, An~Yang, et~al.
\newblock Group sequence policy optimization.
\newblock \emph{arXiv preprint arXiv:2507.18071}, 2025.

\bibitem[Zheng et~al.(2022)Zheng, Chen, Jenkins, and Wang]{zheng2022vlmbench}
Kaizhi Zheng, Xiaotong Chen, Odest~Chadwicke Jenkins, and Xin~Eric Wang.
\newblock {VLMbench}: A compositional benchmark for vision-and-language manipulation.
\newblock In \emph{Proceedings of the NeurIPS Datasets and Benchmarks Track}, 2022.

\bibitem[Zheng et~al.(2023)Zheng, He, and Wang]{zheng2023minigpt}
Kaizhi Zheng, Xuehai He, and Xin~Eric Wang.
\newblock Minigpt-5: Interleaved vision-and-language generation via generative vokens.
\newblock \emph{arXiv preprint arXiv:2310.02239}, 2023.

\bibitem[Zhu et~al.(2025)Zhu, Wang, Chen, Liu, Ye, Gu, Duan, Tian, Su, Shao, et~al.]{zhu2025internvl3}
Jinguo Zhu, Weiyun Wang, Zhe Chen, Zhaoyang Liu, Shenglong Ye, Lixin Gu, Yuchen Duan, Hao Tian, Weijie Su, Jie Shao, et~al.
\newblock Internvl3: Exploring advanced training and test-time recipes for open-source multimodal models.
\newblock \emph{arXiv preprint arXiv:2504.10479}, 2025.

\bibitem[Zou et~al.()Zou, Guo, Yang, Zhang, Hu, and Zhang]{zoudynamath}
Chengke Zou, Xingang Guo, Rui Yang, Junyu Zhang, Bin Hu, and Huan Zhang.
\newblock Dynamath: A dynamic visual benchmark for evaluating mathematical reasoning robustness of vision language models.
\newblock In \emph{The Thirteenth International Conference on Learning Representations}.

\end{thebibliography}
\bibliographystyle{arxiv}


\appendix
\newpage
\section*{Appendix Contents}

\vspace{0.3em}

{\large
  \setlength{\parskip}{0.55em}
  \setlength{\parindent}{0pt}

  \noindent Limitation \dotfill \pageref{app:limitation}\par
  \noindent Related Work \dotfill \pageref{app:related_work}\par
  \noindent Environment Details \dotfill \pageref{app:environment}\par
  \noindent Implementation Details \dotfill \pageref{app:implementation_details}\par
  \noindent Additional Experiment \dotfill \pageref{app:additional_experiments}\par
  \noindent Agent Prompt \dotfill \pageref{app:agent_prompt}\par
  \noindent Embodied Prior Learning Dataset \dotfill \pageref{app:prior_dataset}\par
  \noindent Error Analysis \dotfill \pageref{app:error_analysis}\par
  \noindent Case Analysis \dotfill \pageref{app:case_analysis}\par
}
\newpage


\section{Limitation}
\label{app:limitation}
A key limitation of this work is that all evaluations are conducted in simulated environments, without validation on real-world systems. This reflects a common trade-off in agent research: simulations provide standardized and reproducible benchmarks that greatly reduce time, cost, and safety risks, but they inevitably limit real-world applicability. While such practice is common for LLM/VLM-based agents \citep{zhai2024finetuning,feng2025gigpo}, real-world testing remains crucial for practical deployment. As future work, we plan to explore deploying our ERA training pipelines in real-world environments.

\section{Additional Related Work}
\label{app:related_work}
\textbf{Vision Language Models.}
Vision Language Models (VLMs) have been a popular research domain with their ability to combine multi-modal perception with a strong language backbone. \cite{li2025visual} and other works \citep{yin2024survey,zhang2024vision,awais2025foundation} categorize VLMs into subdomains like vision to text \citep{chen2024expanding,deitke2024molmo,bai2025qwen25vl,li2024llava,li2024mini,zhang2024tinychart}, vision to action \citep{sima2024drivelm}, and text to vision \citep{deng2025emerging,chen2025janus,sun2023emu,zheng2023minigpt}, etc. For vision to action, embodied AI serves as a perfect area, as it provides a natural environment and action interface. \cite{kim2024openvla,huang2023voxposer,xu2024aguvis,wu2024atlas,wu2025gui} pioneer the relevant field by converting vision input into executable actions. VLMs for reasoning comprise a large portion of the suitable applications. For example, existing works \citep{zhang2024mathverse,lu2023mathvista,zhang2024mavis,zoudynamath} utilize VLMs for solving math problems. Besides, video models emerge as an extension to VLMs as well. \cite{chen2025eagle}, \cite{zhang2025videollama} and \cite{tian2025ego} synthesize video information into text for further processing. The interleaving interaction between vision information and text reasoning provides better deployment for VLMs \citep{yuan2024mora,tang2024hawk}.

\textbf{RL for LLMs or VLMs.} Reinforcement Learning from Human Feedback (RLHF) has become a cornerstone of modern LLM alignment \citep{sun2023reinforcement,wang2024reinforcement}. Early work such as InstructGPT established the paradigm of training a reward model from human preferences and using RL to fine-tune the base model, demonstrating substantial improvements in helpfulness and safety \citep{ouyang2022training,stiennon2020learning,yang2024rewards}. In such framework, reward model \citep{sun2025rethinking,yang2024regularizing,luo2025rethinking,shen2025micro} plays a critical role to modeling human preference and guide policy optimization. To avoid explicitly training a reward model, implicit reward models like Direct Preference Optimization (DPO \citep{rafailov2023direct}) have been investigated and successfully applied in scenarios with preference data. Since the success of OpenAI-o1 \citep{jaech2024openai} and DeepSeek-R1 \citep{guo2025deepseek}, reinforcement learning (RL) has been a dominant technique for LLMs post-training, especially for reasoning models. To better handle long autoregressive sequences, new algorithms like GRPO and GSPO \citep{guo2025deepseek,shao2024deepseekmath,zheng2025group} improve stability via group-based comparisons, and offline/self-training methods such as ReST and ReST-MCTS \citep{gulcehre2023reinforced,zhang2024rest} reduce online interaction costs by iteratively filtering and retraining on high-quality outputs. Beyond textual alignment, RL is increasingly applied to reasoning \citep{lightman2023let} and multimodal alignment: LLaVA-RLHF and RLHF-V \citep{sun2023aligning,yu2024rlhf} demonstrate that preference optimization can mitigate hallucinations and strengthen grounding in visual-language tasks.
\section{Environment Details}
\label{app:environment}

\noindent In this section, we describe the two evaluation environments used in our study. This complements Section~\ref{sec:method} by detailing the task interfaces, action spaces, observations, and simulator feedback. The environments span high-level planning and low-level control: \textbf{EB-ALFRED}, a household-task benchmark in AI2-THOR with PDDL-specified subgoals and a discrete action set; and \textbf{EB-Manipulation}, a CoppeliaSim-based robotic control suite that issues 7-DoF end-effector commands with automatic motion planning.

\subsection{EB-ALFRED for High-level Planning Task} \label{ap:details_alfred}
EB-ALFRED is built on the ALFRED dataset and the AI2-THOR simulator~\citep{Shridhar_2020_CVPR,kolve2017ai2thor,deitke2020robothor}, widely recognized for diverse household tasks and realistic environments in embodied AI. The benchmark evaluates an agent’s ability to plan and execute sequences of high-level actions for tasks such as ``Put washed lettuce in the refrigerator.'' Each task is formally represented in the Planning Domain Definition Language (PDDL)~\citep{mcdermott1998pddl}, enabling precise evaluation of task and subgoal completion.

The ALFRED dataset spans seven task types: \textit{Pick} \& \textit{Place}, \textit{Stack} \& \textit{Place}, \textit{Pick Two} \& \textit{Place}, \textit{Clean} \& \textit{Place}, \textit{Heat} \& \textit{Place}, \textit{Cool} \& \textit{Place}, and \textit{Examine in Light}. Following LoTa-Bench’s implementation for household task planning~\citep{choi2024lotabench}, the simulator supports eight high-level action primitives: \textit{pick up}, \textit{open}, \textit{close}, \textit{turn on}, \textit{turn off}, \textit{slice}, \textit{put down}, and \textit{find}. These actions are parameterized by objects (e.g., ``find an apple'' or ``pick up an apple''). The simulator provides both egocentric visual observations and textual feedback, indicating whether an action succeeds or fails (e.g., ``failure to pick up an object because another object is already being held'').

\subsection{EB-Manipulation for Low-level Control Task} \label{ap:details_manip}
EB-Manipulation extends VLMbench~\citep{zheng2022vlmbench} using the CoppeliaSim/V-REP simulator~\citep{rohmer2013v} to control a 7-DoF Franka Emika Panda robotic arm. It comprises four manipulation categories: (1) \textit{Pick} \& \textit{Place Objects}, (2) \textit{Stack Objects}, (3) \textit{Shape Sorter Placement}, and (4) \textit{Table Wiping}, each with diverse instances varying in color, position, shape, and orientation.

The action space is a 7-dimensional vector specifying end-effector translation, rotation, and gripper state. Actions are executed with automatic motion planning in \texttt{ABS\_EE\_POSE\_PLAN\_WORLD\_FRAME} mode, which drives the trajectory from the current to the target pose, reducing the agent’s burden to predicting keypoints essential for task completion.

\section{Implementation Details}
\label{app:implementation_details}

This section provides a comprehensive overview of the implementation details for our proposed ERA framework, complementing the methodology presented in Section~\ref{sec:method} and supporting the reproducibility of the experimental results reported in Section~\ref{sec:exp}. We structure this section into two main parts. First, we elaborate on the \textbf{Algorithm Details}, where we discuss the design rationale behind our EPL training curriculum (\S\ref{app:epl_design}) and provide a detailed breakdown of the reward function used in our RL pipeline (\S\ref{app:reward_design}). Second, we present the \textbf{Training Details}, which includes a complete list of hyperparameters and configurations for both the Embodied Prior Learning stage (\S\ref{app:EPL_config}) and the online Reinforcement Learning stage (\S\ref{app:training_details:rl_details}).

\subsection{Algorithm Details}

\subsubsection{EPL Training Design Details}
\label{app:epl_design}
The sequential training order in EPL is deliberately designed to reflect the hierarchical nature of embodied skill acquisition. Since small VLMs lack environment-grounded perception after generic pretraining, we first fine-tune them on \textbf{environment-anchored priors} to establish spatial grounding, object semantics, and action–context associations tied to the embodied environment. This early stage provides the model with a stable perceptual foundation for interpreting observations and reasoning about spatial / semantic relations. Next, we train on \textbf{trajectory-augmented priors}, which build upon this foundation by injecting step-level reasoning, temporal coherence, and goal-conditioned planning skills. This stage allows the model to integrate grounded perception with structured reasoning traces, thereby enhancing its ability to perform long-horizon tasks. When \textbf{external knowledge priors} are included, they are placed before trajectory-level training to transfer abstract reasoning and cross-domain grounding capabilities from large-scale text or multimodal corpora. Together, this curriculum aligns with the goal of bridging the perception–reasoning–action gap, ensuring that embodied agents acquire both generalizable and task-specific priors before online interaction.

The configurations of the EPL methods in Table~\ref{tab:EPL_ablations} are summarized as follows:
\begin{itemize}
    \item \textbf{Raw Trajectory}: trained on the raw trajectory dataset for 2 epochs.
    \item \textbf{Raw Trajectory + Traj-Aug}: trained on the trajectory-augmented prior dataset for 2 epochs.
    \item \textbf{Raw Trajectory + Env-Anc}: trained first on the environment-anchored prior dataset for 1 epoch, followed by the raw trajectory dataset for 2 epochs.
    \item \textbf{Raw Trajectory + Ext-Know}: trained first on the external knowledge prior dataset for 1 epoch, followed by the raw trajectory dataset for 2 epochs.
    \item \textbf{Raw Trajectory + Traj-Aug + Env-Anc}: trained first on the environment-anchored prior dataset for 1 epoch, followed by the trajectory-augmented prior dataset for 2 epochs.
    \item \textbf{Raw Trajectory + Traj-Aug + Ext-Know}: trained first on the external knowledge prior dataset for 1 epoch, followed by the trajectory-augmented prior dataset for 2 epochs.
\end{itemize}



\subsubsection{RL Reward Design Details}
\label{app:reward_design}

To provide a dense and informative learning signal that balances final task completion with intermediate progress and behavior shaping, we design the reward function $r_t$ at each turn $t$ as a sum of three components:
\[
r_t = r_t^{\text{success}} + r_t^{\text{subgoal}} + r_t^{\text{behavior}}.
\]
The specific values for these components are summarized in Table~\ref{tab:reward_hyperparams}. Below, we provide a detailed breakdown of each component with examples and implementation details.

\paragraph{Reward Hyperparameters.}
The numerical values for each reward component are detailed in Table~\ref{tab:reward_hyperparams}. These values were determined through empirical tuning to balance the different learning objectives.

\begin{table}[h]
\centering
\caption{Hyperparameters for the reward components.}
\label{tab:reward_hyperparams}
\begin{tabular}{@{}lll@{}}
\toprule
Component & Task Type & Value \\
\midrule
\multirow{2}{*}{Success ($r_t^{\text{success}}$)} & High-level (EB-ALFRED) & $+4.0$ \\
& Low-level (EB-Manipulation) & $+3.0$ \\
\midrule
\multirow{2}{*}{Subgoal ($r_t^{\text{subgoal}}$)} & High-level (EB-ALFRED) & $+1.0$ per new subgoal \\
& Low-level (EB-Manipulation) & $+1.0$ per new object approached \\
\midrule
\multirow{2}{*}{Behavior Shaping ($r_t^{\text{behavior}}$)} & High-level (EB-ALFRED) & $-0.5$ for invalid actions \\
& Low-level (EB-Manipulation) & $+0.5$ if $q_t > 0.75$; $-0.5$ if $q_t < 0.25$ \\
\bottomrule
\end{tabular}
\end{table}

\begin{enumerate}[label=(\roman*), itemsep=0.2em, topsep=0.3em]

\item \textbf{Success-based Reward} ($r_t^{\text{success}}$): A sparse, high-magnitude reward is given upon task completion to serve as the primary optimization objective.
\begin{itemize}[leftmargin=*, itemsep=0.1em]
    \item \textit{For high-level planning}, a reward of $r_t^{\text{success}} = +4.0$ is awarded if the task's goal conditions are met. The episode then terminates. For example, for the instruction ``wash the apple and put it in the refrigerator,'' the agent receives this reward only when the environment state confirms that the apple's property is ``isWashed'' and its location is inside the ``refrigerator'' receptacle.
    \item \textit{For low-level manipulation}, a reward of $r_t^{\text{success}} = +3.0$ is given upon successful completion. For instance, if the instruction is to ``stack block A on block B,'' the reward is granted when the environment's physics engine determines that block A is in a stable state on top of block B, satisfying the goal constraints.
\end{itemize}

\item \textbf{Subgoal-based Reward} ($r_t^{\text{subgoal}}$): This component provides a dense signal for achieving intermediate steps, guiding the agent's exploration.
\begin{itemize}[leftmargin=*, itemsep=0.1em]
  \item \textit{For high-level planning}, the environment defines a set of subgoals that must be completed. The agent receives a reward of $r_t^{\text{subgoal}} = +1.0$ each time it achieves a new, previously uncompleted subgoal. For example, for the task ``wash the apple and put it in the refrigerator,'' a key subgoal is changing the apple's state to ``isWashed". When the agent successfully executes the ``wash" action on the apple, it receives a +1.0 reward for completing this subgoal for the first time.
  \item \textit{For low-level manipulation}, we maintain a set of target objects $\mathcal{O}_{\text{target}}$ relevant to the task. The agent is rewarded with $r_t^{\text{subgoal}} = +1.0$ the first time its end-effector $e_t$ enters the vicinity of a target object $o \in \mathcal{O}_{\text{target}}$ at position $p$. This is determined by checking if $\|e_t - p\|_2 < \delta$, where $\delta$ is a small distance threshold. To encourage exploration, this reward is granted only once per unique target object within an episode. The logic is detailed in Algorithm~\ref{alg:subgoal_reward}.
\end{itemize}

\item \textbf{Behavior Shaping Reward} ($r_t^{\text{behavior}}$): This component penalizes incorrect behavior and rewards correctness at the domain-specific level.
\begin{itemize}[leftmargin=*, itemsep=0.1em]
  \item \textit{For high-level planning}, flawed reasoning can lead to semantically invalid actions. Such actions incur a penalty of $r_t^{\text{behavior}} = -0.5$. These invalid actions are defined by the environment's logical constraints. Examples include:
    \begin{itemize}
        \item Attempting to \texttt{Pickup} an object when another is already held.
        \item Attempting to \texttt{Put} an object in a receptacle when not holding that object.
        \item Attempting to \texttt{Open} a receptacle that is already open.
        \item Interacting with an object that is not currently visible.
    \end{itemize}
    This penalty discourages the agent from taking illogical or impossible actions, thereby improving the coherence of its plans.

  \item \textit{For low-level manipulation}, precise control requires accurate visual perception. We reward or penalize the agent based on the quality of its generated visual description. Let the environment contain $N$ objects ordered from left to right, with ground-truth attributes (type, color) given by a sequence of tuples $(d_1, \dots, d_N)$. If the agent's description yields predicted tuples $(\hat{d}_1, \dots, \hat{d}_N)$, we define the matching ratio as $q_t = \frac{1}{N} \sum_{i=1}^N \mathbf{1}\{\hat{d}_i = d_i\}$. If the agent fails to generate a parsable description, $q_t$ is set to 0 to prevent the agent from omitting the description to avoid penalties. The reward is then assigned based on this ratio:
  \[
    r_t^{\text{behavior}} =
    \begin{cases}
      +0.5 & \text{if } q_t > 0.75 \\
      -0.5 & \text{if } q_t < 0.25 \\
      0 & \text{otherwise}
    \end{cases}
  \]
  This reward structure incentivizes the agent to develop robust and accurate visual perception skills. The calculation is detailed in Algorithm~\ref{alg:visual_reward}.
\end{itemize}

\end{enumerate}

\begin{algorithm}[h]
\caption{Pseudocode for Low-Level Subgoal Reward}
\label{alg:subgoal_reward}
\begin{algorithmic}[1]
\State \textbf{Input:} observation, state dictionary target\_objects\_approached
\Procedure{CheckTargetObjectsApproached}{observation, target\_objects\_approached}
\State \textbf{if} gripper\_pose not in observation \textbf{then return} False
\State gripper\_coords $\leftarrow$ observation.gripper\_pose[:3]
\State
\For{each obj\_name, status in target\_objects\_approached}
\If{status == 0} \Comment{Only check un-approached objects}
\State obj\_info $\leftarrow$ observation.object\_informations[obj\_name]
\State obj\_coords $\leftarrow$ obj\_info.pose[:3]
\State distance $\leftarrow$ $\Vert\text{gripper\_coords} - \text{obj\_coords}\Vert_2$
\If{distance $\leq 0.2$}
\State target\_objects\_approached[obj\_name] $\leftarrow$ 1 \Comment{Mark as approached}
\State \textbf{return} True \Comment{New subgoal achieved}
\EndIf
\EndIf
\EndFor
\State \textbf{return} False \Comment{No new target object was approached}
\EndProcedure
\end{algorithmic}
\end{algorithm}

\begin{algorithm}[t]
\caption{Pseudocode for Low-Level Visual Description Reward}
\label{alg:visual_reward}
\begin{algorithmic}[1]
\State \textbf{Input:} Agent's reasoning output think\_text, Environment observation
\State
\Procedure{GetGroundTruthObjects}{observation}
    \State objects\_with\_y $\leftarrow$ []
    \For{obj\_name, obj\_info in observation.object\_informations}
        \State y\_coord $\leftarrow$ obj\_info.pose[1] \Comment{Extract y-coordinate for sorting}
        \State type, color $\leftarrow$ GetProperties(obj\_name)
        \State Add (y\_coord, type, color) to objects\_with\_y
    \EndFor
    \State Sort objects\_with\_y by y\_coord
    \State \textbf{return} list of (type, color) tuples from sorted list
\EndProcedure
\State
\Procedure{ParseVisualDescription}{think\_text}
    \State desc\_text $\leftarrow$ Extract visual description section from think\_text using regex
    \If{desc\_text is empty} \textbf{return} None \EndIf
    \State parsed\_objects $\leftarrow$ []
    \State matches $\leftarrow$ Find all object patterns (e.g., ``a red cube at [...]") in desc\_text
    \For{each match in matches}
        \State words $\leftarrow$ Split match into words
        \State color, type $\leftarrow$ IdentifyColorAndType(words)
        \State Add (color, type) to parsed\_objects
    \EndFor
    \State \textbf{return} parsed\_objects
\EndProcedure
\State
\Procedure{CalculateVisualReward}{think\_text, observation}
    \State predicted\_tuples $\leftarrow$ ParseVisualDescription(think\_text)
    \If{predicted\_tuples is None} \textbf{return} -0.5 \EndIf \Comment{Penalize unparsable description}
    \State
    \State gt\_tuples $\leftarrow$ GetGroundTruthObjects(observation)
    \State N $\leftarrow$ length of gt\_tuples
    \If{N == 0} \textbf{return} 0 \EndIf
    \State
    \State match\_count $\leftarrow$ 0
    \For{i from 0 to min(len(gt\_tuples), len(predicted\_tuples)) - 1}
        \If{predicted\_tuples[i] matches gt\_tuples[i]}
            \State match\_count $\leftarrow$ match\_count + 1
        \EndIf
    \EndFor
    \State
    \State $q\_t$ $\leftarrow$ match\_count / N
    \State
    \If{$q\_t > 0.75$} \textbf{return} +0.5
    \ElsIf{$q\_t < 0.25$} \textbf{return} -0.5
    \Else \textbf{return} 0
    \EndIf
\EndProcedure
\end{algorithmic}
\end{algorithm}

\paragraph{Hyperparameter Considerations.}
The relative magnitudes of the reward components are crucial for effective training.
\begin{itemize}
  \item $r_t^{\text{success}}$ should be larger than any potential cumulative reward from other components to ensure task completion remains the primary goal.
  \item $r_t^{\text{subgoal}}$ controls the incentive for making intermediate progress. Its magnitude should be significant enough to guide exploration but not so large as to create local optima where the agent is satisfied with only completing subgoals.
  \item The penalties for invalid actions and poor visual descriptions should be calibrated to discourage undesired behaviors without making the agent overly risk-averse, which could stifle exploration.
  \item The bonuses for accurate descriptions should provide a meaningful incentive but not dominate the subgoal or success rewards.
\end{itemize}
A careful tuning of these components is necessary to achieve a balance between exploration, behavior shaping, and convergence to successful policies.

\subsection{Training Details}
\label{app:training_details}
\subsubsection{Embodied Prior Learning}
\label{app:EPL_config}
For ERA with the Qwen2.5-VL-3B backbone, we set the input resolution to $500 \times 500$ pixels to balance performance and efficiency, and freeze the vision transformer (ViT) to preserve pre-trained visual representations. The maximum sequence length is 8,192 tokens. Training uses the Adam optimizer \citep{kingma2014adam} with a cosine learning rate schedule and a warm-up ratio of 5\%.

We adopt Embodied Prior Learning (EPL) with a batch size of 16 per dataset. The peak learning rate is $1 \times 10^{-5}$. Our implementation builds upon the AGUVIS framework \citep{xu2024aguvis} and incorporates DeepSpeed optimizations \citep{rajbhandari2020}, BF16 mixed precision, and gradient checkpointing to reduce memory usage.

The EPL-only variant is trained on a cluster of H200-140G GPUs, where the 3B model uses 2 nodes and completes training in approximately 2 hours for EB-Manipulation and 5 hours for EB-ALFRED.

\subsubsection{Reinforcement Learning}
\label{app:training_details:rl_details}

Our online reinforcement learning stage is implemented using a custom framework based on VeRL \citep{sheng2025hybridflow}, tailored for training VLM-based embodied agents. We employ the Proximal Policy Optimization (PPO) algorithm. A key feature of our framework is its ability to perform large-scale parallel rollouts, where multiple agents interact with distinct environment instances simultaneously to accelerate data collection. The following subsections detail the hyperparameters and training procedures for both high-level (EB-ALFRED) and low-level (EB-Manipulation) tasks.

\paragraph{Batching Strategy.}
A crucial aspect of our training setup is the distinction between data collection batching and gradient update batching.
\begin{itemize}[leftmargin=*]
    \item \textbf{Rollout Batch Size} refers to the number of parallel environments used for data collection in each rollout phase. For the high-level EB-ALFRED task, we use a rollout batch size of 50. For the low-level EB-Manipulation task, we use 48 parallel environments. Each environment instance generates a trajectory of up to 30 turns for EB-ALFRED and 15 turns for EB-Manipulation.
    \item \textbf{PPO Mini-Batch and Micro-Batch Size.} During the update phase, the trajectory data collected from all parallel rollouts is aggregated. From this buffer, we sample PPO mini-batches of 16 turn-level experiences. For distributed training, this mini-batch is further divided into micro-batches. For both tasks, we set the per-GPU micro-batch size to 1, meaning each GPU processes one turn-level sample at a time for gradient computation.
\end{itemize}

\paragraph{Policy and Value Network Optimization.}
For both tasks, the actor (policy) and critic (value) networks are initialized from the weights of the model obtained after the Embodied Prior Learning stage. However, optimization details differ significantly between the high-level and low-level tasks to reflect their distinct challenges.

\begin{itemize}[leftmargin=*]
\item For \textbf{high-level planning (EB-ALFRED)}, we use an AdamW optimizer with a learning rate of $1 \times 10^{-6}$ for the actor and $1 \times 10^{-5}$ for the critic. Throughout the RL stage, the vision tower of the VLM is kept frozen. This encourages the agent to learn high-level reasoning and planning capabilities based on fixed visual features, as the task depends more on symbolic understanding than on fine-tuning perceptual abilities.

\item For \textbf{low-level manipulation (EB-Manipulation)}, the actor learning rate is set to $6 \times 10^{-7}$ and the critic learning rate to $1 \times 10^{-5}$. In contrast to the high-level task, we unfreeze and fine-tune the vision tower for both the actor and the critic. This is critical for low-level control, which demands precise spatial understanding and grounding that can be refined through online interaction with the environment.
\end{itemize}

\paragraph{PPO Hyperparameters.}
Our PPO implementation is built upon the turn-wise advantage estimation described in Section~\ref{sec:turnwise_rl_main}. Key hyperparameters were configured as follows for both high-level and low-level tasks. The discount factor was set to $\gamma = 0.99$ and the GAE parameter to $\lambda = 0.99$, placing a slight emphasis on near-term rewards while still accounting for long-term consequences. During policy updates, we used a clipping ratio of $\epsilon = 0.2$ for the PPO objective. The value function loss was also clipped with a range of $0.5$. For each batch of rollout data, we performed a single update epoch ($N_{epochs} = 1$). To encourage exploration and prevent policy collapse, we added an entropy bonus to the actor's loss, with a coefficient of $0.001$. Gradient clipping was applied with a norm of $1.0$ for both the actor and critic to ensure stable training. While our framework supports KL-divergence regularization against the initial SFT policy to prevent large policy deviations, this feature was disabled in our final experiments.

\paragraph{Training Procedure.}
The online training process is organized into PPO iterations, each consisting of a rollout phase and an update phase. We run a total of 15 PPO iterations for EB-ALFRED and 50 iterations for EB-Manipulation. To ensure that value estimates are reliable before they are used to compute advantages for policy updates, we employ a critic warmup phase. For both tasks, the critic network is trained for 3 iterations on data from an initial rollout while the actor's policy is held constant. This stabilization of the value function is crucial for effective and stable PPO training. 
Totally, we use 2 H200-140GB GPU for RL training with roughly 12 hours for EB-ALFRED and EB-Manipulation.

\section{Additional Experiments}
\label{app:additional_experiments}

This section provides supplementary experimental results and detailed analyses that complement the main findings presented in Section~\ref{sec:exp}. The goal is to offer a more granular understanding of how specific design choices within the ERA framework contribute to overall performance and generalization. We structure our analysis around four key areas. First, we investigate the impact of oracle visual descriptions on low-level control, highlighting the critical role of accurate perception. Second, we dissect the environment-anchored prior dataset to evaluate the individual and combined contributions of its components. Third, we provide a more detailed breakdown of our reward design ablation, examining the synergistic effects of different reward signals. Finally, we offer a comprehensive comparison of token-level, bi-level, and our proposed turn-level Generalized Advantage Estimation (GAE) schemes, further validating the stability and effectiveness of turn-level credit assignment.

\subsection{Effect of Rule-based Visual Description for Low-level Control}
\begin{table}[h!]
\centering\small
\caption{
Ablation study on the effect of rule-based visual description in augmented-trajectory prior dataset. 
}
\label{tab:ablation_oracle}
\renewcommand{\arraystretch}{1.1}
\setlength\tabcolsep{6pt}
\resizebox{0.6\linewidth}{!}{

\begin{tabular}{
>{\raggedright\arraybackslash}p{5.5cm} 
>{\centering\arraybackslash}p{1.3cm} 
>{\centering\arraybackslash}p{1.3cm} 
}
\toprule

\textbf{Method} & \textbf{Seen} & \textbf{Unseen} \\

\midrule
Raw Trajectory (baseline)   & 25.0 & 7.3 \\
+ Rule-based visual description & 39.6 & 22.9 \\
+ Traj-Aug                  & 41.4 & 26.1 \\

\bottomrule
\end{tabular}}
\end{table}

While Embodied Prior Learning (EPL) enables agents to learn structured priors from visual trajectories, the raw trajectories often contain \textit{noisy or incorrect visual descriptions} generated by VLMs. Such inaccuracies can misguide perception and hinder effective reasoning about the environment. To isolate and quantify the impact of accurate perception-level grounding, we introduce two variants: (i) \textbf{+ Rule-based visual description}, which enhances the raw trajectories with \textit{rule-based visual descriptions only}, and (ii) \textbf{+ Traj-Aug}, which further \textit{augments trajectories by inserting step-wise reasoning} derived from these rule-based descriptions. This setup enables us to systematically evaluate how precise visual grounding and structured reasoning contribute to embodied learning and generalization.

Table~\ref{tab:ablation_oracle} highlights the impact of incorporating rule-based visual descriptions into EPL. Starting from the raw trajectory baseline, which yields 25.0\% on seen and 7.3\% on unseen environments, adding rule-based visual descriptions substantially improves performance to 39.6\% and 22.9\%, respectively. This demonstrates that accurate visual grounding plays a critical role in bridging perception and action. Further enriching trajectories with structured reasoning through Traj-Aug leads to additional gains, reaching 41.4\% on seen and 26.1\% on unseen environments. \textbf{These results confirm that both accurate visual descriptions and trajectory-level reasoning are essential for enhancing generalization in embodied agents, with the strongest improvements observed in unseen settings.}

\subsection{Effect of Different Components in the Environment-anchored Prior Dataset}
\begin{table}[h!]
\vspace{-6pt}
\centering\small
\caption{
Ablation study on environment-anchored prior dataset on \textsc{EB-ALFRED}, analyzing the impact of training data from different tasks.
}
\label{tab:env_anchored_ablation_alfred}
\renewcommand{\arraystretch}{1.1}
\setlength\tabcolsep{6pt}
\resizebox{0.7\linewidth}{!}{

\begin{tabular}{
>{\centering\arraybackslash}p{5cm} 
>{\centering\arraybackslash}p{1.3cm} 
>{\centering\arraybackslash}p{1.3cm} 
>{\centering\arraybackslash}p{1.3cm} 
}
\toprule

\multirow{2}{*}{\textbf{Data Type}} 
& \multicolumn{3}{c}{\bf EB-ALFRED} \\

\cmidrule(lr){2-4}
~ & \textbf{Avg} & \textbf{Common} & \textbf{Spatial} \\

\midrule

Masked Action Modeling only   & 41 & 38 & 44 \\
Action Sequence Reordering only  & 44 & 40 & 48 \\
All & 47 & 44 & 50 \\

\bottomrule
\end{tabular}}
\end{table}

\begin{table}[h!]
\vspace{-6pt}
\centering\small
\caption{
Ablation study on environment-anchored prior dataset on \textsc{EB-Manipulation}, analyzing the impact of training data from different tasks.
}\label{tab:env_anchored_ablation_manipulation}
\renewcommand{\arraystretch}{1.1}
\setlength\tabcolsep{6pt}
\resizebox{1.0\linewidth}{!}{

\begin{tabular}{
>{\centering\arraybackslash}p{4.5cm} 
>{\centering\arraybackslash}p{1.3cm} 
>{\centering\arraybackslash}p{1.3cm} 
>{\centering\arraybackslash}p{1.3cm} 
>{\centering\arraybackslash}p{1.3cm} 
>{\centering\arraybackslash}p{1.3cm} 
>{\centering\arraybackslash}p{1.3cm} 
}
\toprule

\multirow{2}{*}{\textbf{Data Type}} 
& \multicolumn{6}{c}{\bf EB-Manpulation} \\

\cmidrule(lr){2-7}
~ & \textbf{Avg} & \textbf{Base} & \textbf{Complex} & \textbf{Visual} & \textbf{Common} & \textbf{Spatial} \\

\midrule

No Comb. Grounding   & 36.7 & 45.8 & 37.5 & 45.8 & 35.4 & 18.8 \\
No Relative Grounding  & 36.6 & 47.9 & 33.3 & 47.9 & 33.3 & 20.8 \\ No Absolute Grounding & 35.0 & 43.8 & 33.3 & 45.8 & 33.3 & 18.8 \\
All  & 40 & 45.8 & 41.7 & 47.9 & 37.5 & 27.1 \\

\bottomrule
\end{tabular}}
\vspace{-0.5em}
\end{table}
The Environment-Anchored dataset is designed to enrich embodied prior learning by providing diverse supervision signals derived from the environment. However, it remains unclear which specific components contribute most to improving embodied reasoning. By systematically ablating these components, we aim to understand how different forms of environment-anchored supervision jointly shape embodied representations and generalization across tasks.

On \textbf{EB-ALFRED}, the results in Table~\ref{tab:env_anchored_ablation_alfred} reveal that incorporating both \textit{Masked Action Modeling (MAM)} and \textit{Action Sequence Reordering (ASR)} provides clear additive benefits. Using MAM alone achieves an average success rate of 41.0, while ASR alone slightly outperforms it with 44.0. When the two tasks are combined, performance rises to 47.0 on average, a gain of +6.0 over MAM alone and +3.0 over ASR alone. The improvements are consistent across subsets: on the \textit{Common} split, performance increases from 38.0 and 40.0 to 44.0 (+6.0 and +4.0, respectively), and on the \textit{Spatial} split from 44.0 and 48.0 to 50.0 (+6.0 and +2.0). These results indicate that temporal reasoning (ASR) and context-based inference (MAM) provide complementary supervision signals that jointly enhance high-level planning and perception grounding. On \textbf{EB-Manipulation}, Table~\ref{tab:env_anchored_ablation_manipulation} demonstrates similar synergy among grounding components. Excluding any single grounding dimension leads to a noticeable drop in performance. Without compositional grounding, the average success rate is 36.7; removing relative grounding yields 36.6, and removing absolute grounding further decreases it to 35.0. In contrast, jointly using all three grounding types raises the average to 40.0, a gain of +3.3 to +5.0 points. The benefit is most evident on unseen or spatially complex subsets: on the \textit{Spatial} split, performance improves from 18.8–20.8 to 27.1 (+6.3 to +8.3), and on the \textit{Common} split from 33.3–35.4 to 37.5 (+2.1 to +4.2). \textbf{These gains highlight that absolute, relative, and compositional grounding capture complementary spatial priors, with their joint supervision enabling stronger generalization across diverse embodied settings.}


\subsection{Effect of Different Reward Design in RL}
\begin{table*}[h!]
\centering\small
\caption{
Ablation of RL reward sparsity of unseen task performance
}\label{tab:rl_ablation_process_reward}
\renewcommand{\arraystretch}{1.1}
\setlength\tabcolsep{4pt}
\resizebox{0.9\linewidth}{!}{

\begin{tabular}{
>{\centering\arraybackslash}p{4cm} 
>{\centering\arraybackslash}p{1.3cm} 
>{\centering\arraybackslash}p{1.3cm} 
>{\centering\arraybackslash}p{1.3cm} 
>{\centering\arraybackslash}p{1.3cm} 
>{\centering\arraybackslash}p{1.3cm} 
>{\centering\arraybackslash}p{1.3cm} 
}
\toprule

\multirow{2}{*}{\textbf{Reward Sparsity}} 
& \multicolumn{3}{c}{\bf EB-ALFRED} 
& \multicolumn{3}{c}{\bf EB-Manipulation} \\

\cmidrule(lr){2-4} \cmidrule(lr){5-7}
~ & \textbf{Avg} & \textbf{Common} & \textbf{Spatial} 
   & \textbf{Avg} & \textbf{Common} & \textbf{Spatial} \\

\midrule

Outcome Only   & 46 & 42 & 50      & 41.7 & 47.9 & 35.4 \\
Outcome + Subgoal   & 49 & 44 & 54 & 42.8 & 45.9 & 39.6 \\
Outcome + B.S  &  47 & 44 & 50 & 41.7 & 47.9 & 35.4 \\
Outcome + Subgoal + B.S  & 60 & 54 & 66      & 43.8 & 47.9 & 39.6 \\
\bottomrule
\end{tabular}}
\vspace{-0.5em}
\end{table*}

Reward sparsity poses a significant challenge for credit assignment in long-horizon embodied tasks. To analyze the contribution of different reward components, we conduct a granular analysis on unseen subsets, with detailed results in Table~\ref{tab:rl_ablation_process_reward}.

On the long-horizon planning tasks in \textbf{EB-ALFRED}, dense, process-level rewards provide substantial benefits over a sparse, success-reward-only baseline. While individual components offer moderate gains—subgoal-based rewards improve the average success rate by 3\% and behavior-shaping by 1\%—their combination yields a strong synergistic effect. The full, composite reward function boosts the average success rate from 46\% to 60\%, a 14-point improvement that far exceeds the sum of individual gains. This synergy is particularly impactful on unseen subsets, with performance on \textit{Spatial} and \textit{Common Sense} tasks increasing by 16 and 12 points, respectively. \textbf{These results underscore that a composite reward function is critical for robust generalization in complex planning tasks.} On the contrast, for shorter-horizon, low-level control tasks as in \textbf{EB-Manipulation}, the impact of dense rewards much more limited. The overall gain is modest (41.7\% $\to$ 43.8\%) and the performance on individual subset is close. \textbf{We hypothesize this is because the shorter task horizons (with the shortest task finished by 4–5 steps) allow turn-level GAE to assign credit more effectively even with sparser rewards.} That being said, the combination of sub-goal reward and behavior-shaping reward in low level task still offers positive effect on agent's generalization on unseen tasks.

In summary, \textbf{our analysis confirms that the utility of dense rewards is associated with task horizon length.} That being said, different reward components can be designed to target distinct capabilities, and \textbf{their combination often produces synergistic effects.} This highlights the importance of designing a composite reward function that provides both positive guidance and negative constraints to facilitate effective and generalizable policy learning.

\subsection{Effect of Different GAEs in RL}
\begin{table*}[h!]
\centering\small
\caption{
Ablation of Different GAE in RL
}\label{tab:rl_ablation_gae}
\renewcommand{\arraystretch}{1.1}
\setlength\tabcolsep{4pt}
\resizebox{1.0\linewidth}{!}{

\begin{tabular}{
>{\centering\arraybackslash}p{3.5cm} 
>{\centering\arraybackslash}p{1.3cm} 
>{\centering\arraybackslash}p{1.3cm} 
>{\centering\arraybackslash}p{1.3cm}
>{\centering\arraybackslash}p{1.3cm} 
>{\centering\arraybackslash}p{1.3cm} 
>{\centering\arraybackslash}p{1.3cm} 
>{\centering\arraybackslash}p{1.3cm} 
>{\centering\arraybackslash}p{1.3cm} 
>{\centering\arraybackslash}p{1.3cm}  
>{\centering\arraybackslash}p{1.3cm} 
>{\centering\arraybackslash}p{1.3cm} 
>{\centering\arraybackslash}p{1.3cm} 
}
\toprule

\multirow{2}{*}{\textbf{Different GAE}} 
& \multicolumn{6}{c}{\bf EB-ALFRED} 
& \multicolumn{6}{c}{\bf EB-Manipulation} \\

\cmidrule(lr){2-7} \cmidrule(lr){8-13}
~ & \textbf{Avg} & \textbf{Base} & \textbf{Complex}& \textbf{Visual} & \textbf{Common} & \textbf{Spatial} 
   & \textbf{Avg} & \textbf{Base} & \textbf{Complex}& \textbf{Visual} & \textbf{Common} & \textbf{Spatial} \\

\midrule

Token-level GAE   & 58.8 & 70 & 72 & 56 & 40 & 56    & 40.0 & 43.9 & 47.9 & 37.5 & 45.8 & 25.0 \\
Bi-level GAE   & 60.4 & 72 & 70 & 60 & 44 & 56     & 42.1 & 47.9 & 47.9 & 39.6 & 39.6 & 35.4 \\
Turn-level GAE (Ours)  & 65.2 & 72  & 72 & 62 & 54 & 66 &    48.3 & 56.3 & 47.9 & 50.0 & 47.9 & 39.6   \\

\bottomrule
\end{tabular}}
\vspace{-0.5em}
\end{table*}

Standard token-level advantage estimation creates a granularity mismatch for VLM-based agents, where an entire sequence of generated tokens corresponds to a single environmental action. This can distribute credit improperly and destabilize training. To address this, we compare token-level, bi-level \citep{wang2025vagen}, and our proposed turn-level GAE. As shown in Table~\ref{tab:rl_ablation_gae}, aligning credit assignment with the unit of interaction via turn-level GAE consistently yields superior performance.

On the long-horizon tasks in \textbf{EB-ALFRED}, turn-level GAE significantly improves generalization. It achieves a 65.2\% average success rate, outperforming token-level GAE by 6.4 points. The gains are most pronounced on the unseen \textit{Common Sense} and \textit{Spatial} subsets (+14 and +10 points, respectively), demonstrating that turn-level credit assignment better helps agent learn in complex planning task. The advantage is also evident in the low-level control tasks of \textbf{EB-Manipulation}, where turn-level GAE achieves the highest average success rate (43.9\%). It yields substantial improvements on subsets requiring fine-grained perception, with performance increasing by 12.5 points on \textit{Visual} tasks and 14.6 points on \textit{Spatial} tasks over the token-level baseline.

In contrast, bi-level GAE offers only modest gains, as its method is still limited by a reliance on token-level value estimation. In summary, these results provide strong evidence that \textbf{matching the temporal unit of credit assignment to the agent's action abstraction is critical for learning generalizable policies.} Turn-level GAE proves to be a more stable and effective method for training \textbf{sequence-generating agents in interactive environments}.

\subsection{Error Analysis}
\label{app:error_analysis}

\begin{figure}
\centering
\begin{minipage}[t]{\linewidth}
    \includegraphics[width=0.45\linewidth]{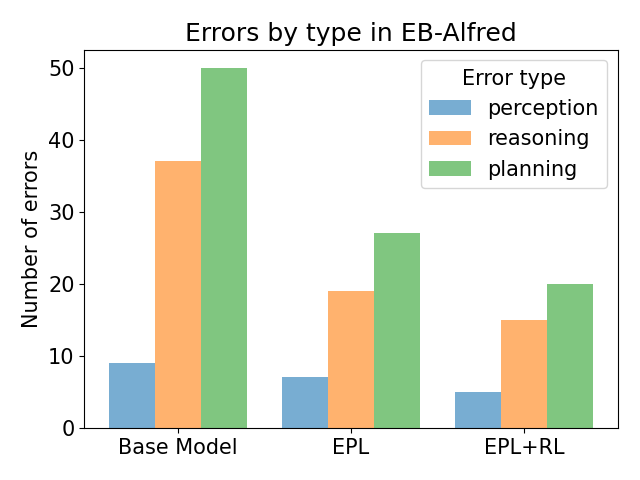}
    \includegraphics[width=0.45\linewidth]{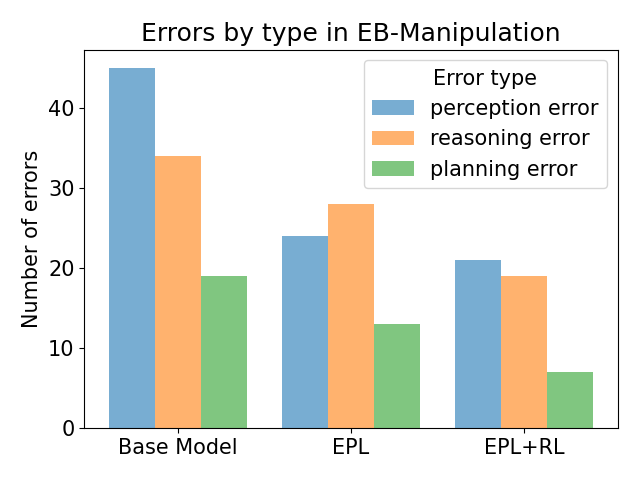}
\end{minipage}
\vspace{-10pt}
\caption{\small{Comparing error statistics in two benchmarks.}}
\label{fig:error_analysis}
\end{figure}

To understand how the EPL and RL stages in ERA reduce different types of errors, we conduct an error analysis on unseen subsets of both the high-level planning task EB-ALFRED (100 tasks) and the low-level control task EB-Manipulation (98 tasks). We categorize into 3 types of error: \textit{(i) Perception errors}: incorrect descriptions of the current state;  
\textit{(ii) Reasoning errors}: mistakes in reasoning about the current state or reflecting on history; \textit{(iii) Planning errors}: mistakes in planning future steps. The results, shown in Figure~\ref{fig:error_analysis}, reveal distinct patterns across task levels. In high-level tasks, reasoning and planning errors are dominant, while in low-level tasks, perception and reasoning errors are more prevalent. Across both settings, EPL and RL consistently reduce errors, but their effects differ in granularity: \textbf{EPL contributes to reducing all error types, whereas RL is especially effective at lowering reasoning and planning errors.} When combined, EPL and RL (i.e., ERA) achieve the lowest error rates across all categories. We provide a deeper case analysis for ERA below.

\clearpage
\section{Agent Prompt}
We provide the agent input prompts used as textual input to ERA for EB-ALFRED and EB-Manipulation.
\label{app:agent_prompt}


\begin{tcolorbox}[colback=gray!5!white, colframe=gray!75!black, 
title=Training System Prompt for EB-ALFRED, boxrule=0.3mm, width=\textwidth, arc=3mm, auto outer arc=true]
\#\# You are a robot operating in a home. Given a task, you must accomplish the task using a defined set of actions to achieve the desired outcome.\newline    
\#\# Action Descriptions and Validity Rules\newline    
• Find: Parameterized by the name of the receptacle to navigate to. So long as the object is present in the scene, this skill is always valid.
\newline
• Pick up: Parameterized by the name of the object to pick. Only valid if the robot is close to the object, not holding another object, and the object is not inside a closed receptacle.
\newline
• Put down: Parameterized by the name of the object to put down to a nearby receptacle. Only valid if the robot is holding an object.
\newline
• Drop: Parameterized by the name of the object to put down. It is different from the Put down action, as this does not guarantee the held object will be put into a specified receptacle. 
\newline
• Open: Parameterized by the name of the receptacle to open. Only valid if the receptacle is closed and the robot is close to the receptacle.
\newline
• Close: Parameterized by the name of the receptacle to close. Only valid if the receptacle is open and the robot is close to the receptacle.
\newline
• Turn on: Parameterized by the name of the object to turn on. Only valid if the object is turned off and the robot is close to the object.
\newline
• Turn off: Parameterized by the name of the object to turn off. Only valid if the object is turned on and the robot is close to the object.
\newline
• Slice: Parameterized by the name of the object to slice. Only valid if the object is sliceable and the robot is close to the object.
\newline
\#\# The available action id (0 - \{len(SKILL SET) - 1\}) and action names are: \{SKILL SET\}.\newline\newline
\#\# Guidelines\newline
1. **Output Plan**: Avoid generating empty plan. Each plan should include no more than 20 actions.\newline
2. **Visibility**: Always locate a visible object by the 'find' action before interacting with it.\newline
3. **Action Guidelines**: Make sure match the action name and its corresponding action id in the output.\newline 
Avoid performing actions that do not meet the defined validity criteria. For instance, if you want to put object in a receptacle, use 'put down' rather than 'drop' actions.\newline
4. **Prevent Repeating Action Sequences**: Do not repeatedly execute the same action or sequence of actions.\newline
Try to modify the action sequence because previous actions do not lead to success.\newline
5. **Multiple Instances**: There may be multiple instances of the same object, distinguished by an index following their names, e.g., Cabinet\_2, Cabinet\_3. You can explore these instances if you do not find the desired object in the current receptacle.\newline
6. **Reflection on History and Feedback**: Use interaction history and feedback from the environment to refine and improve your current plan.\newline
If the last action is invalid, reflect on the reason, such as not adhering to action rules or missing preliminary actions, and adjust your plan accordingly.\newline

**Generation Guide**\newline
- Include the thinking process between \texttt{<|think\_start|>} and \texttt{<|think\_end|>}.\newline
- Include only the target action in \texttt{<|action\_start|>} and \texttt{<|action\_end|>}, i.e., the content inside should be nothing more than \texttt{[action\_id, 'action\_name']}, where the action id is an integer and the action name is the corresponding name. Do not include any other text, such as quotation marks.
\end{tcolorbox}

\begin{tcolorbox}[colback=gray!5!white, colframe=gray!75!black, 
title=Training System Prompt for EB-Manipulation, boxrule=0.3mm, width=\textwidth, arc=3mm, auto outer arc=true]
\#\# You are a Franka Panda robot with a parallel gripper. You can perform various tasks and output a sequence of gripper actions to accomplish a given task with images of your status. The input space, output action space and color space are defined as follows:\newline

** Input Space **\newline
- Each input object is represented as a 3D discrete position in the following format: \texttt{[X, Y, Z]}. \newline
- There is a red XYZ coordinate frame located in the top-left corner of the table. The X-Y plane is the table surface. \newline
- The allowed range of X, Y, Z is \texttt{[0, 100]}.\newline 
- Objects are ordered by Y in ascending order.\newline

** Output Action Space **\newline
- Each output action is represented as a 7D discrete gripper action in the following format: \texttt{[X, Y, Z, Roll, Pitch, Yaw, Gripper state]}.\newline
- X, Y, Z are the 3D discrete position of the gripper in the environment. It follows the same coordinate system as the input object coordinates.\newline
- The allowed range of X, Y, Z is \texttt{[0, 100]}.\newline
- Roll, Pitch, Yaw are the 3D discrete orientation of the gripper in the environment, represented as discrete Euler Angles.\newline 
- The allowed range of Roll, Pitch, Yaw is \texttt{[0, 120]} and each unit represents 3 degrees.\newline
- Gripper state is 0 for close and 1 for open.\newline

** Color space **\newline
- Each object can be described using one of the colors below:\newline
  [\texttt{"red"}, \texttt{"maroon"}, \texttt{"lime"}, \texttt{"green"}, \texttt{"blue"}, \texttt{"navy"}, \texttt{"yellow"}, \texttt{"cyan"}, \texttt{"magenta"}, \texttt{"silver"}, \texttt{"gray"}, \texttt{"olive"}, \texttt{"purple"}, \texttt{"teal"}, \texttt{"azure"}, \texttt{"violet"}, \texttt{"rose"}, \texttt{"black"}, \texttt{"white"}],\newline

** Generation Guide **\newline
- Include the thinking process between \texttt{<|think\_start|>} and \texttt{<|think\_end|>}\newline
- Include only the target action in \texttt{<|action\_start|>} and \texttt{<|action\_end|>}, i.e. the content inside \texttt{<|action\_start|>} and \texttt{<|action\_end|>} should be nothing more than the 7-DoF vector. Do not include any other thing, such as \texttt{''}.
\end{tcolorbox}

\clearpage
\section{Embodied Prior Learning Dataset}
\label{app:prior_dataset}

\subsection{Raw Trajectory}
We provide an example of raw trajectory collected from EmbodiedBench for both EB-ALFRED and EB-Manipulation.

\begin{tcolorbox}[
  colback=gray!5!white, colframe=gray!75!black,
  title=Raw trajectory for EB-ALFRED,
  boxrule=0.3mm, width=\textwidth, arc=3mm,
  auto outer arc=true, breakable
]

\textbf{Step 1:}\par\vspace{4pt}

\noindent
\begin{minipage}[t]{0.2\textwidth}
  \textbf{Input image}\par\vspace{4pt}
  \includegraphics[width=0.9 \linewidth]{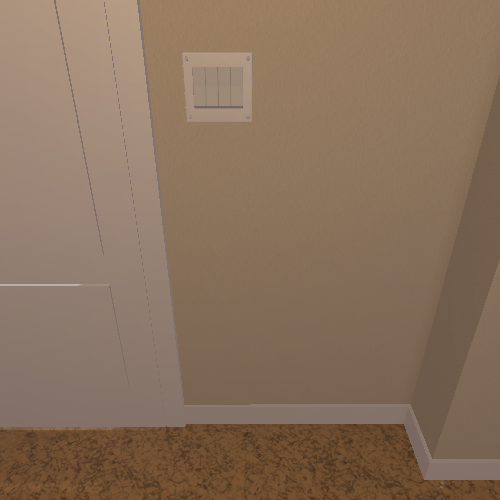}
\end{minipage}\hfill
\begin{minipage}[t]{0.8\textwidth}
  \textbf{Prompt}\par\vspace{4pt}
  instruction: Set plate with a spoon in it on the kitchen table\par
  interaction\_history: []\par
  Based on the above information, please provide the action for the next step
  to complete the task. Think, then act.
\end{minipage}

\vspace{8pt}
\noindent\textbf{Generation}\par
\begingroup\small 
\texttt{<|think\_start|>}This is the first step in the episode, I should start by
formulating the first plan: visual\_description: The image shows a corner of a room
with beige walls and carpeted flooring. There appears to be a vent or air return
mounted on the wall. The space appears to be empty without any visible furniture or
objects. reasoning\_and\_reflection: Since the room appears empty, we'll need to search
for and locate the required items (plate, spoon, and dining table) before we can
complete the task. We'll need to find these items first, then arrange them according
to the instruction. language\_plan: 1. First, locate the plate in the room. 2. Pick up
the plate once found. 3. Find the spoon. 4. Pick up the spoon. 5. Find the dining
table. 6. Put down the spoon on the plate. 7. Place the plate with the spoon on the
dining table.\texttt{<|think\_end|>}\par
\texttt{<|action\_start|>}[31, 'find a Plate']\texttt{<|action\_end|>}
\endgroup\newline

\textbf{Step 2:}\par\vspace{4pt}

\noindent
\begin{minipage}[t]{0.2\textwidth}
  \textbf{Input image}\par\vspace{4pt}
  \includegraphics[width=0.9\linewidth]{figures/alfred/episode_2_step_1.png}
\end{minipage}\hfill
\begin{minipage}[t]{0.8\textwidth}
  \textbf{Prompt}\par\vspace{4pt}
  instruction: Set plate with a spoon in it on the kitchen table\par
  interaction\_history:\par
  {\small\ttfamily
  [\{``step\_id'': 0, ``thinking'': ``This is the first step in the episode, I should
  start by formulating the first plan: visual\_description: The image shows a corner of
  a room with beige walls and carpeted flooring. There appears to be a vent or air return
  mounted on the wall. The space appears to be empty without any visible furniture or
  objects. reasoning\_and\_reflection: Since the room appears empty, we'll need to search
  for and locate the required items (plate, spoon, and dining table) before we can complete
  the task. We'll need to find these items first, then arrange them according to the
  instruction. language\_plan: 1. First, locate the plate in the room. 2. Pick up the
  plate once found. 3. Find the spoon. 4. Pick up the spoon. 5. Find the dining table.
  6. Put down the spoon on the plate. 7. Place the plate with the spoon on the dining
  table.'', ``action'': ``[31, 'find a Plate']'', ``env\_feedback'': ``Last action
  executed successfully.''\}]]
  }\par
  Based on the above information, please provide the action for the next step
  to complete the task. Think, then act.
\end{minipage}

\vspace{8pt}
\noindent\textbf{Generation}\par
\begingroup\small
\texttt{<|think\_start|>}Based on the interaction history and current observation, I am
in the middle of the last plan and will continue carrying it out.\texttt{<|think\_end|>}\par
\texttt{<|action\_start|>}[122, 'pick up the Plate']\texttt{<|action\_end|>}
\endgroup\newline


\textbf{Step 3:}\par\vspace{4pt}
\noindent
\begin{minipage}[t]{0.2\textwidth}
  \textbf{Input image}\par\vspace{4pt}
  \includegraphics[width=0.9\linewidth]{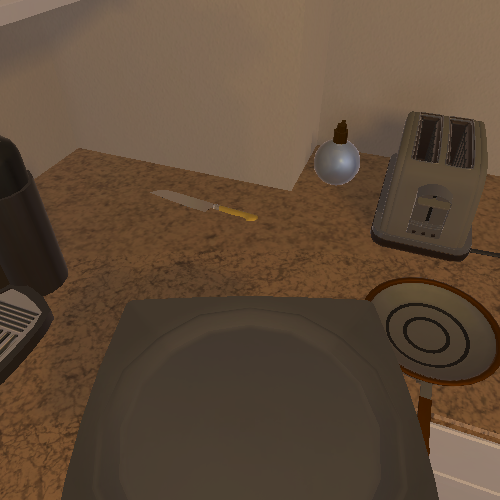}
\end{minipage}\hfill
\begin{minipage}[t]{0.8\textwidth}
  \textbf{Prompt}\par\vspace{4pt}
  instruction: Set plate with a spoon in it on the kitchen table\par
  interaction\_history:\par
  {\small\ttfamily
  [\{``step\_id'': 1, ``thinking'': ``Based on the interaction history and current observation, I am in the middle of the last plan and will continue carrying it out.'', ``action'': ``[122, 'pick up the Plate']'', ``env\_feedback'': ``Last action executed successfully.''\}]]
  }\par
  Based on the above information, please provide the action for the next step
  to complete the task. Think, then act.
\end{minipage}

\vspace{8pt}
\noindent\textbf{Generation}\par
\begingroup\small
\texttt{<|think\_start|>}Based on the interaction history and current observation, I am in the middle of the last plan and will continue carrying it out.\texttt{<|think\_end|>}\par
\texttt{<|action\_start|>}[19, 'find a Spoon']\texttt{<|action\_end|>}
\endgroup\newline

\textbf{Step 4:}\par\vspace{4pt}
\noindent
\begin{minipage}[t]{0.2\textwidth}
  \textbf{Input image}\par\vspace{4pt}
  \includegraphics[width=0.9\linewidth]{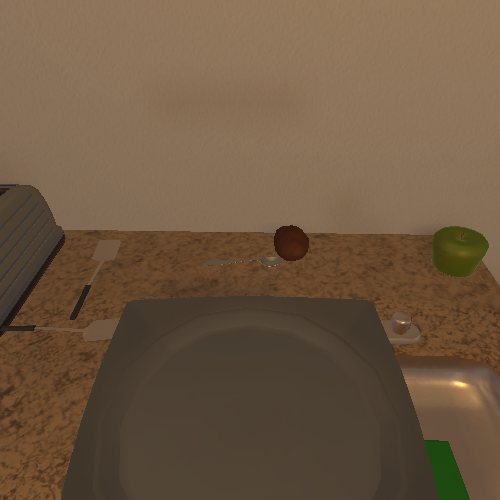}
\end{minipage}\hfill
\begin{minipage}[t]{0.8\textwidth}
  \textbf{Prompt}\par\vspace{4pt}
  instruction: Set plate with a spoon in it on the kitchen table\par
  interaction\_history:\par
  {\small\ttfamily
  [\{``step\_id'': 2, ``thinking'': ``Based on the interaction history and current observation, I am in the middle of the last plan and will continue carrying it out.'', ``action'': ``[19, 'find a Spoon']'', ``env\_feedback'': ``Last action executed successfully.''\}]]
  }\par
  Based on the above information, please provide the action for the next step
  to complete the task. Think, then act.
\end{minipage}

\vspace{8pt}
\noindent\textbf{Generation}\par
\begingroup\small
\texttt{<|think\_start|>}Based on the interaction history and current observation, I am in the middle of the last plan and will continue carrying it out.\texttt{<|think\_end|>}\par
\texttt{<|action\_start|>}[105, 'pick up the Spoon']\texttt{<|action\_end|>}
\endgroup\newline

\textbf{Step 5:}\par\vspace{4pt}
\noindent
\begin{minipage}[t]{0.2\textwidth}
  \textbf{Input image}\par\vspace{4pt}
  \includegraphics[width=0.9\linewidth]{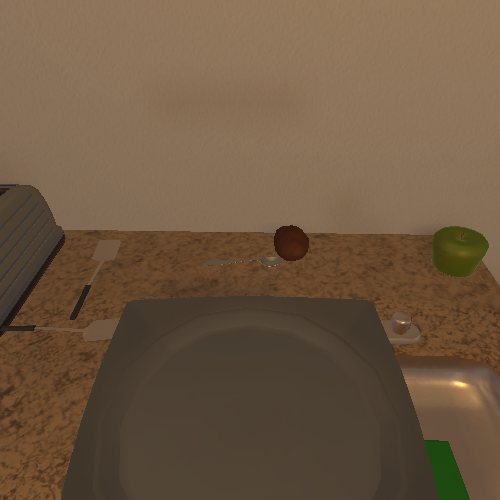}
\end{minipage}\hfill
\begin{minipage}[t]{0.8\textwidth}
  \textbf{Prompt}\par\vspace{4pt}
  instruction: Set plate with a spoon in it on the kitchen table\par
  interaction\_history:\par
  {\small\ttfamily
  [\{``step\_id'': 3, ``thinking'': ``Based on the interaction history and current observation, I am in the middle of the last plan and will continue carrying it out.'', ``action'': ``[105, 'pick up the Spoon']'', ``env\_feedback'': ``Last action is invalid. Robot is currently holding Plate''\}]]
  }\par
  Based on the above information, please provide the action for the next step
  to complete the task. Think, then act.
\end{minipage}

\vspace{8pt}
\noindent\textbf{Generation}\par
\begingroup\small
\texttt{<|think\_start|>}Based on the interaction history and current observation, I have completed the previous plan and can now formulate a new plan: visual\_description: The image shows a kitchen countertop surface. There appears to be some objects on the counter, including what looks like an apple or round fruit. reasoning\_and\_reflection: From the interaction history, the robot successfully found and picked up a plate (Steps 0-1), then found a spoon (Step 2). However, when trying to pick up the spoon (Step 3), the action failed because the robot was already holding the plate. To achieve the goal of placing a plate with a spoon in it on the kitchen table, we need to first put down the plate, pick up the spoon, then put the spoon on the plate, pick up the plate again, and finally place it on the dining table. language\_plan: 1. Find the dining table to put down the plate temporarily. 2. Put down the plate. 3. Find the spoon again. 4. Pick up the spoon. 5. Find the plate. 6. Put the spoon on the plate. 7. Pick up the plate (now with spoon in it). 8. Find the dining table. 9. Put down the plate with spoon on the table.\texttt{<|think\_end|>}\par
\texttt{<|action\_start|>}[18, 'find a DiningTable']\texttt{<|action\_end|>}
\endgroup\newline

\textbf{Step 6:}\par\vspace{4pt}
\noindent
\begin{minipage}[t]{0.2\textwidth}
  \textbf{Input image}\par\vspace{4pt}
  \includegraphics[width=0.9\linewidth]{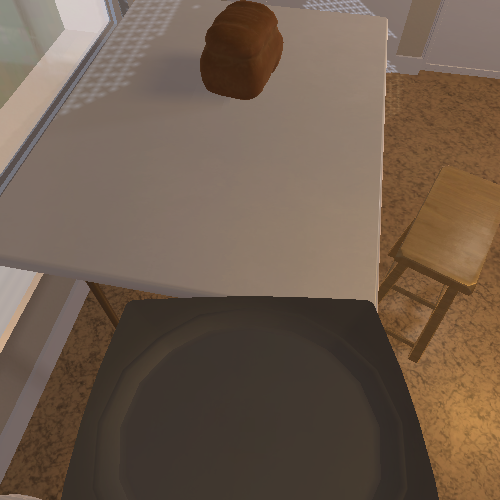}
\end{minipage}\hfill
\begin{minipage}[t]{0.8\textwidth}
  \textbf{Prompt}\par\vspace{4pt}
  instruction: Set plate with a spoon in it on the kitchen table\par
  interaction\_history:\par
  {\small\ttfamily
  [\{``step\_id'': 4, ``thinking'': ``Based on the interaction history and current observation, I have completed the previous plan and can now formulate a new plan: visual\_description: The image shows a kitchen countertop surface. There appears to be some objects on the counter, including what looks like an apple or round fruit. reasoning\_and\_reflection: From the interaction history, the robot successfully found and picked up a plate (Steps 0-1), then found a spoon (Step 2). However, when trying to pick up the spoon (Step 3), the action failed because the robot was already holding the plate. To achieve the goal of placing a plate with a spoon in it on the kitchen table, we need to first put down the plate, pick up the spoon, then put the spoon on the plate, pick up the plate again, and finally place it on the dining table. language\_plan: 1. Find the dining table to put down the plate temporarily. 2. Put down the plate. 3. Find the spoon again. 4. Pick up the spoon. 5. Find the plate. 6. Put the spoon on the plate. 7. Pick up the plate (now with spoon in it). 8. Find the dining table. 9. Put down the plate with spoon on the table.'', ``action'': ``[18, 'find a DiningTable']'', ``env\_feedback'': ``Last action executed successfully.''\}]]
  }\par
  Based on the above information, please provide the action for the next step
  to complete the task. Think, then act.
\end{minipage}

\vspace{8pt}
\noindent\textbf{Generation}\par
\begingroup\small
\texttt{<|think\_start|>}Based on the interaction history and current observation, I am in the middle of the last plan and will continue carrying it out.\texttt{<|think\_end|>}\par
\texttt{<|action\_start|>}[133, 'put down the object in hand']\texttt{<|action\_end|>}
\endgroup\newline

\textbf{Step 7:}\par\vspace{4pt}
\noindent
\begin{minipage}[t]{0.2\textwidth}
  \textbf{Input image}\par\vspace{4pt}
  \includegraphics[width=0.9\linewidth]{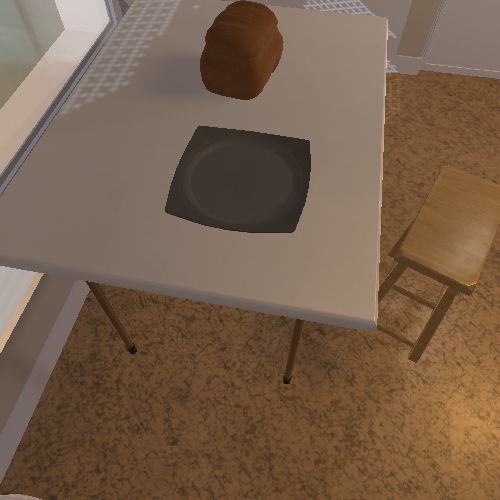}
\end{minipage}\hfill
\begin{minipage}[t]{0.8\textwidth}
  \textbf{Prompt}\par\vspace{4pt}
  instruction: Set plate with a spoon in it on the kitchen table\par
  interaction\_history:\par
  {\small\ttfamily
  [\{``step\_id'': 5, ``thinking'': ``Based on the interaction history and current observation, I am in the middle of the last plan and will continue carrying it out.'', ``action'': ``[133, 'put down the object in hand']'', ``env\_feedback'': ``Last action executed successfully.''\}]]
  }\par
  Based on the above information, please provide the action for the next step
  to complete the task. Think, then act.
\end{minipage}

\vspace{8pt}
\noindent\textbf{Generation}\par
\begingroup\small
\texttt{<|think\_start|>}Based on the interaction history and current observation, I am in the middle of the last plan and will continue carrying it out.\texttt{<|think\_end|>}\par
\texttt{<|action\_start|>}[19, 'find a Spoon']\texttt{<|action\_end|>}
\endgroup\newline

\textbf{Step 8:}\par\vspace{4pt}
\noindent
\begin{minipage}[t]{0.2\textwidth}
  \textbf{Input image}\par\vspace{4pt}
  \includegraphics[width=0.9\linewidth]{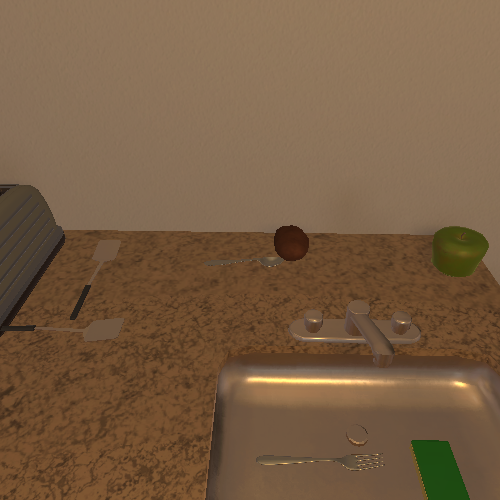}
\end{minipage}\hfill
\begin{minipage}[t]{0.8\textwidth}
  \textbf{Prompt}\par\vspace{4pt}
  instruction: Set plate with a spoon in it on the kitchen table\par
  interaction\_history:\par
  {\small\ttfamily
  [\{``step\_id'': 6, ``thinking'': ``Based on the interaction history and current observation, I am in the middle of the last plan and will continue carrying it out.'', ``action'': ``[19, 'find a Spoon']'', ``env\_feedback'': ``Last action executed successfully.''\}]]
  }\par
  Based on the above information, please provide the action for the next step
  to complete the task. Think, then act.
\end{minipage}

\vspace{8pt}
\noindent\textbf{Generation}\par
\begingroup\small
\texttt{<|think\_start|>}Based on the interaction history and current observation, I am in the middle of the last plan and will continue carrying it out.\texttt{<|think\_end|>}\par
\texttt{<|action\_start|>}[105, 'pick up the Spoon']\texttt{<|action\_end|>}
\endgroup\newline

\textbf{Step 9:}\par\vspace{4pt}
\noindent
\begin{minipage}[t]{0.2\textwidth}
  \textbf{Input image}\par\vspace{4pt}
  \includegraphics[width=0.9\linewidth]{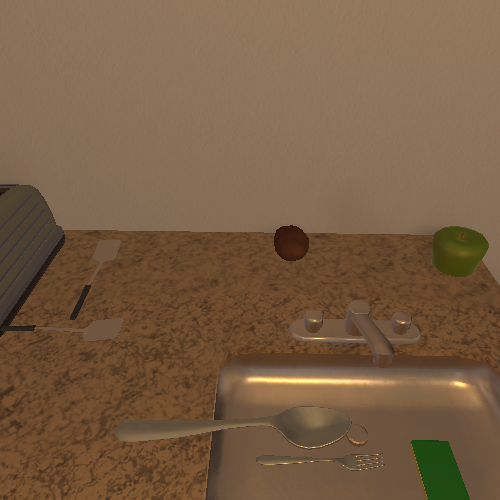}
\end{minipage}\hfill
\begin{minipage}[t]{0.8\textwidth}
  \textbf{Prompt}\par\vspace{4pt}
  instruction: Set plate with a spoon in it on the kitchen table\par
  interaction\_history:\par
  {\small\ttfamily
  [\{``step\_id'': 7, ``thinking'': ``Based on the interaction history and current observation, I am in the middle of the last plan and will continue carrying it out.'', ``action'': ``[105, 'pick up the Spoon']'', ``env\_feedback'': ``Last action executed successfully.''\}]]
  }\par
  Based on the above information, please provide the action for the next step
  to complete the task. Think, then act.
\end{minipage}

\vspace{8pt}
\noindent\textbf{Generation}\par
\begingroup\small
\texttt{<|think\_start|>}Based on the interaction history and current observation, I am in the middle of the last plan and will continue carrying it out.\texttt{<|think\_end|>}\par
\texttt{<|action\_start|>}[31, 'find a Plate']\texttt{<|action\_end|>}
\endgroup\newline

\textbf{Step 10:}\par\vspace{4pt}
\noindent
\begin{minipage}[t]{0.2\textwidth}
  \textbf{Input image}\par\vspace{4pt}
  \includegraphics[width=0.9\linewidth]{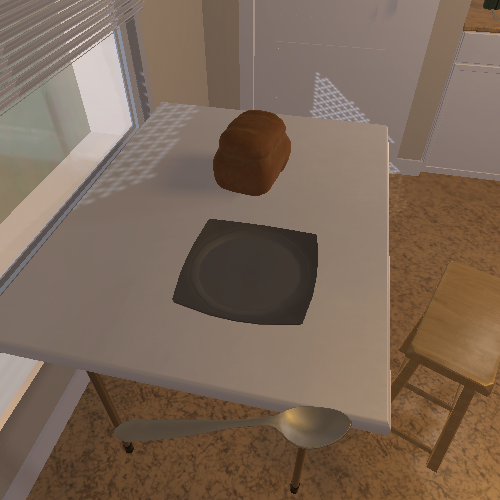}
\end{minipage}\hfill
\begin{minipage}[t]{0.8\textwidth}
  \textbf{Prompt}\par\vspace{4pt}
  instruction: Set plate with a spoon in it on the kitchen table\par
  interaction\_history:\par
  {\small\ttfamily
  [\{``step\_id'': 8, ``thinking'': ``Based on the interaction history and current observation, I am in the middle of the last plan and will continue carrying it out.'', ``action'': ``[31, 'find a Plate']'', ``env\_feedback'': ``Last action executed successfully.''\}]]
  }\par
  Based on the above information, please provide the action for the next step
  to complete the task. Think, then act.
\end{minipage}

\vspace{8pt}
\noindent\textbf{Generation}\par
\begingroup\small
\texttt{<|think\_start|>}Based on the interaction history and current observation, I am in the middle of the last plan and will continue carrying it out.\texttt{<|think\_end|>}\par
\texttt{<|action\_start|>}[133, 'put down the object in hand']\texttt{<|action\_end|>}
\endgroup
\end{tcolorbox}

\begin{tcolorbox}[
  colback=gray!5!white, colframe=gray!75!black,
  title=Raw trajectory for EB-Manipulation,
  boxrule=0.3mm, width=\textwidth, arc=3mm,
  auto outer arc=true, breakable
]

\textbf{Step 1:}\par\vspace{4pt}
\noindent
\begin{minipage}[t]{0.2\textwidth}
  \textbf{Input image}\par\vspace{4pt}
  \includegraphics[width=0.9\linewidth]{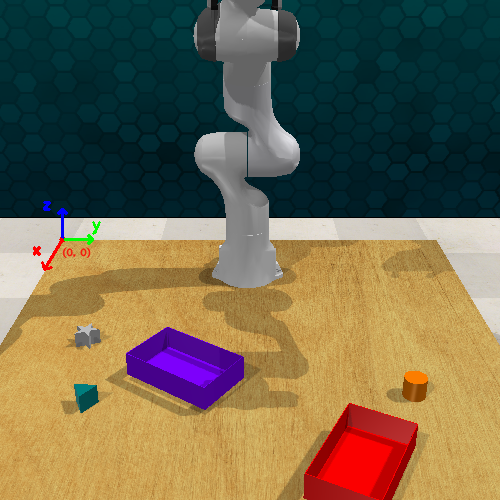}
\end{minipage}\hfill
\begin{minipage}[t]{0.8\textwidth}
  \textbf{Prompt}\par\vspace{4pt}
  \textbf{instruction:} Pick up the star and place it into the red container.\par
  \textbf{interaction\_history:} []\par
  \textbf{additional\_info:} {\ttfamily\{``object 1'': [35, 15, 17], ``object 2'': [57, 18, 18], ``object 3'': [49, 37, 17], ``object 4'': [76, 69, 17], ``object 5'': [54, 81, 18]\}}\par
  Based on the above information, please provide the action for the next step to complete the task. Think, then act.
\end{minipage}

\vspace{8pt}
\noindent\textbf{Generation}\par
\begingroup\small
\texttt{<|think\_start|>}This is the first step in the episode, I should start by formulating the first plan: visual\_description: From left to right, I can see a red star at [35, 15, 17], a teal cube at [57, 18, 18], a purple container at [49, 37, 17], a red container at [76, 69, 17], and an orange cylinder at [54, 81, 18]. reasoning\_and\_reflection: The task is to pick up the star located at [35, 15, 17] and place it into the red container located at [76, 69, 17]. I need to move the gripper to the star's position, close the gripper to grasp it, lift it up, move to the red container's position, and then open the gripper to release the star into the container. language\_plan: 1. Move gripper above the red star at [35, 15, 17]\texttt{\textbackslash n}2. Lower gripper to grasp the star\texttt{\textbackslash n}3. Close gripper to pick up the star\texttt{\textbackslash n}4. Lift the star up\texttt{\textbackslash n}5. Move the star above the red container at [76, 69, 17]\texttt{\textbackslash n}6. Lower the star into the container\texttt{\textbackslash n}7. Open gripper to release the star\texttt{<|think\_end|>}\par
\texttt{<|action\_start|>}[35, 15, 27, 0, 60, 0, 1]\texttt{<|action\_end|>}
\endgroup\newline

\textbf{Step 2:}\par\vspace{4pt}
\noindent
\begin{minipage}[t]{0.2\textwidth}
  \textbf{Input image}\par\vspace{4pt}
  \includegraphics[width=0.9\linewidth]{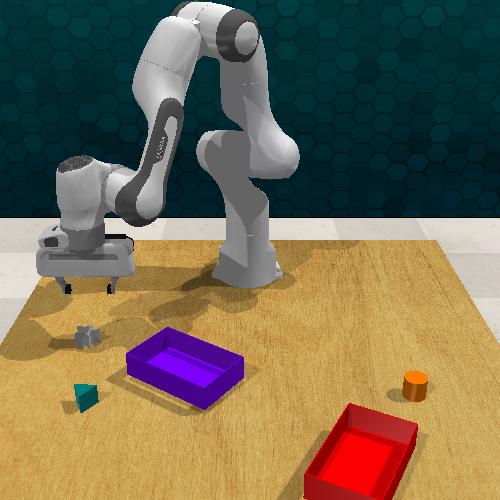}
\end{minipage}\hfill
\begin{minipage}[t]{0.8\textwidth}
  \textbf{Prompt}\par\vspace{4pt}
  \textbf{instruction:} Pick up the star and place it into the red container.\par
  \textbf{interaction\_history:}\par
  {\small\ttfamily[\{``step\_id'': 1, ``thinking'': ``This is the first step in the episode, I should start by formulating the first plan: visual\_description: From left to right, I can see a red star at [35, 15, 17], a teal cube at [57, 18, 18], a purple container at [49, 37, 17], a red container at [76, 69, 17], and an orange cylinder at [54, 81, 18]. reasoning\_and\_reflection: The task is to pick up the star located at [35, 15, 17] and place it into the red container located at [76, 69, 17]. I need to move the gripper to the star's position, close the gripper to grasp it, lift it up, move to the red container's position, and then open the gripper to release the star into the container. language\_plan: 1. Move gripper above the red star at [35, 15, 17]\textbackslash n2. Lower gripper to grasp the star\textbackslash n3. Close gripper to pick up the star\textbackslash n4. Lift the star up\textbackslash n5. Move the star above the red container at [76, 69, 17]\textbackslash n6. Lower the star into the container\textbackslash n7. Open gripper to release the star'', ``action'': ``[35, 15, 27, 0, 60, 0, 1]'', ``env\_feedback'': ``Last action was successful.''\}]]}\par
  \textbf{additional\_info:} {\ttfamily\{``object 1'': [35, 15, 17], ``object 2'': [57, 18, 18], ``object 3'': [49, 37, 17], ``object 4'': [76, 69, 17], ``object 5'': [54, 81, 18]\}}\par
  Based on the above information, please provide the action for the next step to complete the task. Think, then act.
\end{minipage}

\vspace{8pt}
\noindent\textbf{Generation}\par
\begingroup\small
\texttt{<|think\_start|>}Based on the interaction history and current observation, I am in the middle of the last plan and will continue carrying it out.\texttt{<|think\_end|>}\par
\texttt{<|action\_start|>}[35, 15, 17, 0, 60, 0, 0]\texttt{<|action\_end|>}
\endgroup\newline

\textbf{Step 3:}\par\vspace{4pt}
\noindent
\begin{minipage}[t]{0.2\textwidth}
  \textbf{Input image}\par\vspace{4pt}
  \includegraphics[width=0.9\linewidth]{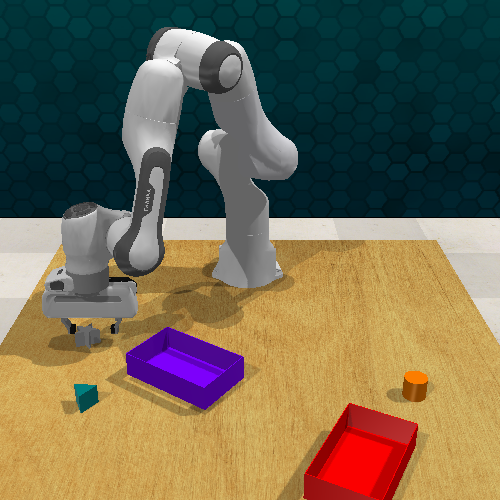}
\end{minipage}\hfill
\begin{minipage}[t]{0.8\textwidth}
  \textbf{Prompt}\par\vspace{4pt}
  \textbf{instruction:} Pick up the star and place it into the red container.\par
  \textbf{interaction\_history:}\par
  {\small\ttfamily[\{``step\_id'': 2, ``thinking'': ``Based on the interaction history and current observation, I am in the middle of the last plan and will continue carrying it out.'', ``action'': ``[35, 15, 17, 0, 60, 0, 0]'', ``env\_feedback'': ``Last action was successful.''\}]]}\par
  \textbf{additional\_info:} {\ttfamily\{``object 1'': [35, 15, 17], ``object 2'': [57, 18, 18], ``object 3'': [49, 37, 17], ``object 4'': [76, 69, 17], ``object 5'': [54, 81, 18]\}}\par
  Based on the above information, please provide the action for the next step to complete the task. Think, then act.
\end{minipage}

\vspace{8pt}
\noindent\textbf{Generation}\par
\begingroup\small
\texttt{<|think\_start|>}Based on the interaction history and current observation, I am in the middle of the last plan and will continue carrying it out.\texttt{<|think\_end|>}\par
\texttt{<|action\_start|>}[35, 15, 27, 0, 60, 0, 0]\texttt{<|action\_end|>}
\endgroup\newline

\textbf{Step 4:}\par\vspace{4pt}
\noindent
\begin{minipage}[t]{0.2\textwidth}
  \textbf{Input image}\par\vspace{4pt}
  \includegraphics[width=0.9\linewidth]{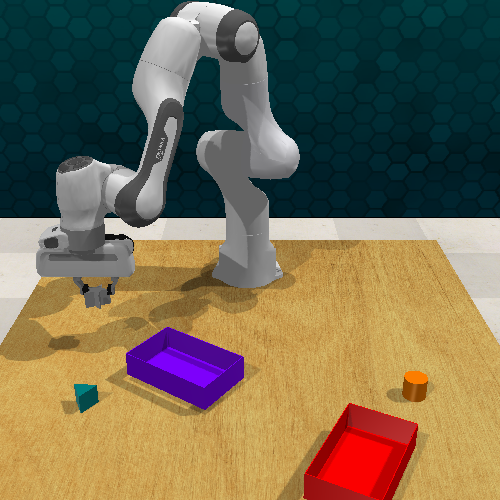}
\end{minipage}\hfill
\begin{minipage}[t]{0.8\textwidth}
  \textbf{Prompt}\par\vspace{4pt}
  \textbf{instruction:} Pick up the star and place it into the red container.\par
  \textbf{interaction\_history:}\par
  {\small\ttfamily[\{``step\_id'': 3, ``thinking'': ``Based on the interaction history and current observation, I am in the middle of the last plan and will continue carrying it out.'', ``action'': ``[35, 15, 27, 0, 60, 0, 0]'', ``env\_feedback'': ``Last action was successful.''\}]]}\par
  \textbf{additional\_info:} {\ttfamily\{``object 1'': [35, 15, 17], ``object 2'': [57, 18, 18], ``object 3'': [49, 37, 17], ``object 4'': [76, 69, 17], ``object 5'': [54, 81, 18]\}}\par
  Based on the above information, please provide the action for the next step to complete the task. Think, then act.
\end{minipage}

\vspace{8pt}
\noindent\textbf{Generation}\par
\begingroup\small
\texttt{<|think\_start|>}Based on the interaction history and current observation, I am in the middle of the last plan and will continue carrying it out.\texttt{<|think\_end|>}\par
\texttt{<|action\_start|>}[76, 69, 27, 0, 60, 0, 0]\texttt{<|action\_end|>}
\endgroup\newline

\textbf{Step 5:}\par\vspace{4pt}
\noindent
\begin{minipage}[t]{0.2\textwidth}
  \textbf{Input image}\par\vspace{4pt}
  \includegraphics[width=0.9\linewidth]{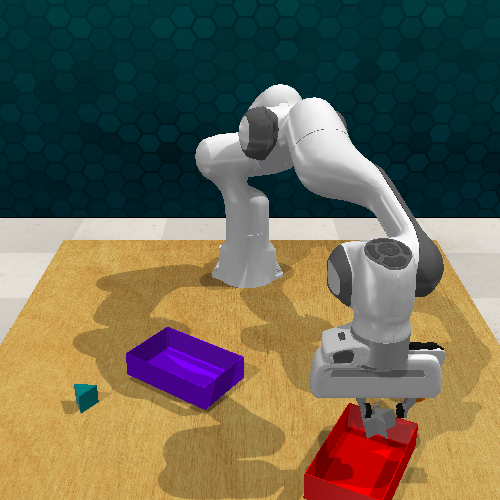}
\end{minipage}\hfill
\begin{minipage}[t]{0.8\textwidth}
  \textbf{Prompt}\par\vspace{4pt}
  \textbf{instruction:} Pick up the star and place it into the red container.\par
  \textbf{interaction\_history:}\par
  {\small\ttfamily[\{``step\_id'': 4, ``thinking'': ``Based on the interaction history and current observation, I am in the middle of the last plan and will continue carrying it out.'', ``action'': ``[76, 69, 27, 0, 60, 0, 0]'', ``env\_feedback'': ``Last action was successful.''\}]]}\par
  \textbf{additional\_info:} {\ttfamily\{``object 1'': [35, 15, 17], ``object 2'': [57, 18, 18], ``object 3'': [49, 37, 17], ``object 4'': [76, 69, 17], ``object 5'': [54, 81, 18]\}}\par
  Based on the above information, please provide the action for the next step to complete the task. Think, then act.
\end{minipage}

\vspace{8pt}
\noindent\textbf{Generation}\par
\begingroup\small
\texttt{<|think\_start|>}Based on the interaction history and current observation, I am in the middle of the last plan and will continue carrying it out.\texttt{<|think\_end|>}\par
\texttt{<|action\_start|>}[76, 69, 17, 0, 60, 0, 1]\texttt{<|action\_end|>}
\endgroup\newline

\end{tcolorbox}

\subsection{Trajectory-augmented Prior Dataset}
We provide an example of Trajectory-augmented Prior dataset for both EB-ALFRED and EB-Manipulation.
\begin{tcolorbox}[
  colback=gray!5!white, colframe=gray!75!black,
  title=Trajectory-augmented Prior for EB-ALFRED,
  boxrule=0.3mm, width=\textwidth, arc=3mm,
  auto outer arc=true, breakable
]

\textbf{Step 1:}\par\vspace{4pt}
\noindent
\begin{minipage}[t]{0.2\textwidth}
  \textbf{Input image}\par\vspace{4pt}
  \includegraphics[width=0.9\linewidth]{figures/alfred/episode_2_step_0.png}
\end{minipage}\hfill
\begin{minipage}[t]{0.79\textwidth}
  \textbf{Prompt}\par\vspace{4pt}
  \textbf{instruction:} Set plate with a spoon in it on the kitchen table\par
  \textbf{interaction\_history:} []\par
  Based on the above information, please provide the action for the next step to complete the task. Think, then act.
\end{minipage}

\vspace{8pt}
\noindent\textbf{Generation}\par
\begingroup\small
\texttt{<|think\_start|>}visual\_description: The image shows a corner of a room with beige walls and carpeted flooring. There appears to be a vent or air return mounted on the wall. The space appears to be empty without any visible furniture or objects. reasoning\_and\_reflection: Since the room appears empty, we'll need to search for and locate the required items (plate, spoon, and dining table) before we can complete the task. We'll need to find these items first, then arrange them according to the instruction. language\_plan: 1. First, locate the plate in the room. 2. Pick up the plate once found. 3. Find the spoon. 4. Pick up the spoon. 5. Find the dining table. 6. Put down the spoon on the plate. 7. Place the plate with the spoon on the dining table.\texttt{<|think\_end|>}\par
\texttt{<|action\_start|>}[31, 'find a Plate']\texttt{<|action\_end|>}
\endgroup\newline

\textbf{Step 2:}\par\vspace{4pt}
\noindent
\begin{minipage}[t]{0.2\textwidth}
  \textbf{Input image}\par\vspace{4pt}
  \includegraphics[width=0.9\linewidth]{figures/alfred/episode_2_step_1.png}
\end{minipage}\hfill
\begin{minipage}[t]{0.79\textwidth}
  \textbf{Prompt}\par\vspace{4pt}
  \textbf{instruction:} Set plate with a spoon in it on the kitchen table\par
  \textbf{interaction\_history:}\par
  {\small\ttfamily[\{``step\_id'': 0, ``thinking'': ``visual\_description: The image shows a corner of a room with beige walls and carpeted flooring. There appears to be a vent or air return mounted on the wall. The space appears to be empty without any visible furniture or objects. reasoning\_and\_reflection: Since the room appears empty, we'll need to search for and locate the required items (plate, spoon, and dining table) before we can complete the task. We'll need to find these items first, then arrange them according to the instruction. language\_plan: 1. First, locate the plate in the room. 2. Pick up the plate once found. 3. Find the spoon. 4. Pick up the spoon. 5. Find the dining table. 6. Put down the spoon on the plate. 7. Place the plate with the spoon on the dining table.'', ``action'': ``[31, 'find a Plate']'', ``env\_feedback'': ``Last action executed successfully.''\}]]}\par
  Based on the above information, please provide the action for the next step to complete the task. Think, then act.
\end{minipage}

\vspace{8pt}
\noindent\textbf{Generation}\par
\begingroup\small
\texttt{<|think\_start|>}visual\_description: The image shows a kitchen counter with a black plate, a knife, a toaster, a pan, and a small round object. The plate is visible and accessible. reasoning\_and\_reflection: The plate has been located on the kitchen counter. The next step is to pick up the plate as planned. After picking up the plate, the next task will be to find and pick up a spoon, then locate the dining table to complete the task. The previous plan is still valid, and we can continue executing it. language\_plan: 1. Pick up the plate. 2. Find the spoon. 3. Pick up the spoon. 4. Find the dining table. 5. Put down the spoon on the plate. 6. Place the plate with the spoon on the dining table.\texttt{<|think\_end|>}\par
\texttt{<|action\_start|>}[122, 'pick up the Plate']\texttt{<|action\_end|>}
\endgroup\newline

\textbf{Step 3:}\par\vspace{4pt}
\noindent
\begin{minipage}[t]{0.2\textwidth}
  \textbf{Input image}\par\vspace{4pt}
  \includegraphics[width=0.9\linewidth]{figures/alfred/episode_2_step_2.png}
\end{minipage}\hfill
\begin{minipage}[t]{0.79\textwidth}
  \textbf{Prompt}\par\vspace{4pt}
  \textbf{instruction:} Set plate with a spoon in it on the kitchen table\par
  \textbf{interaction\_history:}\par
  {\small\ttfamily[\{``step\_id'': 1, ``thinking'': ``visual\_description: The image shows a kitchen counter with a black plate, a knife, a toaster, a pan, and a small round object. The plate is visible and accessible. reasoning\_and\_reflection: The plate has been located on the kitchen counter. The next step is to pick up the plate as planned. After picking up the plate, the next task will be to find and pick up a spoon, then locate the dining table to complete the task. The previous plan is still valid, and we can continue executing it. language\_plan: 1. Pick up the plate. 2. Find the spoon. 3. Pick up the spoon. 4. Find the dining table. 5. Put down the spoon on the plate. 6. Place the plate with the spoon on the dining table.'', ``action'': ``[122, 'pick up the Plate']'', ``env\_feedback'': ``Last action executed successfully.''\}]]}\par
  Based on the above information, please provide the action for the next step to complete the task. Think, then act.
\end{minipage}

\vspace{8pt}
\noindent\textbf{Generation}\par
\begingroup\small
\texttt{<|think\_start|>}visual\_description: The image shows a kitchen counter with a knife, a toaster, a plate, and a bottle. The plate is visible on the counter. reasoning\_and\_reflection: The previous actions successfully located and picked up the plate. Now, the next step is to find a spoon. However, the image shows a knife instead of a spoon. We need to continue searching for the spoon. The dining table is not visible in the current image, so it will need to be located later. language\_plan: 1. Find a Spoon. 2. Pick up the Spoon. 3. Find the dining table. 4. Put down the Spoon on the Plate. 5. Place the Plate with the Spoon on the dining table.\texttt{<|think\_end|>}\par
\texttt{<|action\_start|>}[19, 'find a Spoon']\texttt{<|action\_end|>}
\endgroup\newline

\textbf{Step 4:}\par\vspace{4pt}
\noindent
\begin{minipage}[t]{0.2\textwidth}
  \textbf{Input image}\par\vspace{4pt}
  \includegraphics[width=0.9\linewidth]{figures/alfred/episode_2_step_3.png}
\end{minipage}\hfill
\begin{minipage}[t]{0.79\textwidth}
  \textbf{Prompt}\par\vspace{4pt}
  \textbf{instruction:} Set plate with a spoon in it on the kitchen table\par
  \textbf{interaction\_history:}\par
  {\small\ttfamily[\{``step\_id'': 2, ``thinking'': ``visual\_description: The image shows a kitchen counter with a knife, a toaster, a plate, and a bottle. The plate is visible on the counter. reasoning\_and\_reflection: The previous actions successfully located and picked up the plate. Now, the next step is to find a spoon. However, the image shows a knife instead of a spoon. We need to continue searching for the spoon. The dining table is not visible in the current image, so it will need to be located later. language\_plan: 1. Find a Spoon. 2. Pick up the Spoon. 3. Find the dining table. 4. Put down the Spoon on the Plate. 5. Place the Plate with the Spoon on the dining table.'', ``action'': ``[19, 'find a Spoon']'', ``env\_feedback'': ``Last action executed successfully.''\}]]}\par
  Based on the above information, please provide the action for the next step to complete the task. Think, then act.
\end{minipage}

\vspace{8pt}
\noindent\textbf{Generation}\par
\begingroup\small
\texttt{<|think\_start|>}visual\_description: The image shows a kitchen counter with a spoon, a plate, a green apple, and a coconut. The spoon is on the counter next to the plate. reasoning\_and\_reflection: The previous actions successfully located the plate and spoon. The next step is to pick up the spoon, which is visible on the counter. After picking up the spoon, the plan is to place it on the plate and then find the dining table to complete the task. language\_plan: 1. Pick up the spoon. 2. Place the spoon on the plate. 3. Find the dining table. 4. Place the plate with the spoon on the dining table.\texttt{<|think\_end|>}\par
\texttt{<|action\_start|>}[105, 'pick up the Spoon']\texttt{<|action\_end|>}
\endgroup\newline

\textbf{Step 5:}\par\vspace{4pt}
\noindent
\begin{minipage}[t]{0.2\textwidth}
  \textbf{Input image}\par\vspace{4pt}
  \includegraphics[width=0.9\linewidth]{figures/alfred/episode_2_step_4.png}
\end{minipage}\hfill
\begin{minipage}[t]{0.79\textwidth}
  \textbf{Prompt}\par\vspace{4pt}
  \textbf{instruction:} Set plate with a spoon in it on the kitchen table\par
  \textbf{interaction\_history:}\par
  {\small\ttfamily[\{``step\_id'': 3, ``thinking'': ``visual\_description: The image shows a kitchen counter with a spoon, a plate, a green apple, and a coconut. The spoon is on the counter next to the plate. reasoning\_and\_reflection: The previous actions successfully located the plate and spoon. The next step is to pick up the spoon, which is visible on the counter. After picking up the spoon, the plan is to place it on the plate and then find the dining table to complete the task. language\_plan: 1. Pick up the spoon. 2. Place the spoon on the plate. 3. Find the dining table. 4. Place the plate with the spoon on the dining table.'', ``action'': ``[105, 'pick up the Spoon']'', ``env\_feedback'': ``Last action is invalid. Robot is currently holding Plate''\}]]}\par
  Based on the above information, please provide the action for the next step to complete the task. Think, then act.
\end{minipage}

\vspace{8pt}
\noindent\textbf{Generation}\par
\begingroup\small
\texttt{<|think\_start|>}visual\_description: The image shows a kitchen countertop surface. There appears to be some objects on the counter, including what looks like an apple or round fruit. reasoning\_and\_reflection: From the interaction history, the robot successfully found and picked up a plate (Steps 0-1), then found a spoon (Step 2). However, when trying to pick up the spoon (Step 3), the action failed because the robot was already holding the plate. To achieve the goal of placing a plate with a spoon in it on the kitchen table, we need to first put down the plate, pick up the spoon, then put the spoon on the plate, pick up the plate again, and finally place it on the dining table. language\_plan: 1. Find the dining table to put down the plate temporarily. 2. Put down the plate. 3. Find the spoon again. 4. Pick up the spoon. 5. Find the plate. 6. Put the spoon on the plate. 7. Pick up the plate (now with spoon in it). 8. Find the dining table. 9. Put down the plate with spoon on the table.\texttt{<|think\_end|>}\par
\texttt{<|action\_start|>}[18, 'find a DiningTable']\texttt{<|action\_end|>}
\endgroup\newline

\textbf{Step 6:}\par\vspace{4pt}
\noindent
\begin{minipage}[t]{0.2\textwidth}
  \textbf{Input image}\par\vspace{4pt}
  \includegraphics[width=0.9\linewidth]{figures/alfred/episode_2_step_5.png}
\end{minipage}\hfill
\begin{minipage}[t]{0.79\textwidth}
  \textbf{Prompt}\par\vspace{4pt}
  \textbf{instruction:} Set plate with a spoon in it on the kitchen table\par
  \textbf{interaction\_history:}\par
  {\small\ttfamily[\{``step\_id'': 4, ``thinking'': ``visual\_description: The image shows a kitchen countertop surface. There appears to be some objects on the counter, including what looks like an apple or round fruit. reasoning\_and\_reflection: From the interaction history, the robot successfully found and picked up a plate (Steps 0-1), then found a spoon (Step 2). However, when trying to pick up the spoon (Step 3), the action failed because the robot was already holding the plate. To achieve the goal of placing a plate with a spoon in it on the kitchen table, we need to first put down the plate, pick up the spoon, then put the spoon on the plate, pick up the plate again, and finally place it on the dining table. language\_plan: 1. Find the dining table to put down the plate temporarily. 2. Put down the plate. 3. Find the spoon again. 4. Pick up the spoon. 5. Find the plate. 6. Put the spoon on the plate. 7. Pick up the plate (now with spoon in it). 8. Find the dining table. 9. Put down the plate with spoon on the table.'', ``action'': ``[18, 'find a DiningTable']'', ``env\_feedback'': ``Last action executed successfully.''\}]]}\par
  Based on the above information, please provide the action for the next step to complete the task. Think, then act.
\end{minipage}

\vspace{8pt}
\noindent\textbf{Generation}\par
\begingroup\small
\texttt{<|think\_start|>}visual\_description: The robot is near a table with a loaf of bread on it. The robot is holding a plate. reasoning\_and\_reflection: The robot successfully found the dining table and is ready to put down the plate. The next step is to put down the plate, find the spoon, and then place the spoon on the plate. Finally, the robot will pick up the plate with the spoon and place it on the table. language\_plan: 1. Put down the plate on the table. 2. Find the spoon. 3. Pick up the spoon. 4. Find the plate. 5. Put the spoon on the plate. 6. Pick up the plate with the spoon. 7. Find the dining table. 8. Put down the plate with the spoon on the table.\texttt{<|think\_end|>}\par
\texttt{<|action\_start|>}[133, 'put down the object in hand']\texttt{<|action\_end|>}
\endgroup\newline

\textbf{Step 7:}\par\vspace{4pt}
\noindent
\begin{minipage}[t]{0.2\textwidth}
  \textbf{Input image}\par\vspace{4pt}
  \includegraphics[width=0.9\linewidth]{figures/alfred/episode_2_step_6.png}
\end{minipage}\hfill
\begin{minipage}[t]{0.79\textwidth}
  \textbf{Prompt}\par\vspace{4pt}
  \textbf{instruction:} Set plate with a spoon in it on the kitchen table\par
  \textbf{interaction\_history:}\par
  {\small\ttfamily[\{``step\_id'': 5, ``thinking'': ``visual\_description: The robot is near a table with a loaf of bread on it. The robot is holding a plate. reasoning\_and\_reflection: The robot successfully found the dining table and is ready to put down the plate. The next step is to put down the plate, find the spoon, and then place the spoon on the plate. Finally, the robot will pick up the plate with the spoon and place it on the table. language\_plan: 1. Put down the plate on the table. 2. Find the spoon. 3. Pick up the spoon. 4. Find the plate. 5. Put the spoon on the plate. 6. Pick up the plate with the spoon. 7. Find the dining table. 8. Put down the plate with the spoon on the table.'', ``action'': ``[133, 'put down the object in hand']'', ``env\_feedback'': ``Last action executed successfully.''\}]]}\par
  Based on the above information, please provide the action for the next step to complete the task. Think, then act.
\end{minipage}

\vspace{8pt}
\noindent\textbf{Generation}\par
\begingroup\small
\texttt{<|think\_start|>}visual\_description: The plate is on the kitchen table next to a loaf of bread. There is a stool nearby. reasoning\_and\_reflection: The plate has been successfully placed on the kitchen table. The next step is to find and pick up a spoon to place it on the plate. The previous plan was interrupted after putting down the plate, so the next action is to locate the spoon and continue with the plan to place it on the plate. language\_plan: 1. Find the spoon. 2. Pick up the spoon. 3. Find the plate on the kitchen table. 4. Put the spoon on the plate. 5. Pick up the plate with the spoon in it. 6. Put down the plate with the spoon on the kitchen table.\texttt{<|think\_end|>}\par
\texttt{<|action\_start|>}[19, 'find a Spoon']\texttt{<|action\_end|>}
\endgroup\newline

\textbf{Step 8:}\par\vspace{4pt}
\noindent
\begin{minipage}[t]{0.2\textwidth}
  \textbf{Input image}\par\vspace{4pt}
  \includegraphics[width=0.9\linewidth]{figures/alfred/episode_2_step_7.png}
\end{minipage}\hfill
\begin{minipage}[t]{0.79\textwidth}
  \textbf{Prompt}\par\vspace{4pt}
  \textbf{instruction:} Set plate with a spoon in it on the kitchen table\par
  \textbf{interaction\_history:}\par
  {\small\ttfamily[\{``step\_id'': 6, ``thinking'': ``visual\_description: The plate is on the kitchen table next to a loaf of bread. There is a stool nearby. reasoning\_and\_reflection: The plate has been successfully placed on the kitchen table. The next step is to find and pick up a spoon to place it on the plate. The previous plan was interrupted after putting down the plate, so the next action is to locate the spoon and continue with the plan to place it on the plate. language\_plan: 1. Find the spoon. 2. Pick up the spoon. 3. Find the plate on the kitchen table. 4. Put the spoon on the plate. 5. Pick up the plate with the spoon in it. 6. Put down the plate with the spoon on the kitchen table.'', ``action'': ``[19, 'find a Spoon']'', ``env\_feedback'': ``Last action executed successfully.''\}]]}\par
  Based on the above information, please provide the action for the next step to complete the task. Think, then act.
\end{minipage}

\vspace{8pt}
\noindent\textbf{Generation}\par
\begingroup\small
\texttt{<|think\_start|>}visual\_description: The kitchen counter has a spoon, a fork, a green apple, and a brown object. The sink is visible, and there are some utensils near it. reasoning\_and\_reflection: The previous plan involved putting down the plate to pick up the spoon. The first three actions were executed successfully, and the robot is now ready to pick up the spoon. The next steps involve placing the spoon on the plate, picking up the plate with the spoon, and placing it on the dining table. language\_plan: 1. Pick up the spoon. 2. Find the plate. 3. Put the spoon on the plate. 4. Pick up the plate (now with spoon in it). 5. Find the dining table. 6. Put down the plate with spoon on the table.\texttt{<|think\_end|>}\par
\texttt{<|action\_start|>}[105, 'pick up the Spoon']\texttt{<|action\_end|>}
\endgroup\newline

\textbf{Step 9:}\par\vspace{4pt}
\noindent
\begin{minipage}[t]{0.2\textwidth}
  \textbf{Input image}\par\vspace{4pt}
  \includegraphics[width=0.9\linewidth]{figures/alfred/episode_2_step_8.png}
\end{minipage}\hfill
\begin{minipage}[t]{0.79\textwidth}
  \textbf{Prompt}\par\vspace{4pt}
  \textbf{instruction:} Set plate with a spoon in it on the kitchen table\par
  \textbf{interaction\_history:}\par
  {\small\ttfamily[\{``step\_id'': 7, ``thinking'': ``visual\_description: The kitchen counter has a spoon, a fork, a green apple, and a brown object. The sink is visible, and there are some utensils near it. reasoning\_and\_reflection: The previous plan involved putting down the plate to pick up the spoon. The first three actions were executed successfully, and the robot is now ready to pick up the spoon. The next steps involve placing the spoon on the plate, picking up the plate with the spoon, and placing it on the dining table. language\_plan: 1. Pick up the spoon. 2. Find the plate. 3. Put the spoon on the plate. 4. Pick up the plate (now with spoon in it). 5. Find the dining table. 6. Put down the plate with spoon on the table.'', ``action'': ``[105, 'pick up the Spoon']'', ``env\_feedback'': ``Last action executed successfully.''\}]]}\par
  Based on the above information, please provide the action for the next step to complete the task. Think, then act.
\end{minipage}

\vspace{8pt}
\noindent\textbf{Generation}\par
\begingroup\small
\texttt{<|think\_start|>}visual\_description: The image shows a kitchen counter with a spoon, fork, and a green object on a tray. There is also a brown object and a green apple on the counter. reasoning\_and\_reflection: The previous plan involved finding the plate to place the spoon on it. The spoon is already visible on a tray, which might be the plate. The next step is to confirm the presence of the plate and place the spoon on it. The plan needs to be adjusted to ensure the spoon is placed on the plate and then the plate is moved to the dining table. language\_plan: 1. Find the plate (tray) on the counter. 2. Put the spoon on the plate. 3. Pick up the plate with the spoon. 4. Find the dining table. 5. Put down the plate with the spoon on the dining table.\texttt{<|think\_end|>}\par
\texttt{<|action\_start|>}[31, 'find a Plate']\texttt{<|action\_end|>}
\endgroup\newline

\textbf{Step 10:}\par\vspace{4pt}
\noindent
\begin{minipage}[t]{0.2\textwidth}
  \textbf{Input image}\par\vspace{4pt}
  \includegraphics[width=0.9\linewidth]{figures/alfred/episode_2_step_9.png}
\end{minipage}\hfill
\begin{minipage}[t]{0.79\textwidth}
  \textbf{Prompt}\par\vspace{4pt}
  \textbf{instruction:} Set plate with a spoon in it on the kitchen table\par
  \textbf{interaction\_history:}\par
  {\small\ttfamily[\{``step\_id'': 8, ``thinking'': ``visual\_description: The image shows a kitchen counter with a spoon, fork, and a green object on a tray. There is also a brown object and a green apple on the counter. reasoning\_and\_reflection: The previous plan involved finding the plate to place the spoon on it. The spoon is already visible on a tray, which might be the plate. The next step is to confirm the presence of the plate and place the spoon on it. The plan needs to be adjusted to ensure the spoon is placed on the plate and then the plate is moved to the dining table. language\_plan: 1. Find the plate (tray) on the counter. 2. Put the spoon on the plate. 3. Pick up the plate with the spoon. 4. Find the dining table. 5. Put down the plate with the spoon on the dining table.'', ``action'': ``[31, 'find a Plate']'', ``env\_feedback'': ``Last action executed successfully.''\}]]}\par
  Based on the above information, please provide the action for the next step to complete the task. Think, then act.
\end{minipage}

\vspace{8pt}
\noindent\textbf{Generation}\par
\begingroup\small
\texttt{<|think\_start|>}visual\_description: The kitchen table has a loaf of bread, a plate, and a spoon on it. The spoon is not on the plate. reasoning\_and\_reflection: The robot has successfully found the dining table and put down the plate. It then found and picked up the spoon. The next step is to place the spoon on the plate, pick up the plate with the spoon, and place it on the table. The previous plan was interrupted after picking up the spoon, so the next step is to put the spoon on the plate. language\_plan: 1. Find the plate. 2. Put the spoon on the plate. 3. Pick up the plate with the spoon. 4. Find the kitchen table. 5. Put down the plate with the spoon on the table.\texttt{<|think\_end|>}\par
\texttt{<|action\_start|>}[133, 'put down the object in hand']\texttt{<|action\_end|>}
\endgroup\newline

\end{tcolorbox}

\begin{tcolorbox}[
  colback=gray!5!white, colframe=gray!75!black,
  title=Trajectory-augmented Prior for EB-Manipulation,
  boxrule=0.3mm, width=\textwidth, arc=3mm,
  auto outer arc=true, breakable
]

\textbf{Step 1:}\par\vspace{4pt}
\noindent
\begin{minipage}[t]{0.2\textwidth}
  \textbf{Input image}\par\vspace{4pt}
  \includegraphics[width=0.9\linewidth]{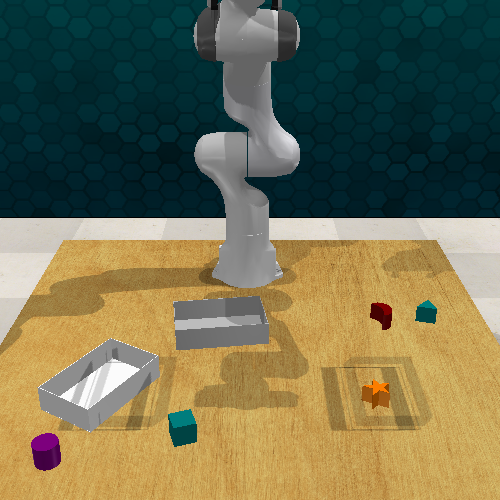}
\end{minipage}\hfill
\begin{minipage}[t]{0.79\textwidth}
  \textbf{Prompt}\par\vspace{4pt}
  \textbf{instruction:} Pick up the star and place it into the silver container.\par
  \textbf{interaction\_history:} []\par
  \textbf{additional\_info:} {\ttfamily\{``object 1'': [73, 15, 18], ``object 2'': [57, 20, 18], ``object 3'': [66, 38, 18], ``object 4'': [33, 43, 17], ``object 5'': [57, 74, 17], ``object 6'': [28, 80, 17], ``object 7'': [26, 91, 17]\}}\par
  Based on the above information, please provide the action for the next step to complete the task. Think, then act.
\end{minipage}

\vspace{8pt}
\noindent\textbf{Generation}\par
\begingroup\small
\texttt{<|think\_start|>}visual\_description: From left to right, I can see a purple cylinder at [73, 15, 18], a white container at [57, 20, 18], a teal cube at [66, 38, 18], a silver container at [33, 43, 17], an orange star at [57, 74, 17], a maroon moon at [28, 80, 17], and a teal triangular at [26, 91, 17]. reasoning\_and\_reflection: The task requires picking up the orange star located at [57, 74, 17] and placing it into the silver container at [33, 43, 17]. I will need to move the gripper to the star position, grasp it, lift it up, move it to the silver container position, and then release it. language\_plan: 1. Move gripper above the orange star\texttt{\textbackslash n}2. Lower gripper and open it to prepare for grasping\texttt{\textbackslash n}3. Close gripper to grasp the star\texttt{\textbackslash n}4. Lift the star up\texttt{\textbackslash n}5. Move the star above the silver container\texttt{\textbackslash n}6. Lower the star into the container\texttt{\textbackslash n}7. Open gripper to release the star\texttt{<|think\_end|>}\par
\texttt{<|action\_start|>}[57, 74, 27, 0, 60, 90, 1]\texttt{<|action\_end|>}
\endgroup\newline

\textbf{Step 2:}\par\vspace{4pt}
\noindent
\begin{minipage}[t]{0.2\textwidth}
  \textbf{Input image}\par\vspace{4pt}
  \includegraphics[width=0.9\linewidth]{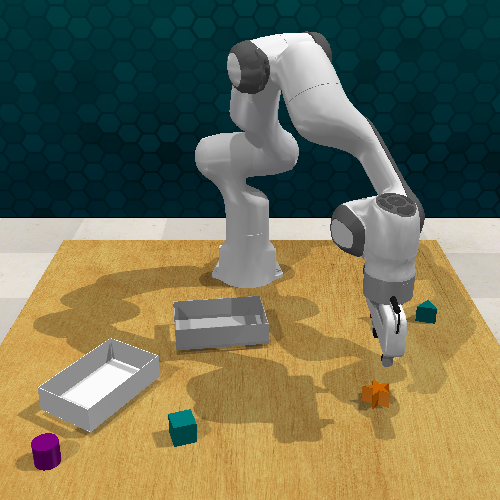}
\end{minipage}\hfill
\begin{minipage}[t]{0.79\textwidth}
  \textbf{Prompt}\par\vspace{4pt}
  \textbf{instruction:} Pick up the star and place it into the silver container.\par
  \textbf{interaction\_history:}\par
  {\small\ttfamily[\{``step\_id'': 0, ``thinking'': ``visual\_description: From left to right, I can see a purple cylinder at [73, 15, 18], a white container at [57, 20, 18], a teal cube at [66, 38, 18], a silver container at [33, 43, 17], an orange star at [57, 74, 17], a maroon moon at [28, 80, 17], and a teal triangular at [26, 91, 17]. reasoning\_and\_reflection: The task requires picking up the orange star located at [57, 74, 17] and placing it into the silver container at [33, 43, 17]. I will need to move the gripper to the star position, grasp it, lift it up, move it to the silver container position, and then release it. language\_plan: 1. Move gripper above the orange star\texttt{\textbackslash n}2. Lower gripper and open it to prepare for grasping\texttt{\textbackslash n}3. Close gripper to grasp the star\texttt{\textbackslash n}4. Lift the star up\texttt{\textbackslash n}5. Move the star above the silver container\texttt{\textbackslash n}6. Lower the star into the container\texttt{\textbackslash n}7. Open gripper to release the star'', ``action'': ``[57, 74, 27, 0, 60, 90, 1]'', ``env\_feedback'': ``Last action was successful.''\}]]}\par
  \textbf{additional\_info:} {\ttfamily\{``object 1'': [73, 15, 18], ``object 2'': [55, 21, 17], ``object 3'': [66, 38, 18], ``object 4'': [34, 46, 18], ``object 5'': [56, 74, 17], ``object 6'': [26, 79, 17], ``object 7'': [26, 91, 17]\}}\par
  Based on the above information, please provide the action for the next step to complete the task. Think, then act.
\end{minipage}

\vspace{8pt}
\noindent\textbf{Generation}\par
\begingroup\small
\texttt{<|think\_start|>}visual\_description: From left to right, I can see a purple cylinder at [73, 15, 18], a white container at [55, 21, 17], a teal cube at [66, 38, 18], a silver container at [34, 46, 18], an orange star at [56, 74, 17], a maroon moon at [26, 79, 17], and a teal triangular at [26, 91, 17]. reasoning\_and\_reflection: The task involves picking up the orange star and placing it into the silver container. I have already moved the gripper above the star. The next step is to lower the gripper and open it to prepare for grasping the star. After that, I will close the gripper to grasp the star, lift it, move it to the container, and release it. I need to ensure that each step is executed precisely to avoid dropping the star or missing the container. language\_plan: 1. Lower gripper and open it to prepare for grasping\texttt{\textbackslash n}2. Close gripper to grasp the star\texttt{\textbackslash n}3. Lift the star up\texttt{\textbackslash n}4. Move the star above the silver container\texttt{\textbackslash n}5. Lower the star into the container\texttt{\textbackslash n}6. Open gripper to release the star\texttt{<|think\_end|>}\par
\texttt{<|action\_start|>}[57, 74, 17, 0, 60, 90, 0]\texttt{<|action\_end|>}
\endgroup\newline

\textbf{Step 3:}\par\vspace{4pt}
\noindent
\begin{minipage}[t]{0.2\textwidth}
  \textbf{Input image}\par\vspace{4pt}
  \includegraphics[width=0.9\linewidth]{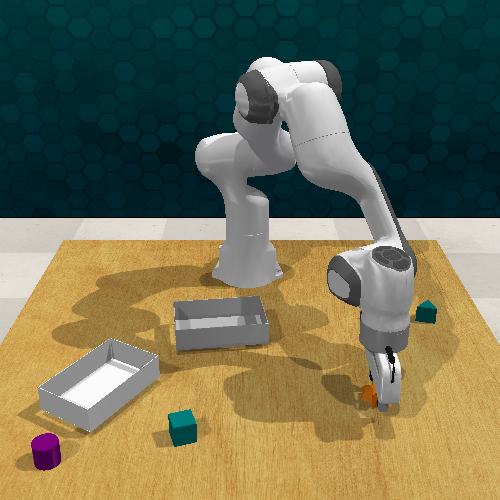}
\end{minipage}\hfill
\begin{minipage}[t]{0.79\textwidth}
  \textbf{Prompt}\par\vspace{4pt}
  \textbf{instruction:} Pick up the star and place it into the silver container.\par
  \textbf{interaction\_history:}\par
  {\small\ttfamily[\{``step\_id'': 1, ``thinking'': ``visual\_description: From left to right, I can see a purple cylinder at [73, 15, 18], a white container at [55, 21, 17], a teal cube at [66, 38, 18], a silver container at [34, 46, 18], an orange star at [56, 74, 17], a maroon moon at [26, 79, 17], and a teal triangular at [26, 91, 17]. reasoning\_and\_reflection: The task involves picking up the orange star and placing it into the silver container. I have already moved the gripper above the star. The next step is to lower the gripper and open it to prepare for grasping the star. After that, I will close the gripper to grasp the star, lift it, move it to the container, and release it. I need to ensure that each step is executed precisely to avoid dropping the star or missing the container. language\_plan: 1. Lower gripper and open it to prepare for grasping\texttt{\textbackslash n}2. Close gripper to grasp the star\texttt{\textbackslash n}3. Lift the star up\texttt{\textbackslash n}4. Move the star above the silver container\texttt{\textbackslash n}5. Lower the star into the container\texttt{\textbackslash n}6. Open gripper to release the star'', ``action'': ``[57, 74, 17, 0, 60, 90, 0]'', ``env\_feedback'': ``Last action was successful.''\}]]}\par
  \textbf{additional\_info:} {\ttfamily\{``object 1'': [73, 15, 18], ``object 2'': [55, 21, 17], ``object 3'': [66, 38, 18], ``object 4'': [32, 40, 17], ``object 5'': [56, 73, 16], ``object 6'': [27, 81, 18], ``object 7'': [26, 91, 17]\}}\par
  Based on the above information, please provide the action for the next step to complete the task. Think, then act.
\end{minipage}

\vspace{8pt}
\noindent\textbf{Generation}\par
\begingroup\small
\texttt{<|think\_start|>}visual\_description: From left to right, I can see a purple cylinder at [73, 15, 18], a white container at [55, 21, 17], a teal cube at [66, 38, 18], a silver container at [32, 40, 17], an orange star at [56, 73, 16], a maroon moon at [27, 81, 18], and a teal triangular at [26, 91, 17]. reasoning\_and\_reflection: The task involves transferring the star from its current position to the silver container. I have already moved the gripper above the star and lowered it to grasp the star. The next step is to close the gripper to secure the star, lift it, move it to the container, and release it. I need to ensure that the gripper is properly aligned and that the star is securely held before moving it to avoid dropping it. language\_plan: 1. Close gripper to grasp the star\texttt{\textbackslash n}2. Lift the star up\texttt{\textbackslash n}3. Move the star above the silver container\texttt{\textbackslash n}4. Lower the star into the container\texttt{\textbackslash n}5. Open gripper to release the star\texttt{<|think\_end|>}\par
\texttt{<|action\_start|>}[57, 74, 25, 0, 60, 90, 0]\texttt{<|action\_end|>}
\endgroup\newline

\textbf{Step 4:}\par\vspace{4pt}
\noindent
\begin{minipage}[t]{0.2\textwidth}
  \textbf{Input image}\par\vspace{4pt}
  \includegraphics[width=0.9\linewidth]{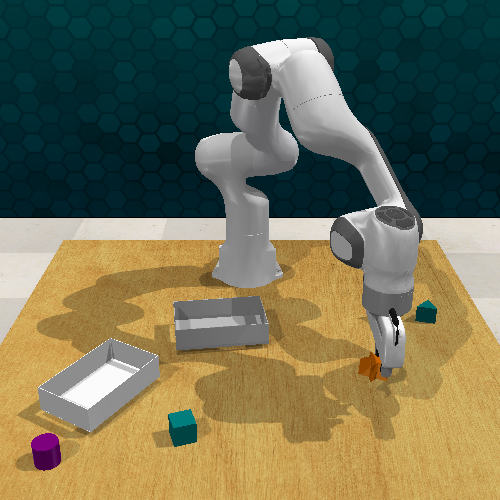}
\end{minipage}\hfill
\begin{minipage}[t]{0.79\textwidth}
  \textbf{Prompt}\par\vspace{4pt}
  \textbf{instruction:} Pick up the star and place it into the silver container.\par
  \textbf{interaction\_history:}\par
  {\small\ttfamily[\{``step\_id'': 2, ``thinking'': ``visual\_description: From left to right, I can see a purple cylinder at [73, 15, 18], a white container at [55, 21, 17], a teal cube at [66, 38, 18], a silver container at [32, 40, 17], an orange star at [56, 73, 16], a maroon moon at [27, 81, 18], and a teal triangular at [26, 91, 17]. reasoning\_and\_reflection: The task involves transferring the star from its current position to the silver container. I have already moved the gripper above the star and lowered it to grasp the star. The next step is to close the gripper to secure the star, lift it, move it to the container, and release it. I need to ensure that the gripper is properly aligned and that the star is securely held before moving it to avoid dropping it. language\_plan: 1. Close gripper to grasp the star\texttt{\textbackslash n}2. Lift the star up\texttt{\textbackslash n}3. Move the star above the silver container\texttt{\textbackslash n}4. Lower the star into the container\texttt{\textbackslash n}5. Open gripper to release the star'', ``action'': ``[57, 74, 25, 0, 60, 90, 0]'', ``env\_feedback'': ``Last action was successful.''\}]]}\par
  \textbf{additional\_info:} {\ttfamily\{``object 1'': [73, 15, 18], ``object 2'': [55, 21, 17], ``object 3'': [66, 38, 18], ``object 4'': [31, 40, 18], ``object 5'': [58, 72, 24], ``object 6'': [27, 81, 18], ``object 7'': [26, 91, 17]\}}\par
  Based on the above information, please provide the action for the next step to complete the task. Think, then act.
\end{minipage}

\vspace{8pt}
\noindent\textbf{Generation}\par
\begingroup\small
\texttt{<|think\_start|>}visual\_description: From left to right, I can see a purple cylinder at [73, 15, 18], a white container at [55, 21, 17], a teal cube at [66, 38, 18], a silver container at [31, 40, 18], an orange star at [58, 72, 24], a maroon moon at [27, 81, 18], and a teal triangular at [26, 91, 17]. reasoning\_and\_reflection: The task is to place the star into the silver container. I have already moved the gripper to the star's position, grasped it, and lifted it up. The next step is to move the star to the position above the silver container, lower it into the container, and release it. I need to ensure the star is securely placed inside the container before releasing it. language\_plan: 1. Move the star above the silver container\texttt{\textbackslash n}2. Lower the star into the container\texttt{\textbackslash n}3. Open gripper to release the star\texttt{<|think\_end|>}\par
\texttt{<|action\_start|>}[33, 43, 25, 0, 60, 90, 0]\texttt{<|action\_end|>}
\endgroup\newline

\textbf{Step 5:}\par\vspace{4pt}
\noindent
\begin{minipage}[t]{0.2\textwidth}
  \textbf{Input image}\par\vspace{4pt}
  \includegraphics[width=0.9\linewidth]{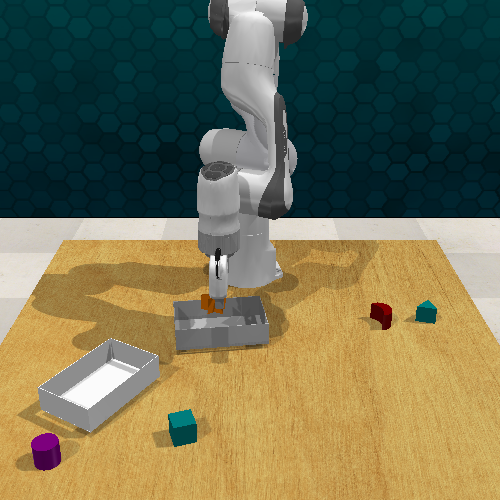}
\end{minipage}\hfill
\begin{minipage}[t]{0.79\textwidth}
  \textbf{Prompt}\par\vspace{4pt}
  \textbf{instruction:} Pick up the star and place it into the silver container.\par
  \textbf{interaction\_history:}\par
  {\small\ttfamily[\{``step\_id'': 3, ``thinking'': ``visual\_description: From left to right, I can see a purple cylinder at [73, 15, 18], a white container at [55, 21, 17], a teal cube at [66, 38, 18], a silver container at [31, 40, 18], an orange star at [58, 72, 24], a maroon moon at [27, 81, 18], and a teal triangular at [26, 91, 17]. reasoning\_and\_reflection: The task is to place the star into the silver container. I have already moved the gripper to the star's position, grasped it, and lifted it up. The next step is to move the star to the position above the silver container, lower it into the container, and release it. I need to ensure the star is securely placed inside the container before releasing it. language\_plan: 1. Move the star above the silver container\texttt{\textbackslash n}2. Lower the star into the container\texttt{\textbackslash n}3. Open gripper to release the star'', ``action'': ``[33, 43, 25, 0, 60, 90, 0]'', ``env\_feedback'': ``Last action was successful.''\}]]}\par
  \textbf{additional\_info:} {\ttfamily\{``object 1'': [73, 15, 18], ``object 2'': [55, 21, 17], ``object 3'': [66, 38, 18], ``object 4'': [34, 41, 25], ``object 5'': [32, 42, 17], ``object 6'': [27, 80, 18], ``object 7'': [26, 91, 17]\}}\par
  Based on the above information, please provide the action for the next step to complete the task. Think, then act.
\end{minipage}

\vspace{8pt}
\noindent\textbf{Generation}\par
\begingroup\small
\texttt{<|think\_start|>}visual\_description: From left to right, I can see a purple cylinder at [73, 15, 18], a white container at [55, 21, 17], a teal cube at [66, 38, 18], an orange star at [34, 41, 25], a silver container at [32, 42, 17], a maroon moon at [27, 80, 18], and a teal triangular at [26, 91, 17]. reasoning\_and\_reflection: The task is to place the star into the silver container. I have already picked up the star and moved it above the container. The next steps involve lowering the star into the container and releasing it. I need to ensure the gripper is positioned correctly above the container before lowering it. Once the star is inside the container, I will open the gripper to release it and then move the gripper back to a neutral position. language\_plan: 1. Move the star above the silver container\texttt{\textbackslash n}2. Lower the star into the container\texttt{\textbackslash n}3. Open gripper to release the star\texttt{\textbackslash n}4. Move gripper back to a neutral position\texttt{<|think\_end|>}\par
\texttt{<|action\_start|>}[33, 43, 17, 0, 60, 90, 1]\texttt{<|action\_end|>}
\endgroup\newline

\end{tcolorbox}

\subsection{Environment-Anchored Prior Dataset}
\label{app:app_env_anchored}
We provide an example of Environment-Anchored Prior dataset for both EB-ALFRED and EB-Manipulation.
\begin{tcolorbox}[colback=gray!5!white, colframe=gray!75!black, 
title=Environment-Anchored Prior Dataset for EB-ALFRED (Masked Action Modeling), boxrule=0.3mm, width=\textwidth, arc=3mm, auto outer arc=true]
\textbf{Prompt}

You are a household assistant. You are given an instruction: ``Place the clean bar soap from the bin into the tub.'' and an incomplete action sequence: [``find a garbagecan'', ``pick up the soapbar'', ``find a sinkbasin'', ``[MASK]'', ``find a Faucet'', ``turn on the Faucet'', ``turn off the Faucet'', ``find a soapbar'', ``pick up the soapbar'', ``find a bathtubbasin'', ``put down the object in hand'']. Please identify the missing action to complete the sequence.\newline

\textbf{Generation}

To complete the logical chain for the instruction ``Place the clean bar soap from the bin into the tub.'', we need to ensure that the sequence of actions aligns with the goal of cleaning the soap and then placing it in the tub. Let's break down the sequence step by step:\newline\newline
1. **Find a garbagecan**: This step is likely to ensure that any debris or used soap can be discarded if necessary.\newline
2. **Pick up the soapbar**: This is the initial step to get the soap ready for cleaning.\newline
3. **Find a sinkbasin**: The sinkbasin is where the soap will be cleaned.\newline
4. **[MASK]**: This is the missing action that needs to be identified.\newline
5. **Find a Faucet**: The faucet is needed to provide water for cleaning the soap.\newline
6. **Turn on the Faucet**: This action provides the water necessary for cleaning.\newline
7. **Turn off the Faucet**: This action stops the water flow after the soap is cleaned.\newline
8. **Find a soapbar**: This step is redundant because the soapbar was already picked up earlier. However, it might be included to ensure the soap is still in hand.\newline
9. **Pick up the soapbar**: This step is also redundant but ensures the soap is ready to be moved.\newline
10. **Find a bathtubbasin**: The bathtubbasin is the final destination for the clean soap.\newline
11. **Put down the object in hand**: This action places the clean soap in the bathtubbasin.\newline\newline
Given the context, the missing action should logically be an action that prepares the soap for cleaning. The most appropriate action to fill the gap is ``put down the object in hand''. This ensures that the soap is placed in the sinkbasin before the water is turned on, allowing it to be properly cleaned. After the soap is cleaned and the faucet is turned off, the sequence continues with finding the bathtubbasin and placing the clean soap in it. Therefore, the missing action is ``put down the object in hand''. To complete the logical chain for the instruction ``Place the clean bar soap from the bin into the tub.'', the missing action is ``put down the object in hand''. The complete and correct sequence is: [``find a garbagecan'', ``pick up the soapbar'', ``find a sinkbasin'', ``put down the object in hand'', ``find a Faucet'', ``turn on the Faucet'', ``turn off the Faucet'', ``find a soapbar'', ``pick up the soapbar'', ``find a bathtubbasin'', ``put down the object in hand''].
\end{tcolorbox}

\begin{tcolorbox}[colback=gray!5!white, colframe=gray!75!black, 
title=Environment-Anchored Prior Dataset for EB-ALFRED (Action Sequence Reordering), boxrule=0.3mm, width=\textwidth, arc=3mm, auto outer arc=true]
\textbf{Prompt}

You are a household assistant. You are given the instruction: ``Relocate the remote from the chair to the lamp.'' The randomized action sequences are [``pick up the remotecontrol'', ``find a floorlamp'', ``find an armchair'', ``turn on the floorlamp'']. Your task is to generate the correct sequence of actions to accomplish the instruction.\newline

\textbf{Generation}

To accomplish the task of relocating the remote from the chair to the lamp, I need to follow a logical sequence of actions. First, I must locate the armchair where the remote is currently placed. Once I have found the armchair, I can proceed to pick up the remote control from it. Next, I need to find the floorlamp where I will place the remote. Finally, to ensure the lamp is functional and the remote is easily accessible, I will turn on the floorlamp. This sequence ensures that I complete the task as instructed, moving the remote from the chair to the lamp and verifying that the lamp is operational. 

As a household assistant, to accomplish the instruction ``Relocate the remote from the chair to the lamp.'', the correct sequence of actions is: [``find an armchair'', ``pick up the remotecontrol'', ``find a floorlamp'', ``turn on the floorlamp''].
\end{tcolorbox}

\begin{tcolorbox}[colback=gray!5!white, colframe=gray!75!black, 
title=Environment-Anchored Prior Dataset for EB-Manipulation (Absolute Coordinate Grounding), boxrule=0.3mm, width=\textwidth, arc=3mm, auto outer arc=true]
\textbf{Input image:}
\begin{center}
\includegraphics[width=0.2\textwidth]{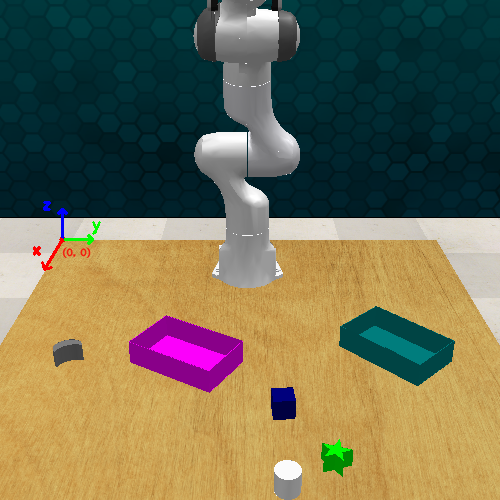}
\end{center}

\textbf{Example 1:}

\textbf{Prompt}

What is the 3D coordinate of the gray moon?
\newline

\textbf{Generation}

[42, 11, 17]
\newline

\textbf{Example 2:}

\textbf{Prompt}

What object is located at [42, 11, 17]?
\newline

\textbf{Generation}

The gray moon
\end{tcolorbox}

\begin{tcolorbox}[colback=gray!5!white, colframe=gray!75!black, 
title=Environment-Anchored Prior Dataset for EB-Manipulation (Relative Coordinate Grounding), boxrule=0.3mm, width=\textwidth, arc=3mm, auto outer arc=true]
\textbf{Input image:}
\begin{center}
\includegraphics[width=0.2\textwidth]{figures/spatial_grounding_example.png}
\end{center}

\textbf{Example 1:}

\textbf{Prompt}

What is the 3D location of the leftmost object?
\newline

\textbf{Generation}

[42, 11, 17]
\newline

\textbf{Example 2:}

\textbf{Prompt}

What is the 3D location of the rightmost object?"
\newline

\textbf{Generation}

[41, 80, 18]
\end{tcolorbox}

\begin{tcolorbox}[colback=gray!5!white, colframe=gray!75!black, 
title=Environment-Anchored Prior Dataset for EB-Manipulation (Combination of Absolute \& Relative Grounding), boxrule=0.3mm, width=\textwidth, arc=3mm, auto outer arc=true]
\textbf{Input image:}
\begin{center}
\includegraphics[width=0.2\textwidth]{figures/spatial_grounding_example.png}
\end{center}

\textbf{Example 1:}

\textbf{Prompt}

Is the object located at [42, 11, 17] the leftmost in the scene?
\newline

\textbf{Generation}

Yes
\newline

\textbf{Example 2:}

\textbf{Prompt}

Is the object located at [42, 11, 17] the rightmost in the scene?
\newline

\textbf{Generation}

No
\end{tcolorbox}

\subsection{External Knowledge Prior Dataset}
We provide an example of External Knowledge Prior dataset for both EB-ALFRED and EB-Manipulation.
\begin{tcolorbox}[colback=gray!5!white, colframe=gray!75!black, 
title=External Knowledge Prior Dataset for EB-ALFRED (OpenO1-SFT), boxrule=0.3mm, width=\textwidth, arc=3mm, auto outer arc=true]
\textbf{Prompt}

\#\# You are a robot who is great at reasoning. Given a instruction, you must generate a thought process and then an output. 

Question: To find the number of positive five-digit numbers that have a 1, 3, 7, or 5 in them.\newline

\textbf{Generation}\newline
Alright, I need to find the number of positive five-digit numbers that have a 1, 3, 5, or 7 in them. First, let's clarify the problem to ensure I understand it correctly. It seems that the task is to count all five-digit numbers where at least one of the digits is either 1, 3, 5, or 7. \newline
To approach this, I can consider the total number of five-digit numbers and then subtract those that do not contain any of the digits 1, 3, 5, or 7. \newline\newline
**Step 1: Calculate the Total Number of Five-Digit Numbers**\newline\newline
Five-digit numbers range from 10000 to 99999. To find the total number of five-digit numbers:\newline\newline
- The first digit (ten-thousands place) can be from 1 to 9 (since it cannot be 0), giving 9 options.\newline
- The remaining four digits (thousands, hundreds, tens, and ones places) can each be from 0 to 9, giving 10 options each.\newline\newline
Total five-digit numbers = $9 \times 10 \times 10 \times 10 \times 10 = 90,000$\newline\newline
**Step 2: Calculate the Number of Five-Digit Numbers Without 1, 3, 5, or 7**\newline\newline
If a number does not contain any of the digits 1, 3, 5, or 7, the allowed digits are 0, 2, 4, 6, 8, and 9.\newline\newline
- The first digit cannot be 0 (to maintain a five-digit number), so it can be 2, 4, 6, 8, or 9. That's 5 options.\newline
- The remaining four digits can be 0, 2, 4, 6, 8, or 9, giving 6 options each.\newline\newline
Number of five-digit numbers without 1, 3, 5, or 7 = $5 \times 6 \times 6 \times 6 \times 6 = 5 \times 6^4$\newline\newline
Calculating $6^4$:\newline
$6^2 = 36$ \newline
$6^3 = 216$ \newline
$6^4 = 1296$\newline\newline
So, number = $5 \times 1296 = 6480$\newline\newline
**Step 3: Subtract to Find the Desired Number**\newline
Number of five-digit numbers with at least one digit being 1, 3, 5, or 7 = Total five-digit numbers - Numbers without 1, 3, 5, or 7\newline\newline
$= 90,000 - 6,480 = 83,520$ \newline
But wait, let's double-check to ensure there are no mistakes in the calculation. \newline\newline
**Verification:**\newline
Total five-digit numbers: 90,000 \newline
Numbers without 1, 3, 5, or 7: 5 (first digit) $\times 6^4$ (remaining digits) $= 5 \times 1296 = 6480$\newline
Subtracting: $90,000 - 6,480 = 83,520$\newline
Everything seems correct. There are no overlapping cases or double-counting issues since the approach uses complementary counting.

\end{tcolorbox}

\begin{tcolorbox}[colback=gray!5!white, colframe=gray!75!black, 
title=External Knowledge Prior Dataset for EB-Manipulation (Spacethinker), boxrule=0.3mm, width=\textwidth, arc=3mm, auto outer arc=true]
\textbf{Input image}
\begin{center}
\includegraphics[width=0.5\textwidth]{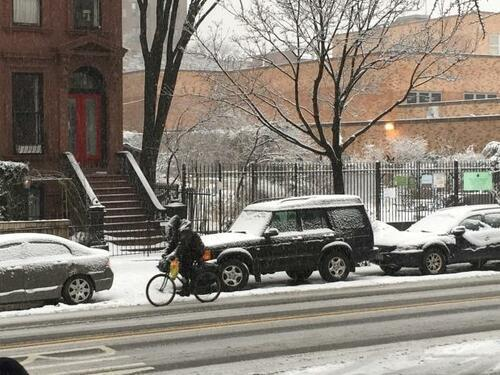}
\end{center}

\textbf{Prompt}

\#\# You are a robot who is great at spatial reasoning. Given a instruction, you must generate a thought process and then an output.\newline
Question: How far apart are the snow-covered parked cars and the snowy street with parked cars?
\newline

\textbf{Generation}

I see a street covered with snow, and several parked cars also dusted with snow. Additionally, there is a cyclist riding along the road. The snow covers both the cars and the roadway, suggesting a contiguous snowy scene. Now, determining the distance between “snow-covered parked cars” and the \"snowy street with parked cars\" suggests considering the separation between two distinct areas within this same scene.

Since both the cars and the street are snow-covered, it might seem as if they are tightly connected in context; however, finding a distance here usually involves looking at the lateral space due to the lack of any actual intervening physical barrier or wide expanse in the image. I estimate this minimal distance to be the space separating the parked cars from the street edge directly in front of them, without a significant gap in the visible scene.

Given these observations, the notion of any small distance like 20.33 inches feels particularly plausible, most likely representing the buffer or visual gap where the snow transitions from the cars to the street with virtually no space. This would particularly make sense if the focus were on something like the offset from the curb which is often minor.

Therefore, although exactness in this scenario might be less definitive visually, conceptually a small distance effectively represents the snow transition directly from car to the street, accounting for negligible separation.

There is essentially no significant distance between the snow-covered parked cars and the snowy street with parked cars, with the understanding that 20.33 inches could be seen as the minimal or visual margin separating the snow on the cars from the street-level snow.
\end{tcolorbox}

\subsection{Prompt for Generating Reasoning Augmentation}
To generate the trajectory-augmented priors as described in Section~\ref{sec:traj}, we leverage GPT-4o, to enrich raw trajectories with structured reasoning. This process involves prompting the model to generate a visual description, a reflection on past actions, and a language plan for following steps. The following prompt is used to elicit these detailed reasoning traces from GPT-4o, transforming action-only sequences into knowledge-rich data for Embodied Prior Learning.
\label{app:4o_prompt}

\begin{tcolorbox}[
  colback=gray!5!white,
  colframe=gray!75!black,
  title=Prompt,
  boxrule=0.3mm,
  width=\textwidth,
  arc=3mm,
  auto outer arc=true
]
For the following task: \{\} \\[4pt]
You have generated the following multi-step plan to complete the task: \\\{\} \\[4pt]
You have executed the first \{\} actions of the plan. 
The next action to be executed is \{\}. 
Now you need to follow the multi-step plan to generate the next multi-step plan including 
\texttt{"visual\_state\_description"}, \texttt{"reasoning\_and\_reflection"}, 
and \texttt{"language\_plan"} in the format of a JSON object. 
Make sure the step number in the language plan starts from 1.
\end{tcolorbox}

\subsection{Rule-based Visual Description Generation for EB-Manipulation}
\label{app:rule_based}
To obtain consistent and interpretable ground-truth visual descriptions, we develop an algorithm that extracts object--color associations from the simulation environment and composes a structured, color-aware description of the scene. The algorithm proceeds in two stages: (i) extraction of object--color pairs via nearest-neighbor color classification in normalized RGB space, and (ii) synthesis of a natural-language description that integrates both semantic and spatial information.

\vspace{0.25em}
\noindent\textbf{Inputs.}
\begin{itemize}[leftmargin=1.5em]
  \item Simulator object tree $\mathcal{T}$; object type \textsc{SHAPE}.
  \item Canonical color map $\kappa:\mathcal{C}\to[0,1]^3$ (predefined color names $\rightarrow$ RGB in $[0,1]$).
  \item Aligned coordinate maps: $\mathrm{SimCoord}:\mathcal{N}\to\mathbb{R}^k$, $\mathrm{RealCoord}:\mathcal{R}\to\mathbb{R}^k$.
\end{itemize}

\noindent\textbf{Outputs.}
\begin{itemize}[leftmargin=1.5em]
  \item Sentence $S$: “From left to right, I can see \dots”
\end{itemize}

\noindent\textbf{Procedure:}
\begin{enumerate}[leftmargin=1.5em]
  \item \textbf{Extract shapes.} Query $\mathcal{T}$ for all \textsc{SHAPE} objects with \texttt{exclude\_base=true} to obtain $\mathcal{S}$.
  \item \textbf{Read attributes.} For each $s\in\mathcal{S}$, record its simulator name $\text{name}(s)$ and normalized color $\text{rgb}(s)\in[0,1]^3$.
  \item \textbf{Classify color.} For each $\text{rgb}(s)$, choose $f_{\text{col}}(s)=\arg\min_{c\in\mathcal{C}}\|\text{rgb}(s)-\kappa(c)\|_2$.
  \item \textbf{Build color map.} Set $\mathrm{color\_dict}[\text{name}(s)]\leftarrow f_{\text{col}}(s)$ (fallback \texttt{unknown} if missing).
  \item \textbf{Align names/coords.} Iterate over paired entries from \texttt{(sim\_name $\rightarrow$ coord)} and \texttt{(real\_name $\rightarrow$ coord)} (the alignment implied by the paired dictionaries).
  \item \textbf{Label composition.} For each pair, form a color-prefixed label:
        \begin{itemize}
          \item Preserve full real name for exceptions $\mathcal{E}=\{\text{``sponge''},\text{``shape sorter''}\}$.
          \item Otherwise, drop the first token of the real name and prefix with color (e.g., “blue block”).
        \end{itemize}
  \item \textbf{Article selection.} Choose “\textit{an}” if the label begins with a vowel, else “\textit{a}”.
  \item \textbf{Ordering (optional, recommended).} Sort objects by the x-coordinate to enforce true “left-to-right” enumeration.
  \item \textbf{Sentence assembly.} Convert each (label, coord) to the phrase “a/an \texttt{label} at \texttt{coord}”; join with commas and a final “and”.
\end{enumerate}

\subsubsection*{Pseudocode for \textsc{Generating Rule-based Visual Description}}
\begin{algorithm}[H]
\caption{\textsc{Generate}$(\mathrm{SimCoord}, \mathrm{RealCoord}, \mathrm{color\_dict})$}
\label{alg:generate}
\begin{algorithmic}[1]
\Require Aligned dictionaries of simulator names $\rightarrow$ coords and real names $\rightarrow$ coords; color map \texttt{color\_dict}
\Ensure Sentence $S$
\State $\mathcal{P} \gets [\,]$ \Comment{list of $(\text{label}, \mathbf{x})$}
\ForAll{paired entries $(\texttt{sim\_name}, \mathbf{x}) \in \mathrm{SimCoord}$ and $(\texttt{real\_name}, \mathbf{y}) \in \mathrm{RealCoord}$}
    \State $c \gets \mathrm{color\_dict}.\texttt{get}(\texttt{sim\_name}, \texttt{unknown})$
    \If{$\texttt{real\_name} \in \{\text{``sponge''}, \text{``shape sorter''}\}$}
        \State $L \gets c ~\Vert~ \texttt{real\_name}$ \Comment{preserve full name}
    \Else
        \State $L \gets c ~\Vert~ (\texttt{real\_name} \text{ with first token removed})$
    \EndIf
    \State Append $(L, \mathbf{x})$ to $\mathcal{P}$
\EndFor
\State \textbf{Optional:} sort $\mathcal{P}$ by the x-axis of $\mathbf{x}$ (left-to-right)
\State $\mathcal{D} \gets [\,]$
\ForAll{$(L, \mathbf{x}) \in \mathcal{P}$}
    \State $a \gets$ \textbf{``an''} if first character of $L$ is a vowel; else \textbf{``a''}
    \State Append string $a~L~\texttt{ at }~\mathbf{x}$ to $\mathcal{D}$
\EndFor
\State Construct sentence $S$ by joining elements of $\mathcal{D}$, prefixed with "From left to right, I can see ".
\State \Return $S$
\end{algorithmic}
\end{algorithm}

\subsection{Rule-based action mapping for EB-ALFRED}
\label{app:rule_based_alfred}
To build our environment-anchored prior datasets (\textit{masked action modeling} and \textit{action sequence reordering}), we start from the ALFRED training set. A central challenge is the mismatch between the action spaces of ALFRED and EB-ALFRED: ALFRED uses high-level, PDDL-style actions (e.g., ``CleanObject"), whereas EB-ALFRED uses a more granular set of phrase-based actions (e.g., ``turn on the faucet").

To reconcile this, we designed a deterministic, rule-based mapper that converts each high-level ALFRED action into a sequence of one or more EB-ALFRED actions. The conversion preserves the semantic intent of the original trajectories while conforming to the target action space. The mapping addresses several cases:
\begin{itemize}[leftmargin=*]
    \item \textbf{Composite Actions}:High-level verbs like ``CleanObject" expand into a logical sequence of primitives (e.g., locate a faucet, turn it on, then turn it off).
    \item \textbf{Context-Dependent Actions}: The mapping for "PutObject" depends on the receptacle. If the target (e.g., a microwave) must be opened first, the sequence is prepended with an ``open the ..." action.
    \item \textbf{Stateful Actions}: For ``ToggleObject," the mapping alternates between ``turn on the ..." and ``turn off the ..." based on the inferred object state.
\end{itemize}
The complete mapping from ALFRED actions to EB-ALFRED action sequences is detailed in Table \ref{tab:action_mapping}. After transforming an original action sequence into the new EB-ALFRED action space using these rules, we executed the mapped sequence in the simulator to verify that all mappings were valid.

\begin{table}[h!]
\centering
\caption{Rule-based mapping from ALFRED actions to EB-ALFRED action sequences. Arguments like ``{obj}" and ``{loc}" are placeholders for specific objects and locations from the original action.}
\label{tab:action_mapping}
\resizebox{\textwidth}{!}{%
\begin{tabular}{l|l}
\toprule
\textbf{ALFRED Action} & \textbf{Mapped EB-ALFRED Action Sequence} \\
\midrule
\texttt{GotoLocation(loc)} & \texttt{find a \{loc\}} \\
\midrule
\texttt{PickupObject(obj)} & \texttt{pick up the \{obj\}} \\
\midrule
\texttt{SliceObject(obj)} & \texttt{slice the \{obj\}} \\
\midrule
\texttt{ToggleObject(obj)} & Alternates between \texttt{turn on the \{obj\}} and \texttt{turn off the \{obj\}}. \\
\midrule
\texttt{NoOp()} & (Ignored; maps to an empty sequence) \\
\midrule
\texttt{PutObject(obj, loc)} & \begin{tabular}[t]{@{}l@{}}If `loc` requires opening (e.g., Fridge, Cabinet): \\ \quad \texttt{1. open the \{loc\}} \\ \quad \texttt{2. put down the object in hand} \\ Else: \\ \quad \texttt{1. put down the object in hand}\end{tabular} \\
\midrule
\texttt{CleanObject(obj)} & \begin{tabular}[t]{@{}l@{}}\texttt{1. put down the object in hand} \\ \texttt{2. find a Faucet} \\ \texttt{3. turn on the Faucet} \\ \texttt{4. turn off the Faucet} \\ \texttt{5. find a \{obj\}} \\ \texttt{6. pick up the \{obj\}}\end{tabular} \\
\midrule
\texttt{CoolObject(obj)} & \begin{tabular}[t]{@{}l@{}}\texttt{1. open the Fridge} \\ \texttt{2. put down the object in hand} \\ \texttt{3. close the Fridge} \\ \texttt{4. open the Fridge} \\ \texttt{5. find a \{obj\}} \\ \texttt{6. pick up the \{obj\}} \\ \texttt{7. close the Fridge}\end{tabular} \\
\midrule
\texttt{HeatObject(obj)} & \begin{tabular}[t]{@{}l@{}}\texttt{1. open the Microwave} \\ \texttt{2. put down the object in hand} \\ \texttt{3. close the Microwave} \\ \texttt{4. turn on the Microwave} \\ \texttt{5. turn off the Microwave} \\ \texttt{6. open the Microwave} \\ \texttt{7. find a \{obj\}} \\ \texttt{8. pick up the \{obj\}} \\ \texttt{9. close the Microwave}\end{tabular} \\
\bottomrule
\end{tabular}%
}
\end{table}
\clearpage
\section{Case Analysis}
\label{app:case_analysis}



\begin{figure}[h!]
\centering
\includegraphics[width=0.8\textwidth,page=2,trim=0 120 0 15,clip]{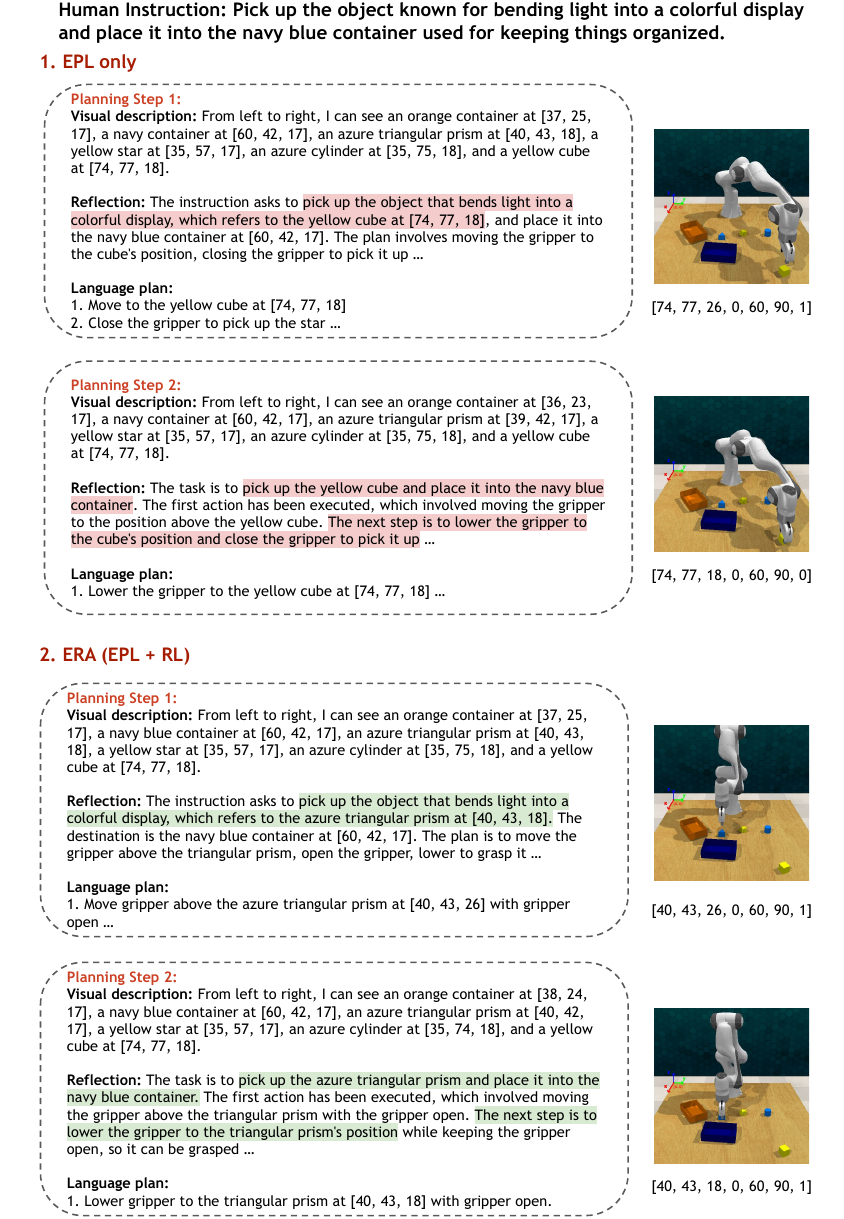}
\caption{Reflection error Example in EB-Manipulation. ERA successfully identified the correct target object: the azure triangular prism, while EPL mistakenly selected the yellow cube.}
\label{fig:eb_alfred_unsuccessful}
\end{figure}

\begin{figure}[h!]
\centering
\includegraphics[width=0.85\textwidth,page=3,trim=0 120 0 15,clip]{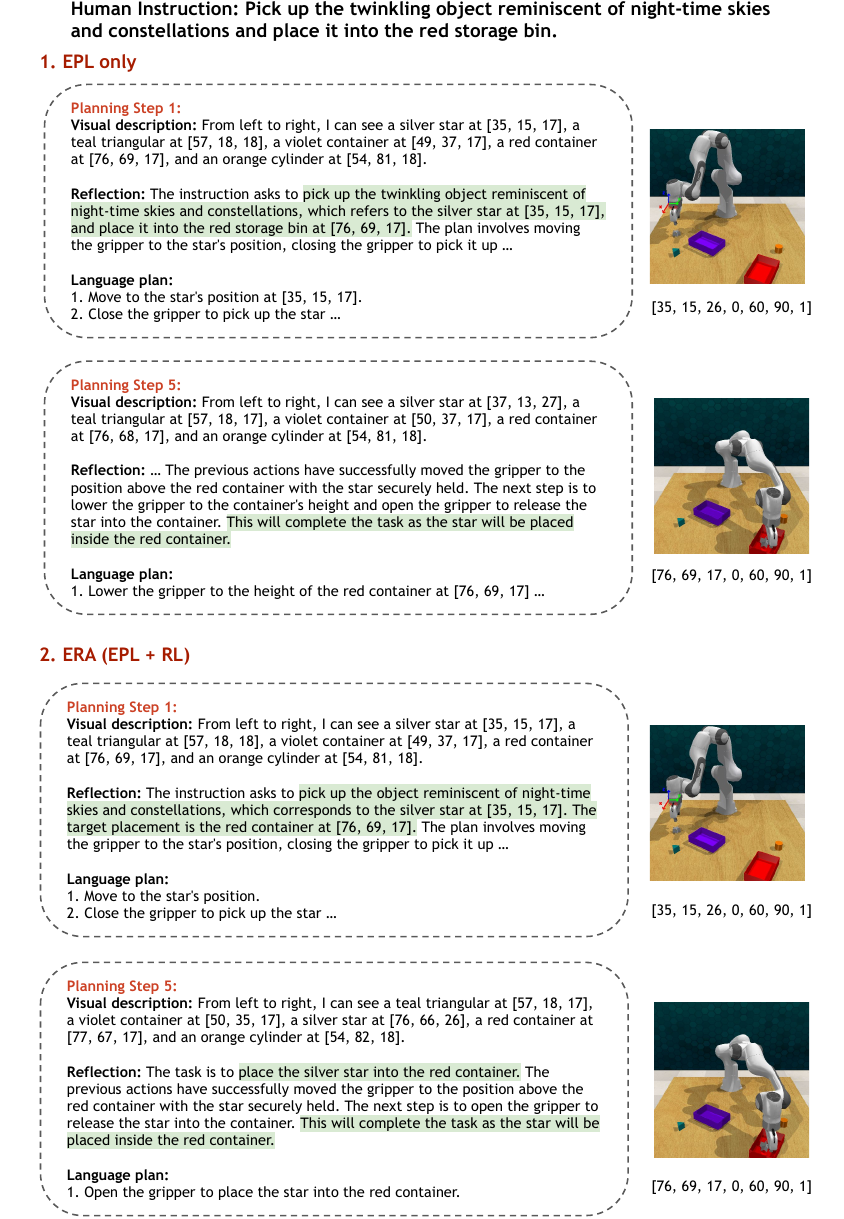}
\caption{Successful reflection Example in EB-Manipulation. Both agents were able to identify the silver star as the target object.}
\label{fig:eb_alfred_successful}
\end{figure}

\begin{figure}[h!]
\centering
\includegraphics[width=0.85\textwidth,page=5,trim=0 120 0 15,clip]{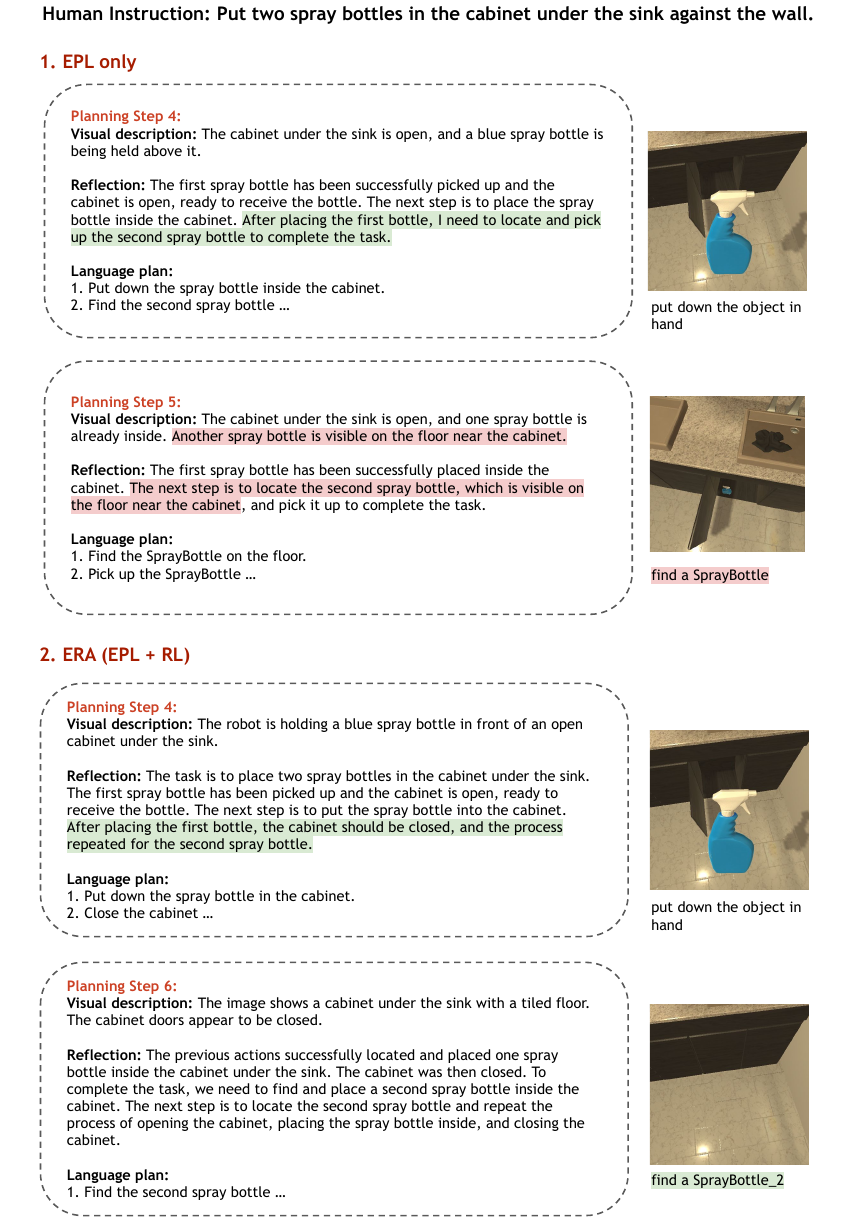}
\caption{Planning error example in EB-ALFRED. ERA successfully identified the second spray bottle as SprayBottle\_2 while EPL repeatedly located the same SprayBottle.}
\label{fig:eb_man_unsuccessful}
\end{figure}

\begin{figure}[h!]
\centering
\includegraphics[width=0.85\textwidth,page=6,trim=0 120 0 15,clip]{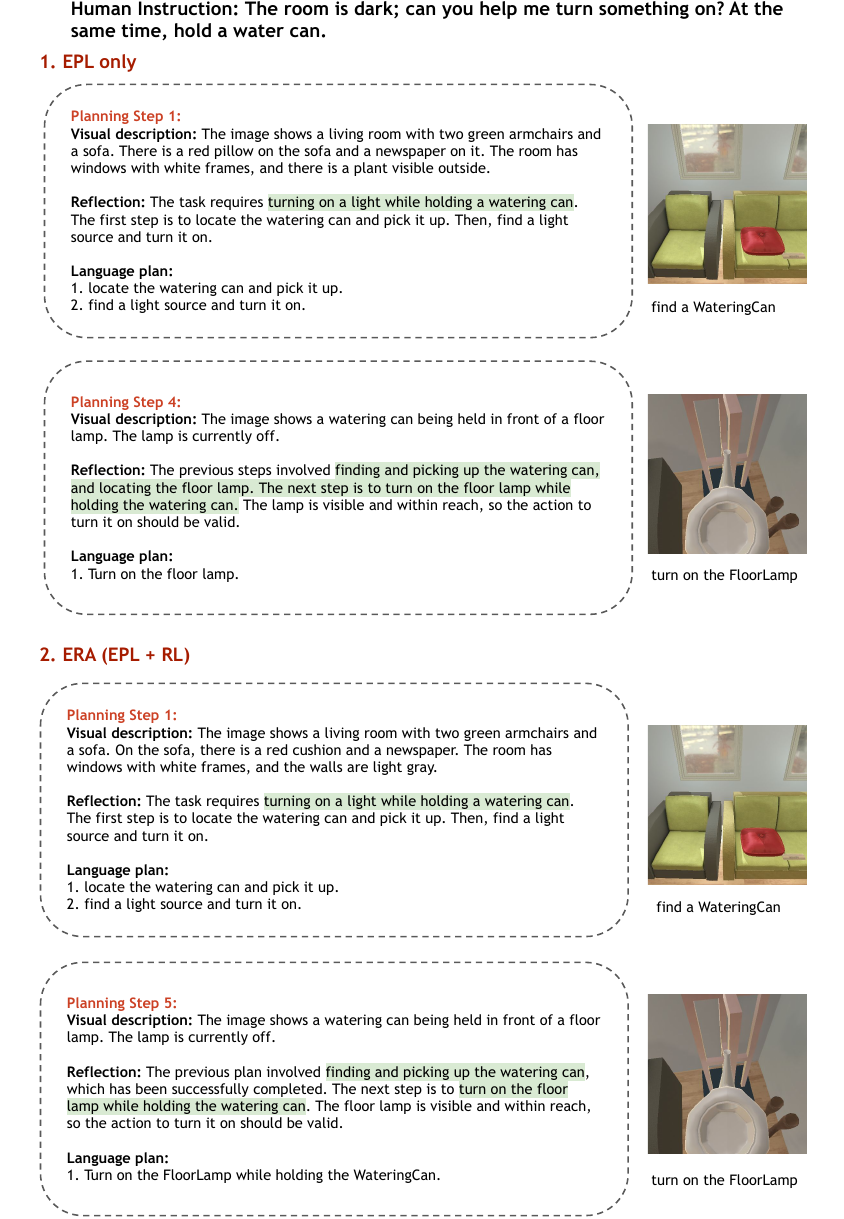}
\caption{Successful reflection example in EB-ALFRED. Both agents were able to identify the FloorLamp as the target object.}
\label{fig:eb_man_successful}
\end{figure}

In this section, we present both successful and unsuccessful planning examples for \textbf{EPL only} and \textbf{ERA (EPL + RL)} across EB-ALFRED and EB-Manipulation. Refer to Figures~\ref{fig:eb_alfred_unsuccessful},~\ref{fig:eb_alfred_successful},~\ref{fig:eb_man_unsuccessful}, and~\ref{fig:eb_man_successful} for detailed reasoning and planning.

\end{document}